\documentclass{article}

\usepackage[final]{corl_2026} 

\title{Articulated 3D Scene Graphs \\ for Open-World Mobile Manipulation}

\author{
  Martin Büchner\textsuperscript{\footnotesize 1} \quad
  Adrian Röfer\textsuperscript{\footnotesize 1} \quad
  Tim Engelbracht\textsuperscript{\footnotesize 2} \quad
  Tim Welschehold\textsuperscript{\footnotesize 1} \\
  \textbf{Zuria Bauer\textsuperscript{\footnotesize 2} \quad
  Hermann Blum\textsuperscript{\footnotesize 2,3} \quad
  Marc Pollefeys\textsuperscript{\footnotesize 2} \quad
  Abhinav Valada\textsuperscript{\footnotesize 1}} \\
  \\
  \textsuperscript{\footnotesize 1}University of Freiburg \quad
  \textsuperscript{\footnotesize 2}ETH Zürich \quad
  \textsuperscript{\footnotesize 3}University of Bonn \\
}

\usepackage{times}

\usepackage[numbers]{natbib}
\usepackage{multicol}
\usepackage{graphicx}
\usepackage{siunitx}
\usepackage{booktabs}
\usepackage{multirow}
\usepackage{tabularx}
\usepackage{siunitx}
\usepackage{threeparttable}
\usepackage{array}
\usepackage{enumitem}
\usepackage{microtype}
\usepackage[table, dvipsnames]{xcolor}
\usepackage{tikz}
\usepackage{pgfplots}
\usepackage{leftindex}
\usepackage{amssymb}
\usepackage{float}
\usepackage{textcomp}
\usepackage{hyperref}

\usepackage{pgfplots}
\pgfplotsset{compat=1.18}
\usepgfplotslibrary{groupplots}

\usepackage{amsmath}
\usepackage{amssymb}

\DeclareMathAlphabet{\mathcal}{OMS}{cmsy}{m}{n}

\def\R{\mathbb{R}}

\def\seThree{\mathfrak{se}(3)}
\def\SEThree{SE(3)}

\newcommand{\norm}[1]{\left\lVert#1\right\rVert}

\newcommand{\prob}[1]{p\left(#1\right)}
\newcommand{\probc}[2]{\prob{#1 \;\middle\vert\; #2}}

\newcommand{\set}[1]{\left\{#1\right\}}

\newcommand{\tuple}[1]{\left\langle#1\right\rangle}

\usepackage{xspace}

\newcommand{\eg}{\mbox{e.\,g.}\xspace}
\newcommand{\ie}{\mbox{i.\,e.}\xspace}

\newcommand{\etal}{\emph{et al.}\xspace}

\renewcommand{\[}{\begin{equation}}
\renewcommand{\]}{\end{equation}}

\renewcommand{\baselinestretch}{0.99}

\definecolor{red}{RGB}{255, 0, 0}   
\definecolor{orange}{RGB}{255, 77, 0}   
\definecolor{green}{RGB}{0, 128, 0}   
\definecolor{purple}{RGB}{160, 32, 240}   
\definecolor{lightblue}{RGB}{52, 155, 235}   
\definecolor{darkmagenta}{RGB}{204, 51, 139} 

\usepackage[capitalize]{cleveref}

\crefname{figure}{Fig.}{Figs.}
\Crefname{figure}{Figure}{Figures}
\crefname{section}{Sec.}{Secs.}
\Crefname{section}{Section}{Sections}
\Crefname{table}{Table}{Tables}
\crefname{table}{Tab.}{Tabs.}
\crefname{algorithm}{Algo.}{Algos.}
\Crefname{algorithm}{Algorithm}{Algorithms}
\crefname{appendix}{Sec.}{Secs.}
\Crefname{appendix}{Section}{Sections}

\usepackage{amsmath}
\usepackage{bigdelim}
\usepackage{url}

\usepackage{flushend}
\usepackage{colortbl}
\usepackage{pifont}
\usepackage{tcolorbox}

\usepackage[ruled,vlined]{algorithm2e}

\usepackage{pgfplots}
\pgfplotsset{compat=1.18}
\usepgfplotslibrary{fillbetween}
\usepgfplotslibrary{groupplots}

\pgfplotsset{
    myplotstyle/.style={
        width=5.5cm,
        height=5cm,
        xlabel={$\eta$},
        xlabel style={font=\small},
        ylabel style={font=\small},
        tick label style={font=\footnotesize},
        grid=major,
        grid style={dashed, gray!40},
        xmin=0.98, xmax=1.0,
        xtick={0.98, 0.985, 0.990, 0.995, 1.0},
        xticklabel style={rotate=45, anchor=east, font=\footnotesize},
        minor tick num=1,
        legend style={font=\tiny, at={(0.97,0.97)}, anchor=north east},
        thick,
    }
}

\usepackage{caption}
\usepackage{subcaption}
\newcommand{\ours}{MoMa-SG}
\newcommand{\ourdataset}{Arti4D-Semantic}

\newcommand{\greyrule}{\arrayrulecolor{black!30}\midrule\arrayrulecolor{black}}

\usepackage[capitalize]{cleveref}
\crefname{section}{Sec.}{Secs.}
\Crefname{section}{Section}{Sections}
\Crefname{table}{Table}{Tables}
\crefname{table}{Tab.}{Tabs.}
\crefname{algorithm}{Alg.}{Algs.}

\begin{document}
\pagestyle{empty}
\maketitle


\begin{abstract}
Semantic scene understanding has enabled robots to reason and perform affordance-driven object interaction. Despite this, real-world manipulation remains fundamentally limited by the inability to predict how objects move and are articulated. Long-horizon mobile manipulation requires closing the gap between semantics, geometry, and kinematics. In this work, we present \emph{MoMa-SG}, a novel framework for building semantic-kinematic 3D scene graphs of articulated scenes containing a myriad of interactable objects. Given RGB-D sequences containing multiple object articulations, we temporally segment object interactions and infer object motion using occlusion-robust point tracking. We lift point trajectories to 3D and estimate articulation models using a novel unified twist estimation formulation that robustly estimates revolute and prismatic joint parameters in a single optimization pass. Next, we associate objects with articulations and detect contained objects by reasoning over parent-child relations at identified opening states. We also introduce the \emph{Arti4D-Semantic} dataset, which provides parent-child relation annotations between container and contained objects together with object axis annotations across 62 in-the-wild RGB-D sequences containing 600 object interactions and three distinct observation paradigms. We extensively evaluate the performance of \emph{MoMa-SG} on two datasets and ablate key design choices of our approach. In addition, real-world experiments on both a quadruped and a mobile manipulator demonstrate robust manipulation of articulated objects in everyday home environments. 
We provide code and data at: \url{https://momasg.cs.uni-freiburg.de}.
\end{abstract}


\begin{figure}[h]
\vspace{-0.3em}
\centering
\includegraphics[trim=0 14cm 0 0cm, clip,width=0.995\columnwidth]{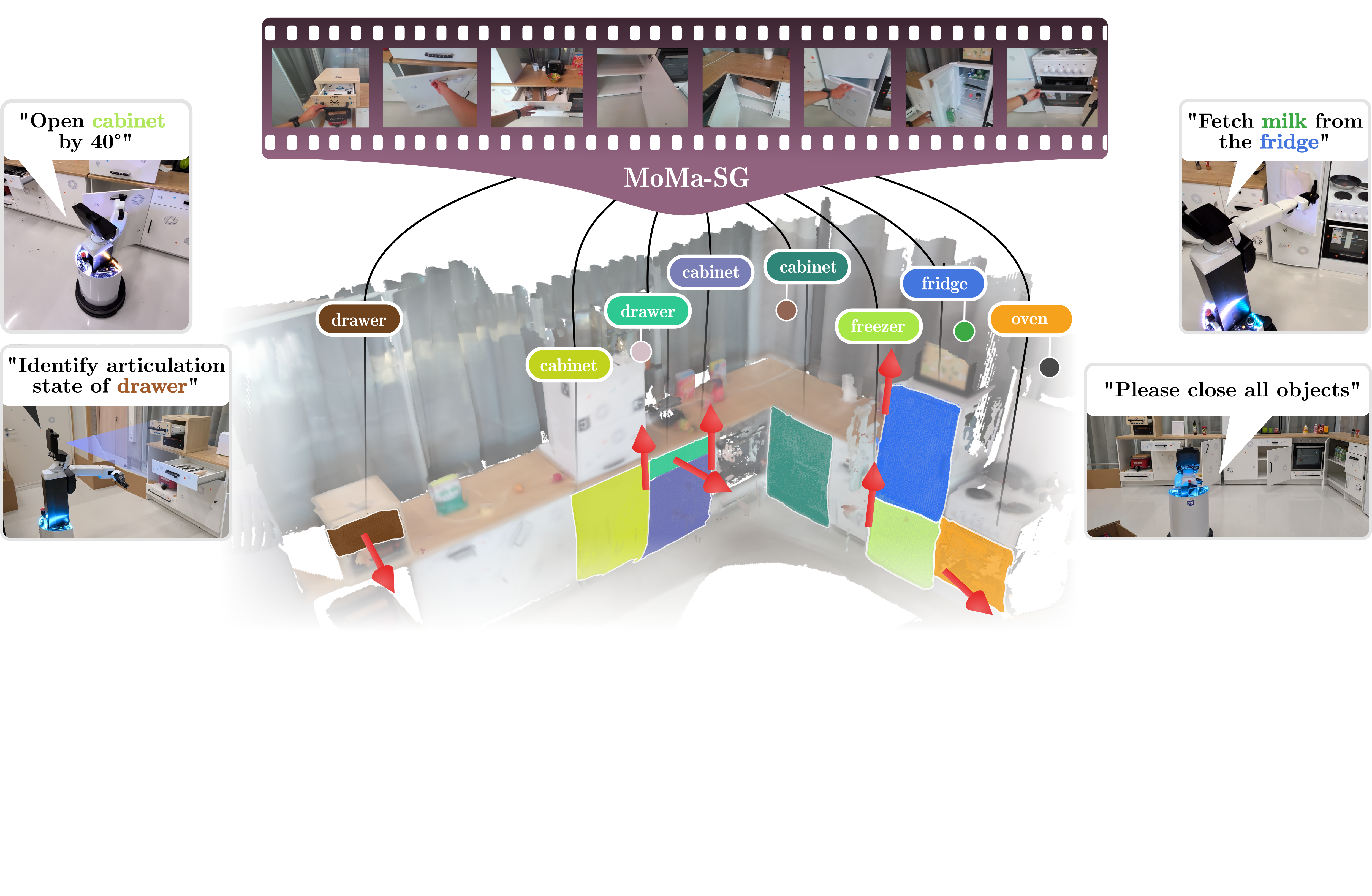}
\caption{\color{black}\ours{} enables the construction of accurate semantic-kinematic 3D scene graphs over articulated scenes and serves as a backbone for long-horizon and language-enabled interactive mobile manipulation.}
\label{fig:teaser}
\vspace{-2em}
\end{figure}


\section{Introduction}
\label{sec:introduction}
Traditional approaches in robotic mapping focus on building metrically accurate representations of quasi-static environments. Recently, 3D scene graphs have proposed sparse abstractions of dense maps that allow structured reasoning over objects, their spatial relationships, and higher-level scene layouts~\cite{armeni20193d, hughes2022hydra, greve2024collaborative}. Beyond purely metric-semantic topological representations, 3D scene graphs have further been equipped with open-vocabulary semantic understanding~\cite{gu2024conceptgraphs, hovsg, koch2024open3dsg, maggio2024clio}, enabling planning from language-specified goals~\cite{honerkamp2024language, more25mohammadi}. Furthermore, these representations increasingly encode functional and affordance-level information, capturing how objects can be interacted with~\cite {zhang2025open, rotondi2025fungraph,engelbracht2024spotlightroboticsceneunderstanding, rotondi2026survey}.

While the aforementioned methods provide a comprehensive manifold for reasoning, planning, and navigation, mobile manipulation tasks require a certain level of kinematic understanding of the surroundings. A significant number of objects in human environments are containers, such as cabinets and drawers, which require an in-depth understanding of their kinematic behavior in order to enable compliant manipulation.
While zero-shot priors of articulation axes are useful, articulated objects can be visually ambiguous, and thus, only observing object interactions yields dependable models that accurately govern their motion. Previous works on articulated object estimation~\cite{katz2008manipulating, arti25werby, buchanan2024online, heppert2022category} are often limited to synthetic data, fiducial markers, or few viewpoints~\cite{liu2023paris, martin2019rbo, xiang2020sapien}.

In this work, we present \ours{} for modeling semantic-kinematic scenes as articulated 3D scene graphs. The ultimate aim of \ours{} is to endow mobile manipulators with the ability to plan, navigate, and manipulate in interactive environments containing numerous articulated objects. We strive to accomplish this from single object interactions, \ie, from in-the-wild RGB-D observations of a human interacting with a series of articulated objects (see \cref{fig:teaser}). We address this problem by one-shot distillation of those observations into a semantic-kinematic 3D scene graph hierarchy. The proposed graph holds explicit kinematic information, \ie, the underlying articulation model for each articulated object, and the semantic relationship among containers and contained objects inferred through interaction. We deliberately avoid assumptions about a fixed set of semantic categories, camera viewpoints, and acting embodiments, allowing \ours{} to operate across a wide range of robots. The observation modes include human demonstrations from both ego-centric, exo-centric as well as robot-centric camera trajectories. We evaluate our proposed method on two distinct mobile manipulators for the task of identifying articulation states as well as opening and closing various articulated objects based on the estimated 3D scene graph. In summary, we make the following contributions:
\begin{itemize}
    \item A unified framework for constructing articulated 3D scene graphs from ego-centric, exo-centric, or robot-centric observations that enables one-shot open-world mobile manipulation.
    \item A novel twist optimization objective that allows for robust estimation of revolute and prismatic articulations from noisy, real-world point-level trajectories.
    \item The novel Arti4D-Semantic benchmark that bridges the fields of semantic scene understanding and object articulation understanding in in-the-wild settings.
    \item We demonstrate real-world mobile manipulation from our articulated scene graphs in two distinct environments across two robotic embodiments.
    \item We make the dataset and code publicly available to foster future research in this domain.
\end{itemize}
\section{Related Work}

\noindent{\textbf{Mobile Manipulation from Kinematic Environment Models}:}
Given a kinematic model of an articulated object, it can be manipulated reliably using a form of compliant control~\cite{jain2009pulling,ruhr2012generalized,buchanan2024online}, even in mobile settings. Arduengo~\etal~\cite{arduengo2021robust} demonstrate learning articulations from a human demonstration and then acts on these with a mobile robot. KinScene~\cite{hsu2024kinscene} presents a similar system, though with the robot exploring articulations itself in a similar manner as~\cite{ruhr2012generalized}. Other approaches use kinematic models to drive robotic controllers~\cite{stuede2019door,mittal2022articulated}. While grasping is commonly the focus of these interactions, \cite{rofer2022kineverse} shows how to generate pushing interactions, including contact point selection. Using only forward-kinematic prediction from an articulation model is also sufficient to realize mobile manipulation through learned controllers~\cite{honerkamp2023n}.

\noindent{\textbf{Articulated Object Estimation}:}
Katz~\etal~\cite{katz2008manipulating} presented the first approach, extracting 2D articulations and processing relative motions of point clusters. Sturm~\etal~\cite{sturm2011probabilistic} extend this to a full framework for extracting kinematic graphs from observations of relative object motions. 
As most articulated objects in everyday life feature a single degree of freedom (DoF), high-dim. representations have been reduced to a single vector in $\mathfrak{se}(3) \subseteq \R^6$~\cite{jain2021screwnet,heppert2022category}. ArtiPoint~\cite{arti25werby} combines point track observations with twist estimation from in-the-wild observations. The most recent approaches utilize Gaussian Splatting or Neural Radiance Fields (NeRF) to reconstruct articulated objects~\cite{jiang2022ditto,liu2025building,liu2023paris}. All of these methods require objects to be observed in different articulation states. To overcome this challenge robotically, a long line of work has investigated the use of interaction to produce motions in a suspected articulated object~\cite{jain2009pulling,ruhr2012generalized,martin2014online,martin2022coupled}. These approaches require prior knowledge of the type of interaction that might cause motion. Learning this type of prior from visual observations has been studied by~\cite{mo2021where2act,xu2022universal,eisner2022flowbot3d,zhang2023flowbot++}. Buchanan~\etal~\cite{buchanan2024online,buchanan2026online} couple learned priors with robotic interaction and tactile exploration to estimate articulations. 
Closest to \ours{}, ArtiPoint~\cite{arti25werby} shares a similar point tracking paradigm; however utilizes a less robust estimator for axis estimation.

\noindent{\textbf{3D Scene Graphs}:}
3D Scene Graphs (SGs) unify geometry, semantics, and relational reasoning for robotic perception. Early work constructed SGs from static 3D reconstructions~\cite{rotondi2026survey, armeni20193d, wald2020learning3dsg}, later extended to incremental RGB-D fusion with online updates and persistent object identity~\cite{wu2021scenegraphfusionincremental3dscene}. More recent approaches introduce open-vocabulary and hierarchical SGs that support unconstrained semantic labeling and language grounding, primarily for navigation and long-horizon exploration~\cite{gu2024conceptgraphs, hovsg}. Dynamic SG formulations model temporal graph evolution and persistent world memory, but typically assume rigid object models~\cite{Rosinol20rss-dynamicSceneGraphs}. Scene graphs have also been leveraged as an interface between perception and task planning~\cite{ju2025momagraph}.
A complementary line of work treats functionality and affordances as first-class graph components, introducing interaction-relevant nodes and relations bridging perception and action~\cite{rotondi2025fungraph, zhang2025open, engelbracht2024spotlightroboticsceneunderstanding}. Only very recent efforts integrate articulation into SGs by modeling parts and kinematic relations at the scene level~\cite{yu25pandora}.

Closest to our work is Pandora~\cite{yu25pandora}, however, it assumes full object visibility and fixed open-close interaction patterns, whereas \ours{} supports partial observability, diverse articulation motions, and embodiment-agnostic mobile manipulation.
\section{Approach}
\label{sec:method}
\begin{figure*}
\includegraphics[width=\linewidth]{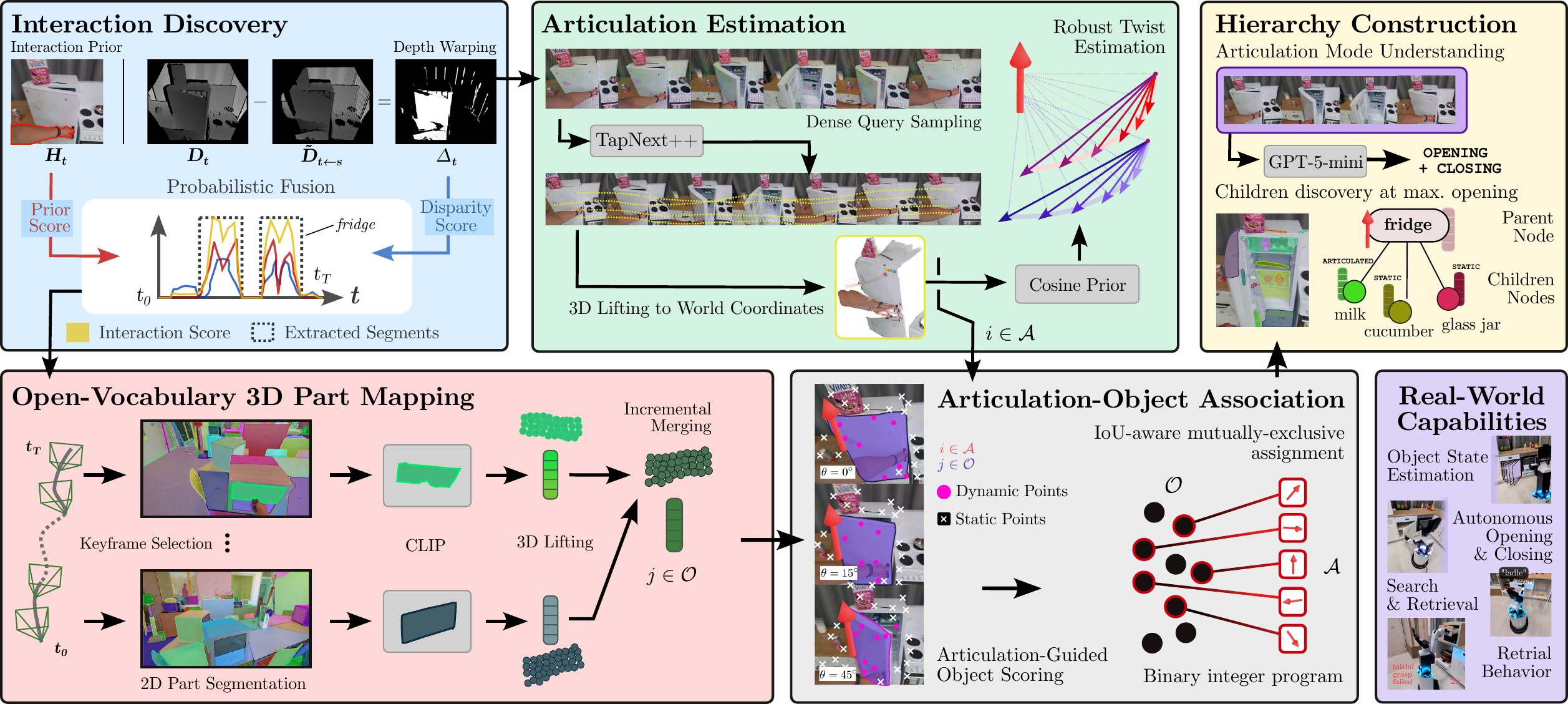}
\caption{\color{black} Overview of \ours{}: We first discover interaction segments (\cref{sec:interaction_disc}), then attain object articulation models $\mathcal{A}$ by estimating twists from point trajectories (\cref{sec:articulation_estimation}). We match mapped objects $\mathcal{O}$ against articulations and discover objects contained in respective articulated parents (\cref{sec:articulated_scene_graph}).}
\label{fig:overview}
\vspace{-2.0em}
\end{figure*}

In the following, we detail our proposed approach \ours{} as depicted in \cref{fig:overview} for building accurate and actionable semantic-kinematic 3D scene graphs from articulation demonstrations. We first construct a set of interaction segments (Sec.~\ref{sec:interaction_disc}), perform point tracking within those segments, and estimate object articulation models (Sec.~\ref{sec:articulation_estimation}). This is followed by a 3D part mapping stage, an object-to-articulation association, and the discovery of contained objects (\cref{sec:articulated_scene_graph}).

\subsection{Interaction Discovery}
\label{sec:interaction_disc}
Given a scene-level RGB-D sequence $\set{\mathbf{I}_t, \mathbf{D}_t}_{t=1}^T$ of consecutive interactions with multiple articulated objects, we first extract a set of $N$ temporal interaction segments $\mathcal{S} = \{(t_{s}^{(n)}, t_{e}^{(n)})\}_{n=1}^N$ capturing dynamic object motion. To do this efficiently, we propose combining two independent signals: First, an interaction prior via segmentation of, \eg, hands or end-effectors, and second, a depth disparity measure that encodes 3D scene dynamics. These complementary signals allow tracking interactions with low dynamics or strong occlusions, \eg, hidden hands when opening large doors. For each RGB frame $\mathbf{I}_t$, we obtain the interaction prior as a segmentation mask $\mathbf{H}_t$ using YOLOv11~\cite{khanam2024yolov11}. We compute the depth disparity signal within a time window $s_H$ between $\mathbf{D}_t$ and $\mathbf{D}_{t-s_H}$ by warping the previous depth map into the current view and measuring the depth deviation. We apply normalization and median filtering, and fuse those two signals using a conditional probability model, see Appendix~\ref{sec:interaction_disc_supp}.

\subsection{Articulation Estimation}
\label{sec:articulation_estimation}
For each segment $s_n \in \mathcal{S}$, we optimize a 1-DoF model of the articulation $\mathcal{A}_i = \{\xi, \theta_s,\ldots,\theta_e,M\}$, where $\xi$ parametrizes the articulation model, $\theta_j$ are articulation states, i.e. the opening states at each time-step $t_j \in s_n$, and $M$ is a semantic token characterizing the observed motion.

\noindent{\textbf{3D Point Tracking and Filtering}}: 
In order to be occlusion-robust, we opt for a point tracking-based approach in favor of a meshing approach as chosen by Pandora~\cite{yu25pandora}. While ArtiPoint~\cite{arti25werby} initializes its tracking process from keypoints sampled around hand masks, \ours{} employs dense grid sampling of keypoints at the initial and terminal frames. Next, we employ TapNext++~\cite{tapnextpp2026} to obtain point tracks $\mathcal{X}_n \in \R^{T_n \times F \times 2}$ across all $T_n$ frames of $s_n$, including their visibilities $\mathcal{V}_n \in \{0, 1\}^{T_n \times F}$ throughout each interaction.
We then lift $\mathcal{X}_{n}$ to 3D using RGB-D depth and compensate for camera motion by transforming into world coordinates. 
Next, we smooth point tracks~\cite{arti25werby} and separate $\mathcal{X}_{n}$ into static and dynamic points. We filter out tracks with negligible length and employ DBSCAN using a rigidity metric in order to obtain a coherent cluster spanning the most extensive spatial volume, as these tracks represent the full range of the articulation.

\noindent{\textbf{Regularized Twist Estimation}:}
We estimate object articulation models using screw theory, which allows us to represent arbitrary rigid-body motion. A screw motion is parameterized by a twist $\xi = \tuple{\omega, v} \in \seThree$, where $\omega \in \mathbb{R}^{3}$ and $v \in \mathbb{R}^{3}$ are the rotational and translational velocity components. In the context of articulations, a key advantage of the twist representation is its capacity to model diverse articulation types, including revolute, prismatic, and screw joints, while also being differentiable, thereby enabling gradient-based optimization~\cite{jain2021screwnet,heppert2022category,buchanan2024online, arti25werby, kim2025screwsplat}. Furthermore, given a configurational parameter $\theta$, we are able to produce the equivalent $\SEThree$ rigid-body transform using the \emph{exponential map} $\text{exp} : \seThree \rightarrow \SEThree$, while the inverse of the exponential map is referred to as the \emph{logarithmic map} $\text{log}: \SEThree \rightarrow \seThree$, converting back to tangent space~\cite{barfoot2024state}. 

Despite its advantages, we find that the twist parameterization is prone to observational noise and keypoint drift when applied directly on point trajectories in uncontrolled settings~\cite{arti25werby, buchanan2024online}, which is why one typically first estimates object poses and then applies pose-based twist estimation. However, fitting representative local frames to objects observed in-the-wild remains challenging. In order to resolve this, we propose a novel regularization scheme for point-based twist estimation that addresses the magnitude of $\omega$ and $v$. We find that under real-world observations, \eg, pure prismatic motion, which ideally would represent a screw motion of infinite pitch, often shows a non-negligible rotational component $\|\omega\|$. Thus, when employing vanilla twist estimation from point trajectories, one could determine the underlying articulation type (prismatic or revolute) from the ratio $\|\omega\| / \|v\|$ and parse the twist by, \eg, neglecting the rotational component of an identified prismatic motion. In practice, finding such a decision rule is only possible while incurring significant type and axis errors.

Instead, we propose a scaled dot product prior among vectors cast from the observed point trajectories, which are secants of circles with ideally finite (revolute) or infinite (prismatic) radius. The reasoning behind this is that vectors drawn from prismatic motions generally show little to no angular deviation. On the contrary, revolute motions produce distinct angular disagreements. As such, we sample vectors of the form $\mathbf{x}_i = \mathbf{p}_i^\text{t} - \mathbf{p}_i^\text{0}$, where $\mathbf{p}_i^\text{0} \in \mathbb{R}^3$ represents an initial point of a track and $\mathbf{p}_i^\text{t} \in \mathbb{R}^3$ represents a point along the track. From this, we construct a set of vectors $\mathbf{X}\in\mathbb{R}^{M \times 3}$ across all point tracks for which we compute the scaled dot product of all pairs $C_{ij} = \frac{\mathbf{x}_i^\top\mathbf{x}_j}{\|\mathbf{x}_i\| \, \|\mathbf{x}_j\|}$,
where we enforce that $\mathbf{x}_i$ and $\mathbf{x}_j$ do not \textit{terminate} at the same time step $t$. 
Next, we compute the median $\eta$ of the upper-triangular component of the matrix holding all $C_{ij}$ and feed $\eta$ to a shifted sigmoid function $\sigma(k(\eta-\eta^{*}))$ with bias $\eta^{*}$, which is a tunable hyperparameter. We impose that for revolute motion, the absolute dot product $|\omega^{\top}v| \approx 0$ as $\omega \perp v$. Similarly, for prismatic motion, we expect a diminishing rotational term $\|\omega\|$ as solely $v$ governs the observed translational motion. 
We define $\lambda_{pris}=\sigma(k(\eta-\eta^{*}))$ and $\lambda_{rev}=1-\sigma(k(\eta-\eta^{*}))$ and obtain the following optimization problem:
\begin{equation}
\label{eq:articulation_regularized_opt}
\begin{aligned}
\hat{\xi}^*, \theta^*_1, \ldots, \theta^*_e
&= \min_{\hat{\xi}, \theta_s, \ldots, \theta_e}
\sum_{m=s}^e \sum_{j=1}^{|P_m|}
\norm{p^{t+\epsilon}_j - \exp(\hat{\mathbf{\xi}}\theta_m) \cdot p^t_j} + \alpha \left(
\lambda_{rev} \left| \frac{\omega^{\top}v}{\norm{\omega}\norm{v}} \right|
+ \lambda_{pris} \|\omega\|
\right),
\end{aligned}
\end{equation}
where the first term represents the original twist optimization from point trajectories and the second term constitutes our dot product prior with $\alpha$ as an optional weighting term. We solve for an optimal twist $\xi$ as well as observed configurations $\theta_m^{*}$ using non-linear least-squares optimization. Note that the regularization term is fully differentiable while we do not suffer a loss of generality.

\noindent{\textbf{Mode Understanding}:}
The obtained articulation estimate in Eq.~\ref{eq:articulation_regularized_opt} yields configurational scalars $\theta$. In general, however, we cannot infer from that whether we, \eg, observed an opening or a closing action. Thus, we feed a small set of informative frames contained in $s_n$ to \textit{GPT-5-mini} and prompt for the observed articulation mode $M_i$, which is constrained to be \texttt{OPENING}, \texttt{CLOSING}, or combined motions such as \texttt{OPENING-CLOSING} or \texttt{CLOSING-OPENING}. We use a simple heuristic as a critic to verify whether the VLM output is consistent with the distribution of $\theta$'s. 

\subsection{Articulated 3D Scene Graph Construction}
\label{sec:articulated_scene_graph}
Building on the previous results, we construct a hierarchical 3D scene graph $\mathcal{G}=(\mathcal{V},\mathcal{E})$ consisting of an object node layer $\mathcal{O}$ representing objects and a layer of contained objects $\mathcal{C}$ that represent children nodes $k$ to $j \in \mathcal{O}$. A small subset $\mathcal{O}^{\mathcal{A}} \subseteq \mathcal{O}$ corresponds to the articulated objects associated with articulations $\mathcal{A}$. This representation is supported by a geometric point cloud layer. In this work, we assume that the observed articulations $\mathcal{A}$ are either prismatic or revolute. An edge $(j,k)$ thus connects an articulated parent object $j$ to a contained child node $k \in \mathcal{C}$. We note that child nodes are either of static or articulated type, \ie, they either follow the same articulation model as their parent or remain static. Intuitive examples for this are a mug hidden behind the kitchen cabinet door as opposed to a milk carton that sits in the fridge door and moves along with its parent. In our approach, we construct $\mathcal{O}$ and $\mathcal{A}$ independently and later match objects against articulations to obtain $\mathcal{O}^{\mathcal{A}}$.

\noindent{\textbf{3D Part Mapping}:} In order to construct $\mathcal{O}$, we employ an incremental mapping framework that fuses 2D segmentation masks from \emph{Semantic-SAM}~\cite{li2023semantic} across non-interaction keyframes, which provides dependable part instances. To allow for open-world deployment, we equip each object mask $o \in \mathcal{O}$ with a CLIP feature to render it semantically retrievable using open-vocabulary scoring. In contrast to \emph{ConceptGraphs}~\cite{gu2024conceptgraphs}, we do not rely on semantic similarities during mask merging, as it frequently produces erroneous merges for object parts. We incrementally merge new detections into existing objects while evaluating whether the object and detection are \textit{isolated} masks, \ie, they are not intersecting the image boundaries. We detail these rules in Appendix Sec.~\ref{sec:part_mapping_supp}.

\noindent{\textbf{Matching Articulations Against Objects}:}
In the following, we identify the object parts $\mathcal{O}^{\mathcal{A}} \subseteq \mathcal{O}$ that are undergoing interactions given the estimated articulations $\mathcal{A}$. Per segment $s_n$, we first retrieve the $k$-closest objects in $\mathcal{O}$ given their 3D centroids, and the mean position of the respective 3D point tracks. As we aim to find the 3D part segment that, under motion, best explains the observed point trajectories, we \textit{replay} close-by objects in $\mathcal{O}$ under the estimated articulation model $\mathcal{A}_{i}$ and evaluate whether the motion pattern matches with our 2D point tracks $\mathcal{X}_n$. 
In particular, we measure the ratio of dynamic keypoints falling into the mask of object $j$ as well as static keypoints falling into its exterior. Summing these two terms yields $p_{ij}$ representing the cost of matching articulation $i$ and object $j$, which we further detail in the Appendix Sec.~\ref{sec:supp_obj_articulation_matching}.
In general, we face under- and over-segmentation of parts, leading to conflicting masks. To compensate for this, we strive to find a mutually exclusive object-to-articulation matching. This comes with two distinct considerations: a) we aim to avoid assigning a single object $j$ to two different articulations in the case of under-segmentation, and b) we aim to minimize mutual 3D overlap among assigned objects. We thus formulate a binary integer program (BIP) that ensures that a single object $j \in \mathcal{O}$ is only matched against an articulation $i \in \mathcal{A}$ once while minimizing the mutual 3D overlap among all assigned object parts:
\begin{equation}
    \underset{m,y,z}{\operatorname{min}} \sum_{i\in \mathcal{A}}\sum_{j\in\mathcal{O}}{p_{ij}m_{ij}} + \lambda\sum_{(j,k)\in \mathcal{P}}{q_{jk}z_{jk}},
\end{equation}
where $p_{ij}$ is the cost of assigning articulation $\mathcal{A}_{i}$ to object $j$ as described above, $q_{jk} \geq 0$ represents the IoU between object $j$ and $k$. In addition, $m_{ij} \in \{0,1\}$ becomes 1 if articulation $i$ is assigned to object $j$, $y_{j} \in \{0,1\}$ becomes $1$ if object $j$ is assigned to any articulation, and lastly, $z_{jk} \in \{0,1\}, j,k \in \mathcal{O}$ denotes if both $y_j=1$ and $y_k = 1$. In order to satisfy the above requirements, we impose the following constraints: 
\begin{equation}
\begin{aligned}
    \sum_{j\in\mathcal{O}} m_{ij}= 1\,\,\forall i \in \mathcal{A}, \quad 
    \sum_{i\in\mathcal{A}} m_{ij} \le 1 \,\,\forall j \in \mathcal{O}, \quad
    z_{jk} \le y_j, z_{jk} \le y_k, \text{ and } z_{jk} \geq y_j + y_k -1, 
\end{aligned}
\end{equation}
with the last enforcing $z_{jk} = y_j\,\wedge y_k$. Solving this BIP thus provides us with a set of objects $\mathcal{O}^{\mathcal{A}}$ constituting the optimal assignment of articulations $\mathcal{A}$ and objects $\mathcal{O}$.

\noindent{\textbf{Discovery of Contained Objects}:} 
In the final stage of our approach, we discover contained objects and arrange them as child nodes $\mathcal{C}$ of the found articulated parents $\mathcal{O}^{\mathcal{A}}$. We achieve this by evaluating the estimated articulation mode $M$ and identifying the configuration $\theta_{max}$ representing the maximum-open frame. We then retrieve the 2D projected articulated parent mask under two configurations: maximum opening and closed states. Subsequently, we query a class-agnostic instance segmentation method~\cite{mobile_sam} and identify whether objects show significant overlap and containment with respect to the articulated part mask at the time of opening or the closed object mask. In case of significant agreement with the articulated part, we initialize an \texttt{ARTICULATED} edge $(i,c)$ between parent and child. Conversely, we initialize a \texttt{STATIC} edge when the child object covers the closed object mask.

\section{Experimental Evaluation}
\vspace{-0.2em}
We evaluate \ours{} on several different tasks ranging from articulation estimation to 3D part segmentation and contained object discovery. We conduct experiments on \textit{Arti4D}~\cite{arti25werby} as well as its labeled counterpart \textit{Arti4D-Semantic}, which is introduced in the following, and an articulated object split of the large-scale DROID dataset~\cite{khazatsky2024droid}. We provide metrics and further results on interaction segmentation, exo/robot-centric splits, and mobile manipulation in the Appendix Secs.~\ref{sec:approach_supp},~\ref{sec:supp_experiments}.

\noindent{\textbf{Arti4D-Semantic}:}
As part of this work, we present the \textit{\ourdataset{}} benchmark dataset, which, to the best of our knowledge, is the first hierarchical 3D scene graph dataset tailored towards scene-level articulated object understanding, connecting the previously isolated research areas of articulated object estimation and semantic 3D mapping. It extends \textit{Arti4D}~\cite{arti25werby} with interaction segments, labeled articulated 3D parts with semantic categories, and labels of objects contained within articulated objects as well as the relation to their respective parent object. Beyond the ego-centric sequences of \textit{Arti4D}, \textit{\ourdataset{}} adds exo-centric and robot-centric demonstrations via a teleoperated mobile manipulator. Overall, the dataset contains 62 RGB-D sequences across 5 scenes depicting human interactions with drawers, cabinets, doors, and sliding cabinets, providing instance and semantic labels coupled with axes of motion, see Fig.~\ref{fig:qualitative} and Appendix Sec.~\ref{sec:supp_arti4d_semantic}.

\noindent{\textbf{Baselines \& Implementation Details}:}
We compare against ArtiPoint~\cite{arti25werby}, which relies on factor graph estimation of pose trajectories using the estimator of Buchanan~\etal\cite{buchanan2024online} or Sturm~\etal\cite{sturm2011probabilistic}. ArtiPoint is already tuned towards Arti4D. In addition, we compare against ArtGS~\cite{liu2025building}, which employs Gaussian splatting-based reconstruction to identify axes. As it is intended to run on quasi-static scenes, we adapt it by masking hands using YOLOv9~\cite{wang2024yolov9} similar to ArtiPoint~\cite{arti25werby}, and provide it with three frames before, during, and after interaction, respectively. We also compare against Pandora~\cite{yu25pandora}, which constructs articulated scene graphs from egocentric vision by initializing flood fills near hands towards static, non-articulated objects in order to segment the underlying articulated object. To adapt Pandora to Arti4D, we provide it with real-world depth, ground-truth closed states given the interaction window ground truth and manually provide states of maximum opening. 

For comparability, all methods in Tab.~\ref{tab:articulation_estimation},~\ref{tab:articulated_parts} use ground truth interaction segments.  
We run TapNext++~\cite{tapnextpp2026} forward and backward along each interaction using a query grid resolution of 64 x 64 and require minimum point track spans of $\qty{0.15}{\meter}$.
We use $\eta^{*}=0.995$ and embed the twist estimator in a RANSAC loop with 100 iterations over 30\% splits. The approach requires 24 GB of VRAM. We provide additional experiments and settings in the Appendix. 
\renewcommand{\arraystretch}{0.95}
\begin{table*}[t]
\color{black}
\centering
\scriptsize
\caption{\centering{Articulated object estimation results on Arti4D. We report axis-angle errors $\theta_{err}$, pos. errors $d_{L2}$ of rev. joints, type accuracy, and mode recognition quality. We conduct the ablation study on Arti4D, see top 6 rows.}}
\setlength\tabcolsep{2.0pt}
\begin{threeparttable}
\begin{tabular}{l|cccccc|cccc}
      \toprule
          & \multicolumn{6}{c|}{\textbf{Arti4D}} & \multicolumn{4}{c}{\textbf{DROID}}  \\
            \midrule
          \multirow{2}{*}{Method} & Prismatic & \multicolumn{2}{c}{Revolute} & \multicolumn{2}{c}{Joint Type [\%]} & Mode & Prismatic & \multicolumn{2}{c}{Revolute} & Type\\
 & $\theta_{err}$[$^\circ$] & $\theta_{err}$[$^\circ$] & $d_{L2}$[m] & Acc. & Recall Pris./Rev. & Acc. [\%] & $\theta_{err}$[$^\circ$] & $\theta_{err}$[$^\circ$] & $d_{L2}$[m] & Acc.[\%] \\
        \midrule
        Ours w/ CoTracker3 & 13.190 & 22.982 & 0.091 & 88.4 & 91.7\,/\,86.6 & \multirow{6}{*}{98.3} & \multicolumn{4}{c}{\cellcolor{gray!30}}  \\ 
        Ours w/ 32x32 queries & 14.982 & 18.424 & 0.110 & 86.1 & 82.5\,/\,92.9 & & \multicolumn{4}{c}{\cellcolor{gray!30}}\\
        Ours w/ 96x96 queries & 09.263 & 16.704 & 0.066 & 93.8 & 91.5\,/\,100.0 & & \multicolumn{4}{c}{\cellcolor{gray!30}}\\
        Ours w/o $\xi$-regul. + w/o RANSAC & 34.089 & 23.175 & 0.103 & 90.9 & 87.7\,/\,97.0 & & \multicolumn{4}{c}{\cellcolor{gray!30}} \\ 
        Ours w/o $\xi$-regular. + RANSAC & 21.427  & 17.119 & 0.103 & 90.2 & 87.4\,/\,95.2 &  & \multicolumn{4}{c}{\cellcolor{gray!30}}  \\ 
        Ours w/o RANSAC & 11.205 & 20.382 & 0.113 & 90.3 & 87.6\,/\,95.6 & & \multicolumn{4}{c}{\cellcolor{gray!30}} \\ 
        \greyrule
         ArtGS~\cite{liu2025building} & 52.070 & 62.638 & 0.301 & 61.9 & 100.0\,/\,00.0 & -  & - & - & - & -\\
         Pandora~\cite{yu25pandora} & 46.814 & 50.620 & \underline{0.195} & 55.9 & 77.2\,/\,44.4 & - & - & - & - & -\\
         ArtiPoint~\cite{arti25werby} w/~\cite{sturm2011probabilistic} & \underline{17.754} & 28.467 & 0.313 & \underline{87.2} & 85.4\,/\,90.5 & - & 28.47 & 27.33 & 0.333 & \underline{62.2} \\ 
         ArtiPoint w/o prior~\cite{arti25werby} & 22.953 & 27.057 & 0.269 & 77.6 & 71.3\,/\,89.9 & - & \underline{34.68} & 26.10 & 0.284 & 61.8 \\ 
         ArtiPoint~\cite{arti25werby} & 23.272 & \underline{26.358} & 0.248 & 77.6 & 70.9\,/\,90.6 & - & 35.88 & \underline{25.43} & \underline{0.278} & 61.1 \\ 
         MoMa-SG (ours) & \textbf{10.374} & \textbf{14.800} & \textbf{0.087} & \textbf{90.9} & 87.7\,/\,95.8 & 98.3 & \textbf{07.15} & \textbf{16.91} & \textbf{0.115} & \textbf{89.5}\\ 
      \bottomrule
\end{tabular}
\footnotesize
\end{threeparttable}
\label{tab:articulation_estimation}
\color{black}
\vspace{-2.5em}
\end{table*}
\renewcommand{\arraystretch}{1.0
}

\noindent{\textbf{Interaction Segmentation}:}
\label{sec:eval_interaction}
We evaluate \ours{} on segmenting interactions temporally in Tab.~\ref{tab:interaction_segmentation} against Pandora, which casts spheres around hands and measures overlap with a static scene reconstruction, a simple hidden Markov model (HMM) with our two observation signals, and ArtiPoint. \ours{} demonstrates the highest 1D-IoU, segment recall, and segment-based IoU, whereas the HMM shows significant precision at low recalls. Furthermore, we outperform GPT-5-mini as a naive baseline, which operates on sampled 3~Hz video. We deduce that introducing the warp prior (Sec.~\ref{sec:interaction_disc}) leads to a significant performance increase compared to ArtiPoint.

\label{sec:eval_articulation_estimation}
\vspace{-0.2em}
\noindent{\textbf{Articulation Estimation}:}
As shown in Tab.~\ref{tab:articulation_estimation}, \ours{} outperforms all baselines in terms of prismatic and revolute axis errors while also exhibiting increased type accuracy. We find that ArtiPoint, when not using ground truth axis types at twist parsing, yields inferior results compared to \ours{}. We further observe that Pandora suffers from hand and object occlusions, which result in incomplete meshes and severely complicate articulation estimation. We provide further results on robustness and statistical significance of the proposed estimator in Appendix Sec.~\ref{sec:articulation_eval_supp}.
Ablating the benefit of the proposed twist regularization, we observe higher axis and position errors for revolute joints without regularization. Furthermore, we find higher axis and type accuracy when increasing the point query resolution of TapNext++, which however also increases memory consumption. Furthermore, we find that mode understanding accuracy drops considerably when employing GPT-5-nano instead of GPT-5-mini (69.3\%\,/\,98.3\%). In additional ablations, we observe significantly higher errors and lower type accuracy when relying on CoTracker3~\cite{karaev2024cotracker3} utilizing sparse queries based on GFTT~\cite{shi1994good} keypoints, or when not employing a RANSAC-based estimator. 
We visualize qualitative results in \cref{fig:qualitative}. 
Subsequently, we compare \ours{} with ArtiPoint, as the best-performing baseline, on an articulated object split of DROID in Tab.~\ref{tab:articulation_estimation} and observe similar boosts under our method. As an interaction prior, we obtain robot masks via SAM3~\cite{carion2025sam3segmentconcepts}.

\begin{figure*}[h!]
\centering
\footnotesize
\setlength{\tabcolsep}{0.01cm}
\includegraphics[trim=0 0cm 0 0cm, clip, width=\linewidth]{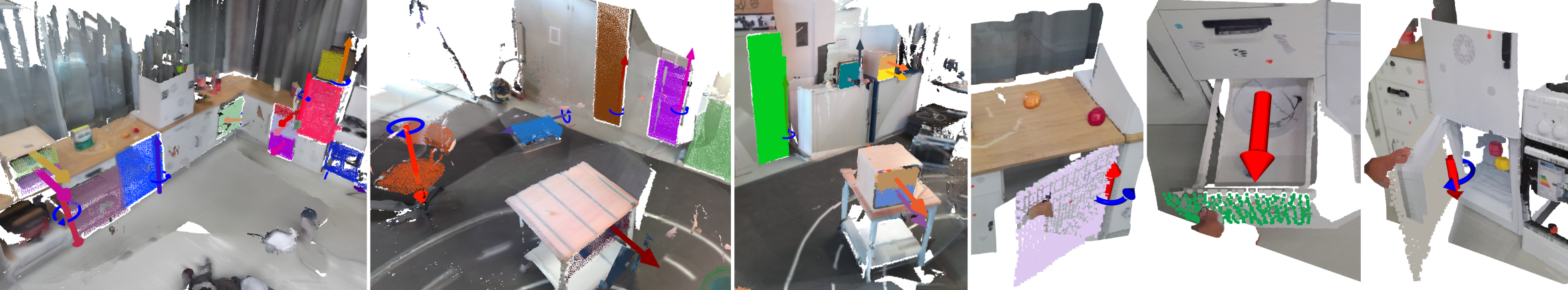}
\caption{Qualitative results of \ours{} on \ourdataset{}: Estimated axis positions and corresponding object masks. As demonstrated, we observe minimal errors for a large number of prismatic objects and small errors on revolute objects. Revolute axes are depicted with a blue torus, prismatic axes without.} 
\label{fig:qualitative}
\vspace*{-0.8em}
\end{figure*}

\begin{figure}[htbp]
    \centering
    \includegraphics[trim=0cm 10cm 5cm 3cm, clip,width=0.49\linewidth,height=5cm]{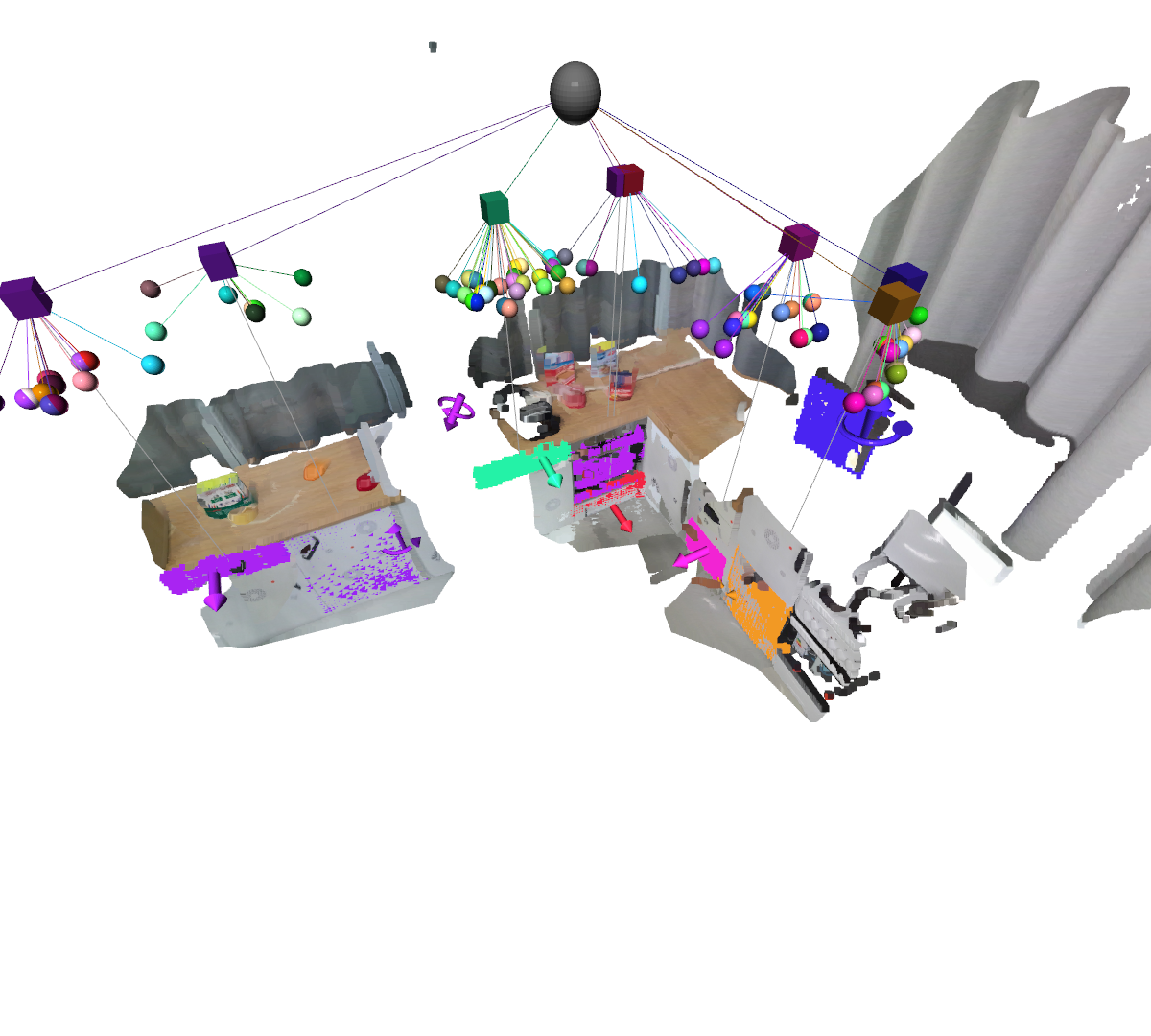}
    \includegraphics[trim=5cm 10cm 0cm 0cm, clip, width=0.49\linewidth,height=5cm]{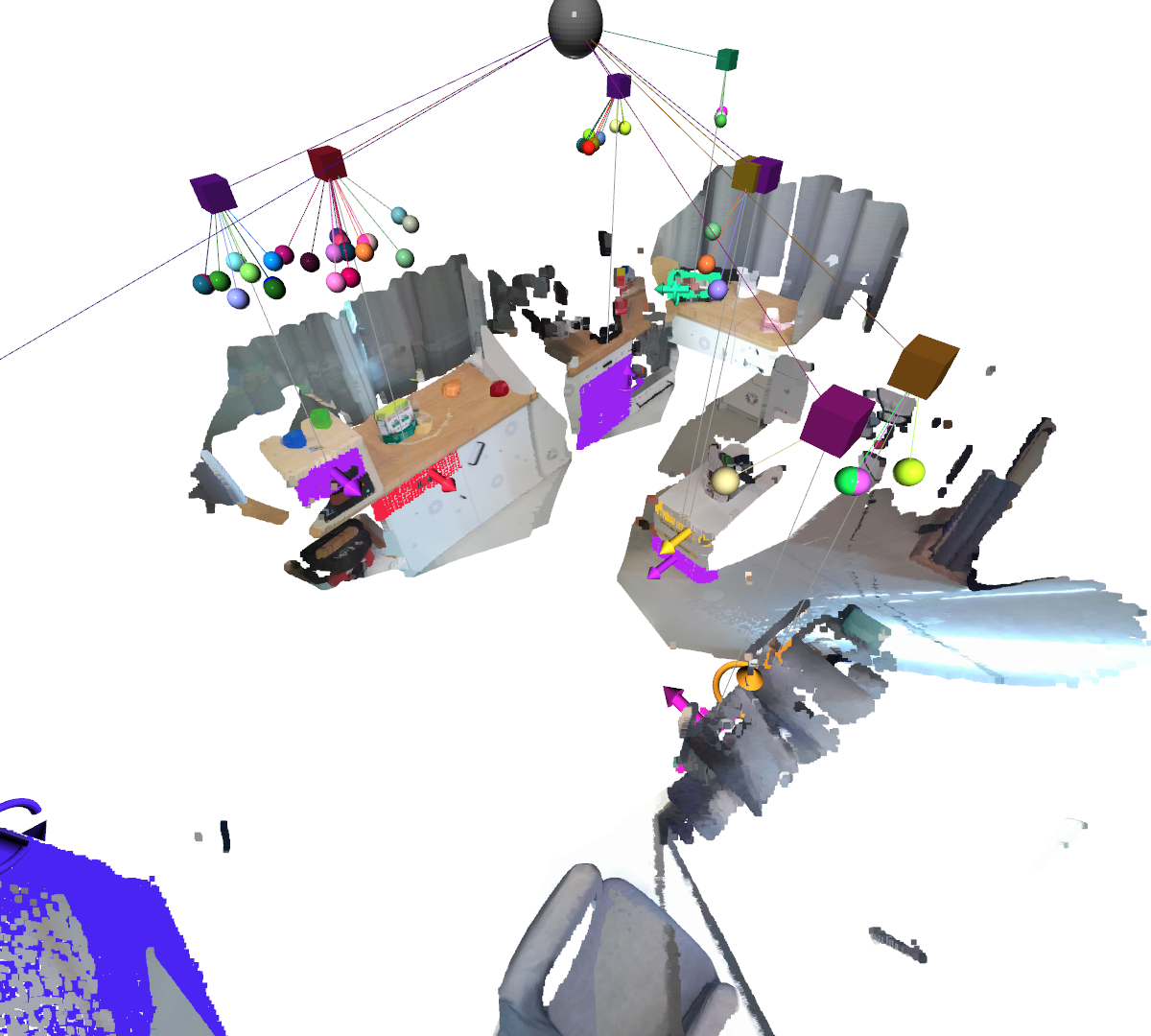}
    \caption{Articulated 3D scene graphs: The top node represents the root node, the cube-shaped nodes represent articulated objects (parents), while the smaller, lower nodes represent discovered child nodes with respect to their articulated parents.}
    \label{fig:articulated_3dsg}
    \vspace{-1.2em}
\end{figure}

\renewcommand{\arraystretch}{0.95}
\begin{table}[t]
\color{black}
\centering
\scriptsize
\caption{\footnotesize{Articulated 3D part segmentation (left) and contained object discovery (right) evaluation on Arti4D-Semantic: The upper four rows refer to \textit{free} articul.-indep. objects $\mathcal{O}$, the lower section contains objects $\mathcal{O}^{\mathcal{A}}$ matched against articulations. The IoU denotes mutual overlap under a linear assignment. The recall indicates the GT overlap at various IoUs. For object discovery (right), we only evaluate objects paired with GT objects on the, the IoU denotes the highest-achieved overlap between a predicted child node and all GT children. The relation accuracy measures correctly identified parent-child edges. Evaluation on RH201 split of Arti4D-Semantic.
}}
\setlength\tabcolsep{6.3pt}
\begin{threeparttable}
\begin{tabular}{l|ccccc|ccc}
      \toprule
         & \multicolumn{5}{c|}{3D Part Mapping} & \multicolumn{3}{c}{Contained Object Discovery} \\
         \midrule
        \multirow{2}{*}{Method} & Object & \multirow{2}{*}{IoU} & & Recall & & \multirow{2}{*}{IoU} & Recall~[\%] & Relation Acc.~[\%] \\
        & Set  & & \textit{@0.25} & \textit{@0.50} & \textit{@0.75} & & @IoU=0.25 & \texttt{STATIC} / \texttt{ARTIC.} \\
        \midrule
         OpenMask3D~\cite{takmaz2023openmask3d} & \multirow{4}{*}{$\mathcal{O}$} & 0.082 & 0.087 & 0.019 & 0.000 & \multicolumn{3}{c}{\cellcolor{gray!30}} \\
         ConceptGraphs~\cite{gu2024conceptgraphs} & & 0.438 & 0.332 & 0.127 & 0.018 & \multicolumn{3}{c}{\cellcolor{gray!30}} \\
         Ours w/o bound.-resp.~merging &  & 0.481 & 0.807 & 0.534 & 0.086 & \multicolumn{3}{c}{\cellcolor{gray!30}} \\
         MoMa-SG (ours) & & \textbf{0.533} & \textbf{0.824} & \textbf{0.646} & \textbf{0.163} & \multicolumn{3}{c}{\cellcolor{gray!30}} \\ 
         \greyrule
         Pandora~\cite{yu25pandora} & \multirow{5}{*}{$\mathcal{O}^{\mathcal{A}}$} & 0.065 & 0.012 & 0.000 & 0.000 & 00.02 & 00.35 & 19.77 \\
         Ours w/ CoTracker3 &  & 0.221 & 0.370 & 0.218 & 0.027 &
         09.14 & 16.67 & 59.23 \\ 
         Ours w/ Mobile-SAM & &  0.417 & 0.672 & 0.453 & 0.112 & 09.31 & 35.66 & 67.08\\ 
         Ours w/ greedy matching &  & \underline{0.423} & \underline{0.703} & \underline{0.453} & \underline{0.110} & \underline{09.50} & \underline{37.43} & \underline{67.19} \\ 
         MoMa-SG (ours) & &\textbf{0.434} & \textbf{0.720} & \textbf{0.471} & \textbf{0.111} &
         \textbf{11.14} & \textbf{42.60} & \textbf{90.56} \\ 
      \bottomrule
\end{tabular}
\footnotesize

\end{threeparttable}
\label{tab:articulated_parts}
\color{black}
\vspace{-0.4em}
\end{table}
\renewcommand{\arraystretch}{1.0}

\label{sec:eval_objects}

\noindent{\textbf{Object Understanding}:}
Precise articulation estimation is paramount for matching objects against articulations and when identifying contained objects. Our object understanding experiments in Tab.~\ref{tab:articulated_parts} are two-fold. We investigate how well articulated 3D parts are segmented, and whether those segments allow us to identify contained objects within various articulated objects. We evaluate the segmentation performance in two distinct settings, being \textit{free} objects $\mathcal{O}$ and articulated objects \textit{matched} against articulations $\mathcal{O}^{\mathcal{A}}$. 
Pandora does not provide \textit{free} objects as it directly segments articulated parts during articulation estimation, except for contained objects.

Overall, we find that \ours{} significantly outperforms Pandora as the latter struggles to identify complete object meshes. We further find drastically decreased performance when using CoTracker3 as a point tracker as it yields less uniform and dense part coverage throughout the tracking stage, which ultimately complicates the subsequent object-articulation association. We further demonstrate that employing part masks stemming from Mobile-SAM instead of Semantic-SAM leads to a slight performance decrease ($\sim$2\,\%pt).
In the free mask setting, we observe that 3D object merging solely from geometric overlap outperforms ConceptGraphs~\cite{gu2024conceptgraphs}, which relies on semantic similarities for object merging, which frequently produces false merges, \eg, when two neighboring drawers are undersegmented. In addition, we demonstrate that merging objects based on whether their 2D masks intersect with the image boundaries significantly improves mapping. Furthermore, the proposed BIP for object-articulation matching further increases performance compared to greedy matching.
Regarding contained objects, we face a compounding error problem. Inferior results in articulation estimation and object segmentation are detrimental to the detection of child nodes. In general, \ours{} outperforms Pandora in both discovered child nodes and relation understanding as displayed in Tab.~\ref{tab:articulated_parts}. Nonetheless, class-agnostic child instance segmentation may lead to frequent errors in the granularity of detected objects. 
We depict two constructed articulated scene graphs in Fig.~\ref{fig:articulated_3dsg}.

\begin{figure*}
    \centering
    \includegraphics[width=1.0\linewidth]{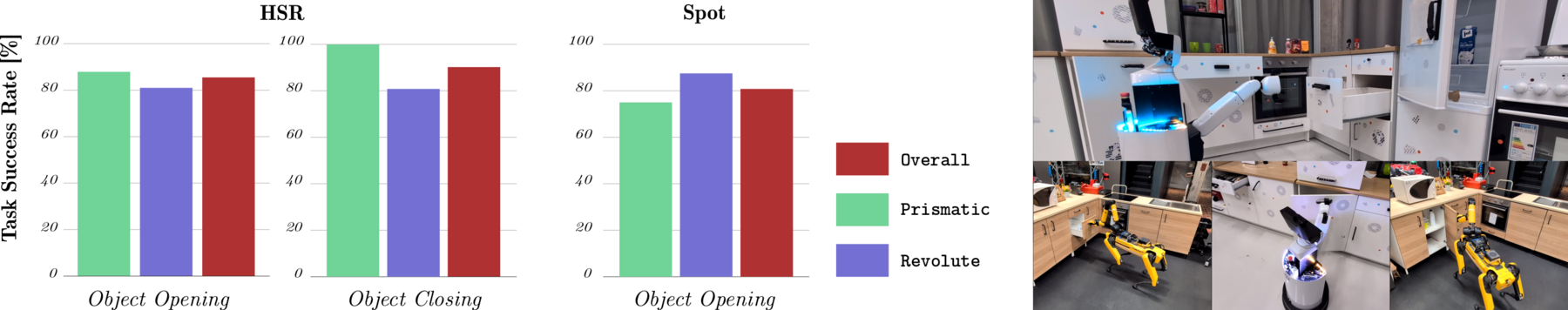}
    \caption{We portray our real-world mobile manipulation environments of the HSR as well as the Spot. We demonstrate that MoMa-SG allows for robust mobile manipulation even under various articulation states of objects. The sample size for each robot-task combination is 50.}
    \label{fig:real-world-moma}
    \vspace{-1em}
\end{figure*}

\vspace{-0.1em}
\noindent{\textbf{Real-World Mobile Manipulation}:}
To demonstrate the real-world utility of \ours{}, we conduct two experiments. As outlined in the Appendix Tab.~\ref{tab:real_world_articulation_state_estimation}, \ours{} allows to precisely estimate real-world articulation states of objects, enabling autonomous articulated object manipulation independent of the configuration state.

In addition, we demonstrate how \ours{} serves as an embodiment-agnostic representation for mobile manipulation of articulated objects. The experiment is conducted in two environments (\cref{fig:real-world-moma}), where we detect object handles online using SAM3~\cite{carion2025sam3segmentconcepts}, estimate grasps using their predominant axes, and match those against all objects using a Hungarian assignment. We sample end-effector trajectories given those grasps and the estimated twists. We evaluate two robotic embodiments, the HSR and a Boston Dynamics Spot, tasked with operating the articulated objects represented in the graph, and measure overall task success of opening and closing objects from random articulation states as reported in \cref{fig:real-world-moma}, where both robots achieve over 80$\%$ success across articulation types and environments. We are able to attribute the majority of failed actions to gripping failures. Furthermore, we provide details on the mobile manipulation approach and portray language-enabled long-horizon mobile manipulation in the Appendix Sec.~\ref{sec:moma_supp} and~\ref{sec:supp_experiments}.
\vspace{-0.4cm}

\section{Conclusion}
We presented \ours{} as a novel embodiment-agnostic method for distilling in-the-wild object articulations into articulated 3D scene graphs. This allows for robust mobile manipulation in dynamic human-centered scenes. In addition, we have presented the novel Arti4D-Semantic dataset as the first benchmark for evaluating articulated 3D scene graphs including object containment. In future work, we aim to address semi-static rearrangements.

\noindent{\textbf{Limitations}:}
Reducing reliance on accurate camera poses and depth sensing remains desirable. To date, scaled monocular depth still lacks the fidelity required for consistent articulation estimation in in-the-wild settings. We further identify track filtering and robust estimation under noise as the primary bottlenecks, which we aim to address through learned filters in future work. Additionally, efficient interaction and articulation recognition continue to be a limiting factor.

\acknowledgments{This work was funded by the BrainLinks-BrainTools center of the University of Freiburg, an academic grant from NVIDIA, and Toyota Motor Europe (TME) supported this project with an HSR robot. Hermann Blum is also with the Lamarr Institute for AI and ML. }

\bibliography{references}  
\clearpage
\appendix
\renewcommand{\baselinestretch}{1}
\setlength{\belowcaptionskip}{0pt}

\setcounter{section}{0}
\setcounter{equation}{0}
\setcounter{figure}{0}
\setcounter{table}{0}
\setcounter{page}{1}

\makeatletter

\renewcommand{\thesection}{A.\arabic{section}}
\renewcommand{\thesubsection}{A.\arabic{section}-\Alph{subsection}}
\renewcommand{\thetable}{A.\arabic{table}}
\renewcommand{\thefigure}{A.\arabic{figure}}

\begin{center}
\vspace{-5ex}
{\Large\bf 
Articulated 3D Scene Graphs \\\vspace{0.5ex} for Open-World Mobile Manipulation
} \\
\vspace{2.5ex}

\large{- Appendix -}\\
\vspace{0.6cm}

\normalsize{Martin Büchner\textsuperscript{\footnotesize 1}\quad
Adrian Röfer\textsuperscript{\footnotesize 1}\quad
Tim Engelbracht\textsuperscript{\footnotesize 2}\quad
Tim Welschehold\textsuperscript{\footnotesize 1}\quad
Zuria Bauer\textsuperscript{\footnotesize 2}\quad \\[0.2cm]
Hermann Blum\textsuperscript{\footnotesize 2,3}\quad
Marc Pollefeys\textsuperscript{\footnotesize 2}\quad
Abhinav Valada\textsuperscript{\footnotesize 1}}\\
\vspace{0.3cm}
\normalsize{\textsuperscript{\footnotesize 1}University of Freiburg \quad\quad {\textsuperscript{\footnotesize 2}ETH Zürich \quad\quad \textsuperscript{\footnotesize 3}University of Bonn}}

\end{center}

\begin{figure}[htbp]
    \centering
    \includegraphics[trim=0cm 0cm 0cm 0cm, clip, width=\linewidth]{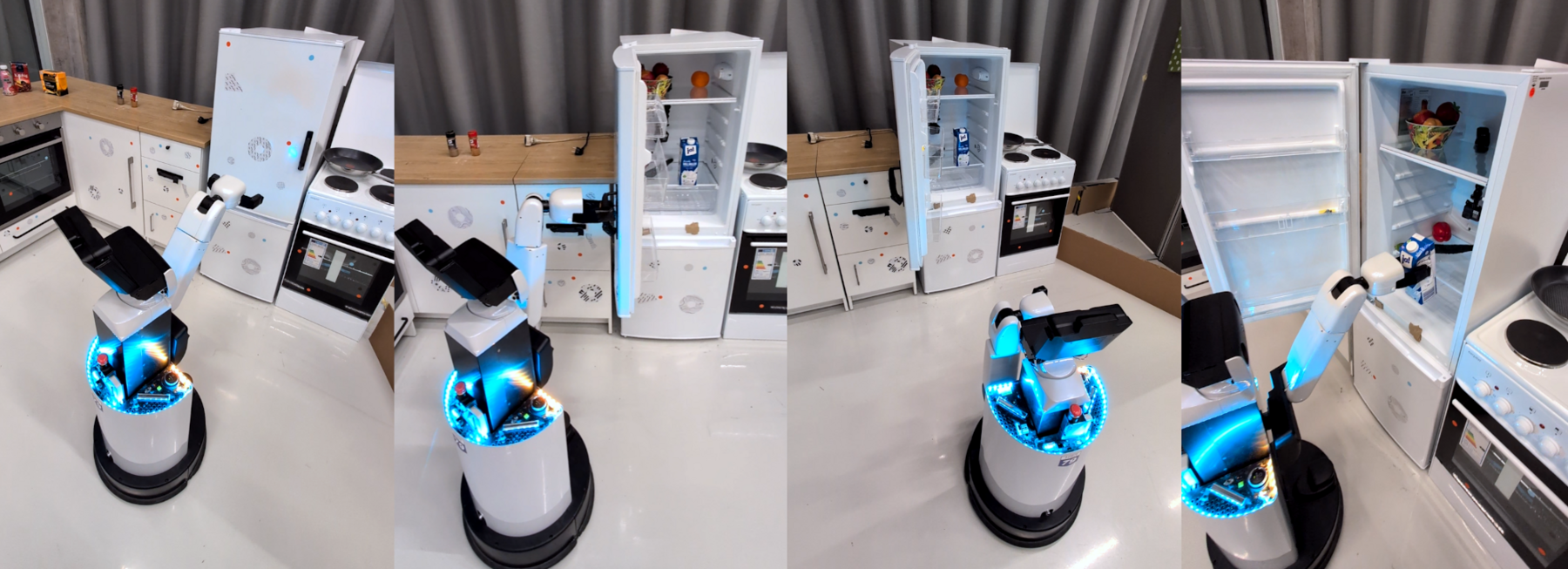}
    \captionof{figure}{%
    We demonstrate the ability of \ours{} to map natural language commands (e.g., “get the milk from the fridge”) to an appropriate sequence of high-level actions, such as opening the fridge, inspecting its contents, and retrieving the target object. All of this is done from open-vocabulary language features and object articulation models estimated from ego-centric video in a one-shot fashion while not relying on any fiducial markers.
    }
    \label{fig:retrieve}
\end{figure}

Throughout this supplementary material, we extend the approach of \ours{} in \cref{sec:approach_supp}, provide further experimental results in \cref{sec:supp_experiments}, and add details on the proposed \ourdataset{} benchmark dataset in \cref{sec:supp_arti4d_semantic}. We visualize a global articulated 3D scene graph output of \ours{} in \cref{fig:articulated_3dsg}.

In particular, we explain the following approach aspects: Interaction segment discovery in \cref{sec:interaction_disc_supp}, our regularized twist estimator including comparisons against privileged estimators assuming joint type knowledge in \cref{sec:supp_twist}, the open-vocabulary 3D part mapping scheme in \cref{sec:part_mapping_supp}, and the proposed object-to-articulation matching scheme in \cref{sec:supp_obj_articulation_matching}. Finally, we provide details on the real-world mobile manipulation stack in \cref{sec:moma_supp}, which includes global localization in ego-centric maps, online grasp generation, and end-effector trajectory sampling for performing object opening and closing actions in the real world as well as mobile manipulation from natural language inputs.

In \cref{sec:supp_experiments}, we explain the utilized metrics (\cref{sec:metrics_supp}), compare the interaction discovery approach of \ours{} in \cref{sec:interaction_segmentation_eval_supp}, and evaluate the articulation estimation and articulated 3D part mapping in \cref{sec:articulation_eval_supp} and \cref{sec:part_mapping_eval_supp} while relying on predicted interaction segments instead of ground truth segments. Further, we present real-world articulation state estimation results in \cref{sec:real_world_long_horizon_moma}, provide a robustness study of \ours{} in \cref{sec:eval_robustness_study_supp}, and report runtime statistics in \cref{sec:runtime_supp}.

\section{Approach Details on \ours{}}
\label{sec:approach_supp}

\subsection{Interaction Discovery}
\label{sec:interaction_disc_supp}
In the following, we outline our interaction discovery approach briefly introduced in \cref{sec:interaction_disc}. Quantitative results on interaction segmentation are provided in \cref{sec:interaction_segmentation_eval_supp}. Further, we provide overall results of \ours{} using predicted interaction segments in \cref{tab:articulation_pred_seg} and \cref{tab:scene_results_part_seg_pred_seg}.

\noindent\textbf{Interaction Prior and Depth Disparity:}
For each RGB frame $\mathbf{I}_t$, we extract an interaction mask $\mathbf{H}_t$ using a pre-trained YOLOv11~\cite{khanam2024yolov11} model. We combine this interaction signal with a depth disparity measure that measures whether the scene stays static between depth map $\mathbf{D}_t$ and a previous depth map $\mathbf{D}_{t-s_H}$, where $s_H$ denotes a fixed stride. For this, we warp $\mathbf{D}_{t-s_H}$ into the camera view of $\mathbf{D}_t$, obtaining $\mathbf{\tilde{D}}_{t \leftarrow s}$. Given the camera poses at the respective timestamps $\mathbf{T}_{t-s_H}$ and $\mathbf{T}_{t}$ and the projection function from point cloud to depth map $\mathbf{D} = \textrm{proj}(\mathbf{p})$, we obtain a depth disparity $\Delta_t$ (\cref{fig:overview}):
\begin{align}
\Delta_t &=
\left|
\mathbf{D}_t - \mathbf{\tilde{D}}_{t \leftarrow s_H}
\right|\\
\mathbf{\tilde{D}}_{t \leftarrow s_H} &= \textrm{proj}\left( \mathbf{T}_{t}^{-1} \mathbf{T}_{t-s_H} \textrm{proj}^{-1} (\mathbf{D}_{t-s_H})   \right)
\end{align}

\noindent\textbf{Probabilistic Interaction Segmentation:}
In the following, we combine the depth disparity $\Delta_t$ and the interaction mask $\mathbf{H}_t$. We threshold $\Delta_t > \tau$ to obtain a binary mask and compute the pixel-wise sum and normalization of both masks as $h_t$ and $\delta_t$. We further apply median filtering over a time horizon $\kappa$. 

To fuse the two signals, we define the interaction score $\prob{s}$ as the probability of $t$ being part of an interaction segment. We then interpret the normalized scores $h_t, \delta_t \in \lbrack 0, 1\rbrack$ as probabilities: $\delta_t = \prob{d=\top}$ represents the probability of having dynamic object motion in the scene, and likewise, $h_t =\prob{h=\top}$ measures the probability of interacting with the scene. Assuming that $\prob{d}$ and $\prob{h}$ are independent, we obtain the following conditional probability model:
\begin{equation}
\begin{aligned}
    \prob{s} &= \sum_{d,h\in\{\top, \bot\}} \prob{s,d,h} = \sum_{d,h} \probc{s}{d,h} \prob{d,h} = \sum_{d,h} \probc{s}{d,h} \prob{d} \prob{h}
\end{aligned}
\end{equation}
The values for $\probc{s}{d,h}$ are tuned on held-out data and describe the conditional probability for a frame belonging to an interaction segment, given the observation of the interaction mask and the dynamics in the scene. The resulting interaction score is depicted in \cref{fig:overview}. 
We parse the segments by identifying consecutive non-zero intervals of length $T$ and disregard segment lengths below $T_{min}$ and above $T_{max}$.

\subsection{Regularized Twist Estimation:}
\label{sec:supp_twist}
\begin{figure*}[ht]
\centering
\footnotesize
\includegraphics[trim=0cm 0cm 0cm 1cm, clip, width=\linewidth]{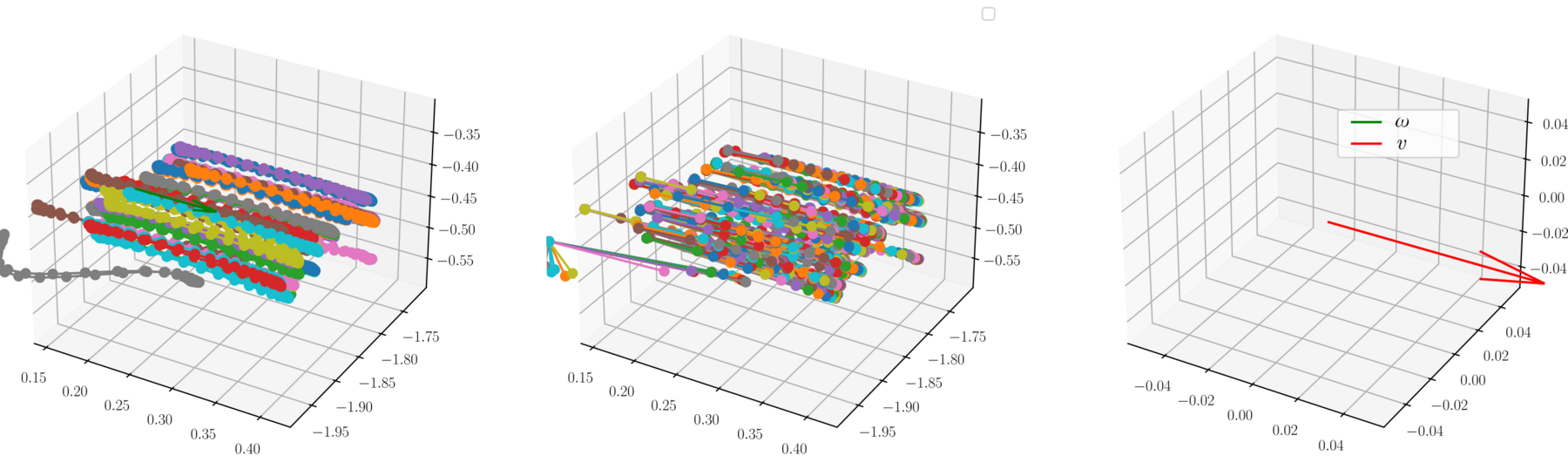} 
\caption{Estimated twist components given an observed prismatic motion. We depict observed 3D point trajectories (left), sampled point pairs used for articulation estimation (middle), and the estimated articulation model (right).} 
\label{fig:twist_prismatic}
\end{figure*}

\begin{figure*}[ht]
\centering
\footnotesize
\includegraphics[trim=0cm 0cm 0cm 0cm, clip, width=\linewidth]{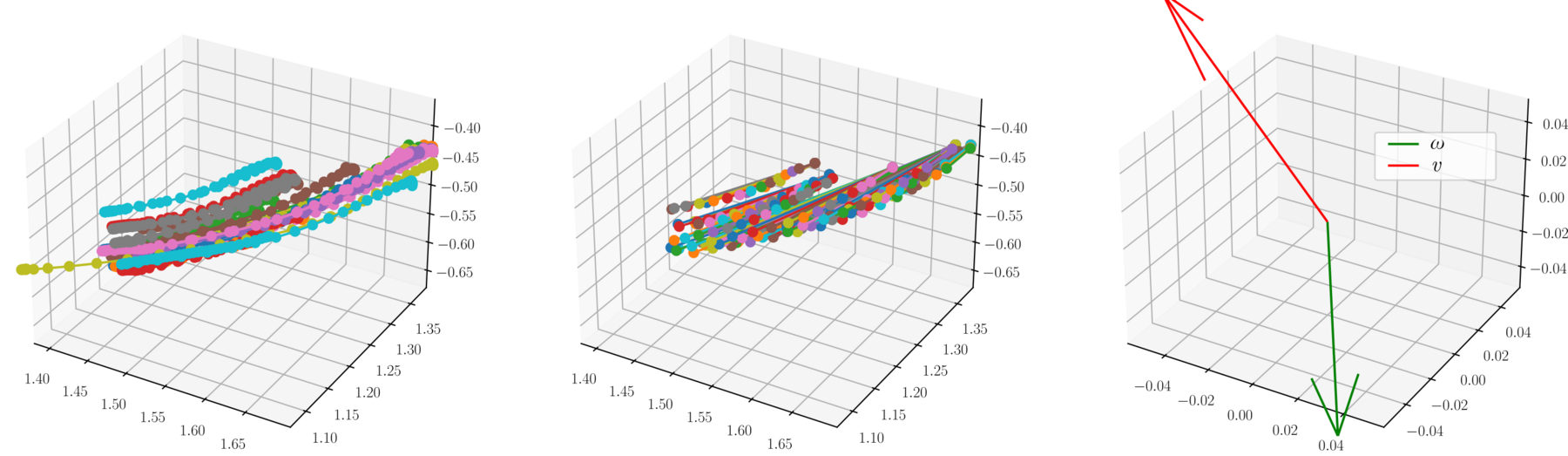} 
\caption{Estimated twist components given an observed revolute motion of small angular range using our proposed regularized twist estimation scheme. We visualize point tracks (left), sampled point pairs (middle), and rotational and translation twist components (right).}
\label{fig:twist_revolute}
\end{figure*}

We expand upon the proposed geometric prior for regularized twist estimation introduced in \cref{sec:articulation_estimation} by depicting two estimation outputs. As depicted in \cref{fig:twist_prismatic}, we robustly estimate a \textit{prismatic} twist that is governed solely by its translation component $v$ while $\|\omega\|$ approximates zero, even under partially noisy trajectories. Similarly, we display the estimation result on a revolute joint in \cref{fig:twist_revolute}, which is prone to misclassification as a prismatic joint as its observed angular range is rather small (approx. $\qty{20}{^{\circ}}$). Nonetheless, due to the geometric prior regularization term, we observe orthogonal twist components, effectively rendering the articulation as the revolute case of twists as described in \cref{sec:articulation_estimation}.
 
\begin{figure*}
\centering
\footnotesize
\includegraphics[trim=0cm 0cm 0cm 0cm, clip, width=\linewidth]{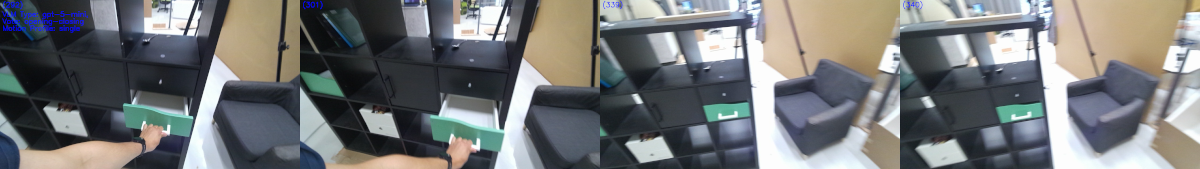} 
\caption{Articulation mode prompt consisting of four concatenated keyframes along the interaction.} 
\label{fig:mode_prompt}
\end{figure*}

Furthermore, in \cref{fig:mode_prompt}, we display the mode understanding prompt to \texttt{GPT-5-mini}, which is turned into the articulation mode estimate $M_i$. We sample four keyframes along the interaction and concatenate them. As a textual prompt, we provide: \textit{"Is the following series of images a consecutive 'opening-closing', 'closing-opening', or just 'opening', or 'closing'. Please answer with one of these four options."}.

\begin{figure}[htbp]
\centering
\footnotesize
\includegraphics[trim=0 0cm 0 0cm, clip, width=0.325\linewidth]{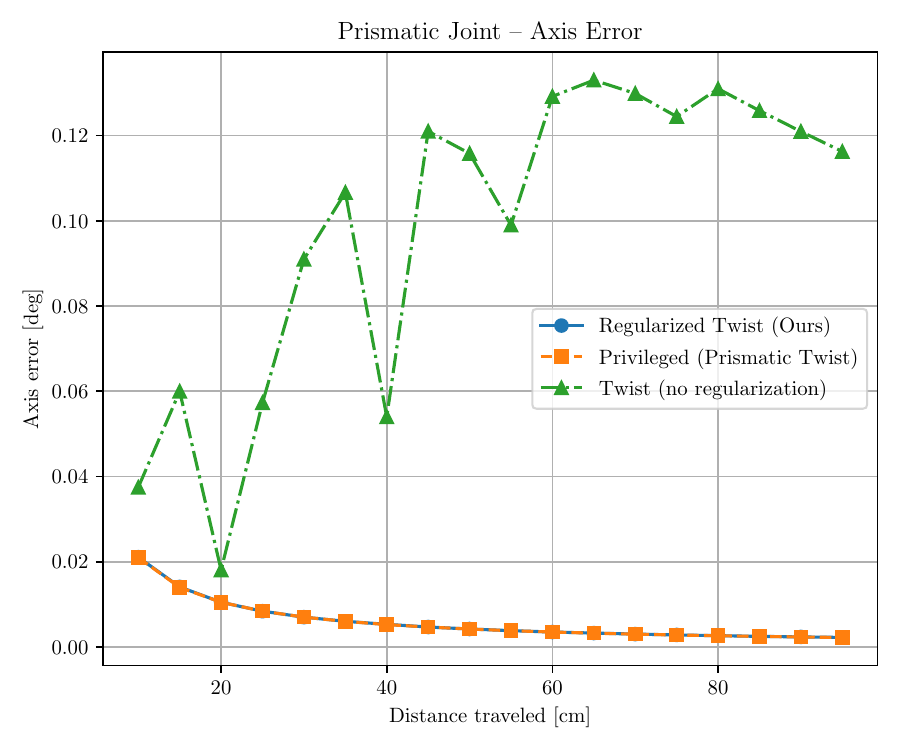}
\includegraphics[trim=0 0cm 0 0cm, clip, width=0.325\linewidth]{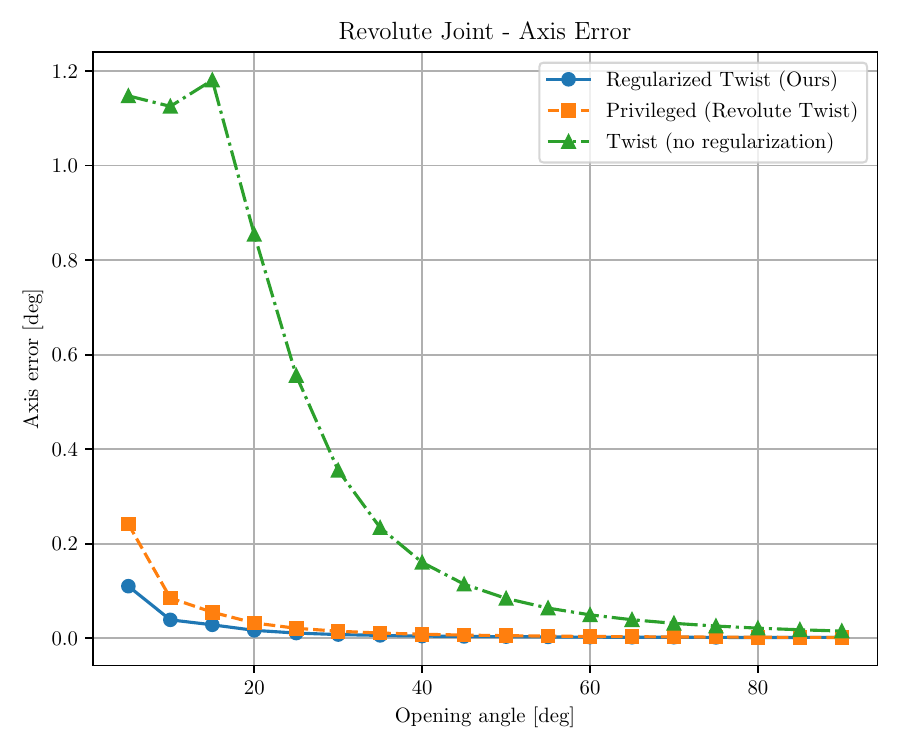}
\includegraphics[trim=0 0cm 0 0cm, clip, width=0.325\linewidth]{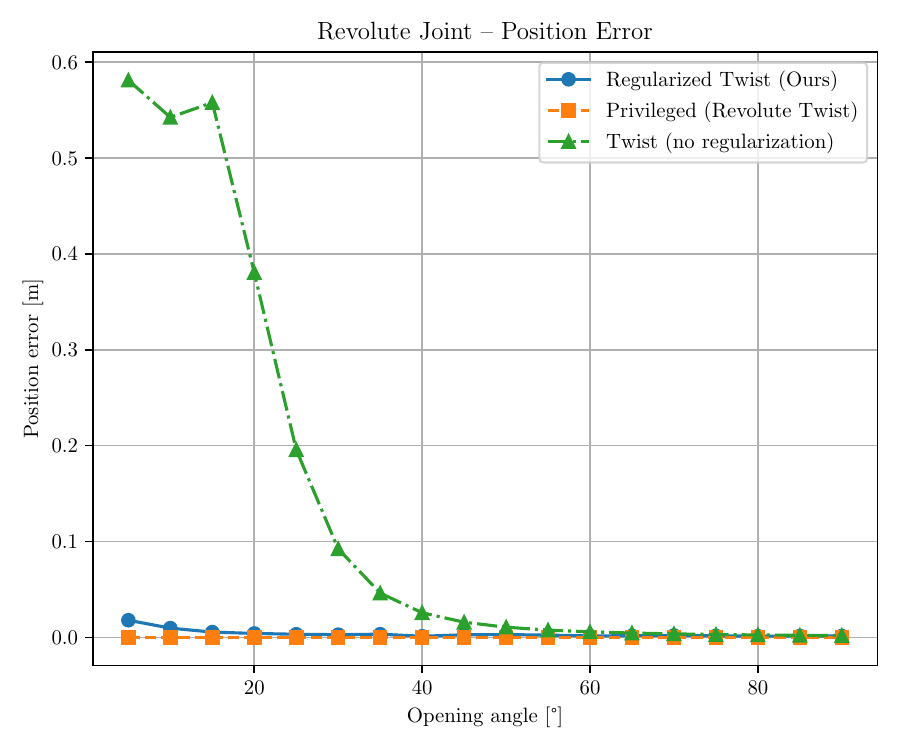}
\caption{Axis and position prediction errors of a regularized twist estimator (ours), under a privileged purely prismatic/revolute model (prismatic/revolute twist), and a non-regularized twist at various articulation configurations sampled synthetically and subject to point-level Gaussian noise.} 
\label{fig:axis_errors_twist}
\end{figure}

Furthermore, we compare on synthetic data using point-level Gaussian noise to which degree our estimator falls short with respect to privileged estimators assuming joint type knowledge, \ie, a purely prismatic or revolute joint estimator. As depicted in \cref{fig:axis_errors_twist}, we find that our regularized estimator performs on par with a privileged estimator, ultimately validating the proposed cosine prior introduced in \cref{sec:articulation_estimation}. In addition to that, we compare the regularized twist estimator against a non-regularized counterpart, which similarly relies on point pairs but does not incorporate a geometric type prior, see \cref{fig:axis_errors_twist}, across both prismatic and revolute joints. Note that we do not apply any RANSAC-based robustification. We find that the non-regularized twist estimator is producing axis errors that are approximately 10 times higher than for its regularized counterpart. Similarly, we observe revolute translational errors that are one magnitude greater. We would like to point out that these experiments are conducted on synthetic data with stationary noise. On real-world data, we find a vastly different trajectory distribution that is governed by severe non-stationary point drift, incomplete tracks due to occlusions, and depth noise, especially at object boundaries, which renders the regularization even more crucial for robust estimation.

In \cref{sec:eval_robustness_study_supp} and \cref{fig:twist_eta_study}, we further evaluate the robustness of the regularized twist estimator under non-stationary noise and occlusions on real-world data while investigating the influence of the chosen hyperparameter $\eta$ governing type discrimination.

\subsection{Open-Vocabulary 3D Part Mapping:}
\label{sec:part_mapping_supp}

\begin{figure*}[ht]
\centering
\footnotesize
\includegraphics[trim=0cm 0cm 0cm 0cm, clip, width=\linewidth]{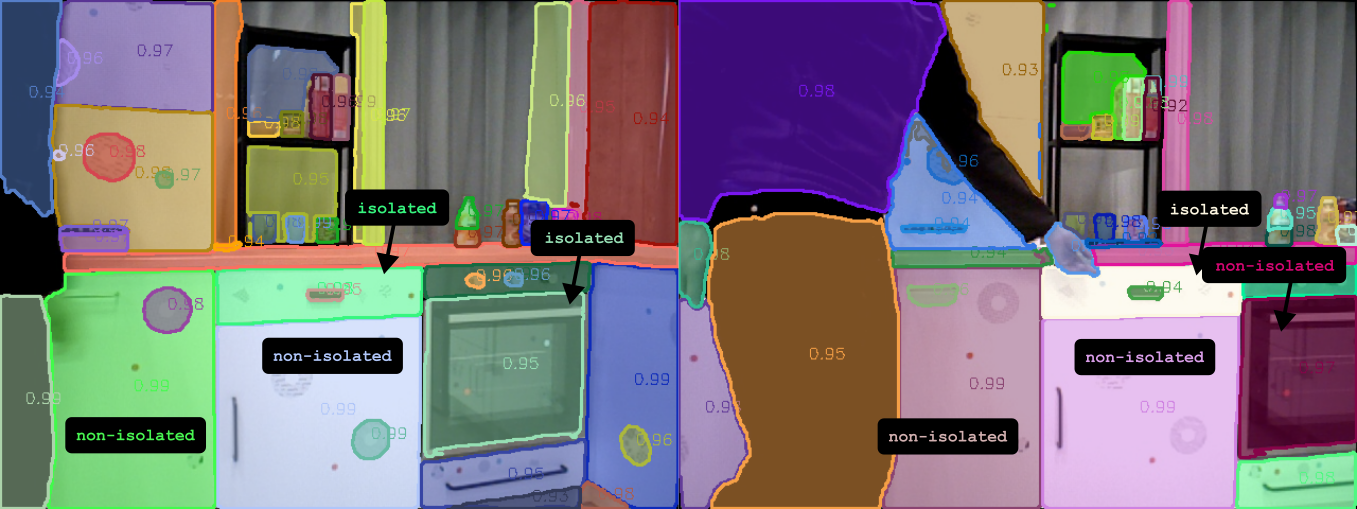}
\caption{We depict dense 2D segmentation masks across two distinct keyframes and denote masks considered as isolated and non-isolated, respectively. The scene stems from the HSR split of Arti4D-Semantic.} 
\label{fig:keyframes}
\end{figure*}

\begin{figure}[htbp]
\centering
\footnotesize
\includegraphics[trim=0cm 5cm 0cm 10cm, clip, width=\linewidth]{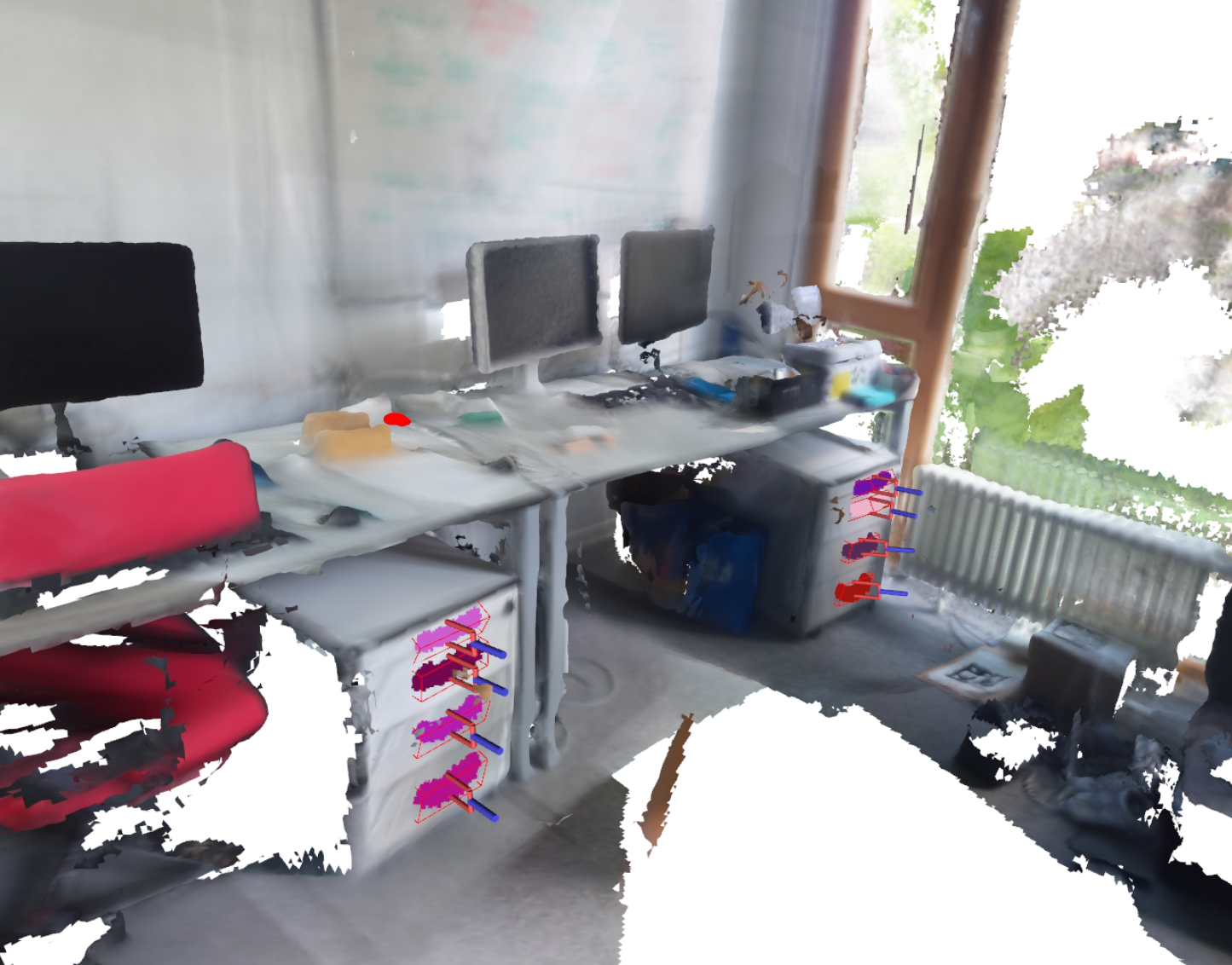}
\caption{Selection of grasps constructed using our handle detection pipeline employing SAM3~\cite{carion2025sam3segmentconcepts} and predominant axis estimation. The blue axis of each grasp constitutes its approach axis.} 
\label{fig:grasps_rr080}
\end{figure}

\noindent{\textbf{Keyframe Construction:}}
In order to construct $\mathcal{O}$, we perform 3D part mapping on the frames that are not part of interaction segments $\mathcal{S}$. Different from classical 3D scene graph construction methods~\cite{hovsg, gu2024conceptgraphs}, we employ a keyframe selection heuristic that does not utilize a fixed stride. Owing to the fact that interactive camera trajectories captured in the real world are highly affected by dynamic camera motion, as opposed to those stemming from simulation or typical indoor datasets~\cite{replica19arxiv, ramakrishnan2021hm3d}, we select keyframes based on small motion blur. To this end, we compute the image Laplacian across all considered frames and verify whether it exceeds a certain threshold. In general, we find that especially thin boundaries between, \eg, close-by drawers require small motion blur to be accurately segmented as depicted in \cref{fig:grasps_rr080}. In addition, we only gather keyframes if a certain distance $\gamma > \gamma^{*}$ is traveled and a rotational difference $\beta > \beta^{*}$ is exceeded in between two camera poses $\mathbf{T}_{k-1}$ and $\mathbf{T}_{k}$:
\begin{equation}
    \gamma = \left\| \mathbf{t}_{k} - \mathbf{t}_{\text{k-1}} \right\| \quad,\quad \beta =\operatorname{arccos}\left(\frac{\operatorname{tr}(\mathbf{R}_{k-1} \mathbf{R}_{k}^{-1})}{2}\right),
\end{equation}
where $\mathbf{R} \in SO(3)$ represents the rotational component and $\mathbf{t} \in \mathbb{R}^3$ the translational component. The above heuristic yields all frames taken as input to our 3D mapping stage. We visualize two exemplary keyframes in \cref{fig:keyframes}.

\begin{algorithm}[ht]
\SetAlgoLined
\DontPrintSemicolon
\SetKwFunction{FMerge}{MergeDetectionsToObjects}
\SetKwFunction{FMergeObj}{Merge}
\SetKwFunction{FOverlap}{overlap}
\SetKwInOut{Input}{Input}\SetKwInOut{Output}{Output}

\Input{Detections $\mathcal{D}$, Objects $\mathcal{O}$, Similarity matrix $\mathbf{S}$}
\Output{Updated free objects $\mathcal{O}$}

\For{$i \gets 0$ \KwTo $|\mathbf{S}|_{rows} - 1$}{
    $j \gets \text{argmax}(S_i)$\;
    $s_{ij} \gets \mathbf{S}({i,j})$\;
    $d \gets \mathcal{D}_i$\;
    $o \gets \mathcal{O}_j$\;
    
    \eIf{$o.\text{isolated}$}{
        \If{$s_{ij} \geq \epsilon_{isolated}$ \textbf{or} \FOverlap{$d, o$} $> \epsilon_{overlap}$}{
            $O_j \gets$ \FMergeObj{$o, d$}\;
        }
    }{
        \eIf{$d.\text{isolated}$}{
            \If{\FOverlap{$o, d$} $> \epsilon_{overlap}$}{
                $o \gets$ \FMergeObj{$o, d$}\;
                $o.\text{isolated} \gets \text{True}$\;
            }
        }{
            $o \gets$ \FMergeObj{$o, d$}\;
        }
    }
}
\Return{$\mathcal{O}$}\;
\caption{Incremental Merging}
\label{alg:merging}
\end{algorithm}

\noindent{\textbf{Part-Aware Mapping:}}
In this paragraph, we expand upon our 3D part mapping approach by providing additional details on the incremental merging scheme. To formalize this, we provide Alg. 1. During incremental mapping, we obtain a list of detections $\mathcal{D}$ per considered keyframe and aim to aggregate those as objects $\mathcal{O}$. At each mapping step, we compute the axis-aligned bounding-box IoU across all elements of $\mathcal{D}$ as well as $\mathcal{O}$, which yields the similarity matrix $\mathbf{S}$. We sort and threshold $\mathbf{S}$, and iteratively evaluate object-detection pairs, denoted as $(d,o)$ in the following. In general, we aim to find part segmentation masks with a certain level of confidence, which are attained by verifying masks via merging given significant point cloud-based overlap. We found that, in particular, part segmentation is prone to under- and over-segmentation, which frequently leads to erroneous merges when utilizing classical pipelines~\cite{gu2024conceptgraphs, hovsg}. While it is generally permissible to deal with wrong segmentation in typical search, navigation, or retrieval tasks, we find that mobile manipulation settings involving articulated objects require a higher level of fidelity. We aim to increase segmentation accuracy by evaluating whether the original 2D segmentation mask stemming from Semantic-SAM~\cite{li2023semantic} belonging to either $o$ or $d$ is isolated, \ie, it does not intersect image boundaries. We depict this differentiation for a number of objects on the \texttt{HSR} split in \cref{fig:keyframes}. If $o$ stems from an isolated mask and $s_{ij} > \epsilon_{isolated}$ or the point cloud-based overlap of $d$ wrt. $o$ is greater than $\epsilon_{overlap}$ we merge $d$ into $o$. If $o$ is not isolated, \ie, it was originally observed at the image boundaries, we first check whether $d$ is isolated and, if true, check whether $o$ is a strong inlier wrt. to $d$ and merge $d$ into $o$. In this fashion, we update the status of $o$ to be an isolated object, as the recent merge of $d$ verified this. In case both $d$ and $o$ are non-isolated, we proceed with merging $d$ into $o$ under the initial thresholding of $\mathbf{S}$. We opt for this approach as we observed numerous cases of inconclusive segmentations hindering object articulation understanding.

\begin{figure}[htbp]
    \centering
    \includegraphics[width=1.0\linewidth]{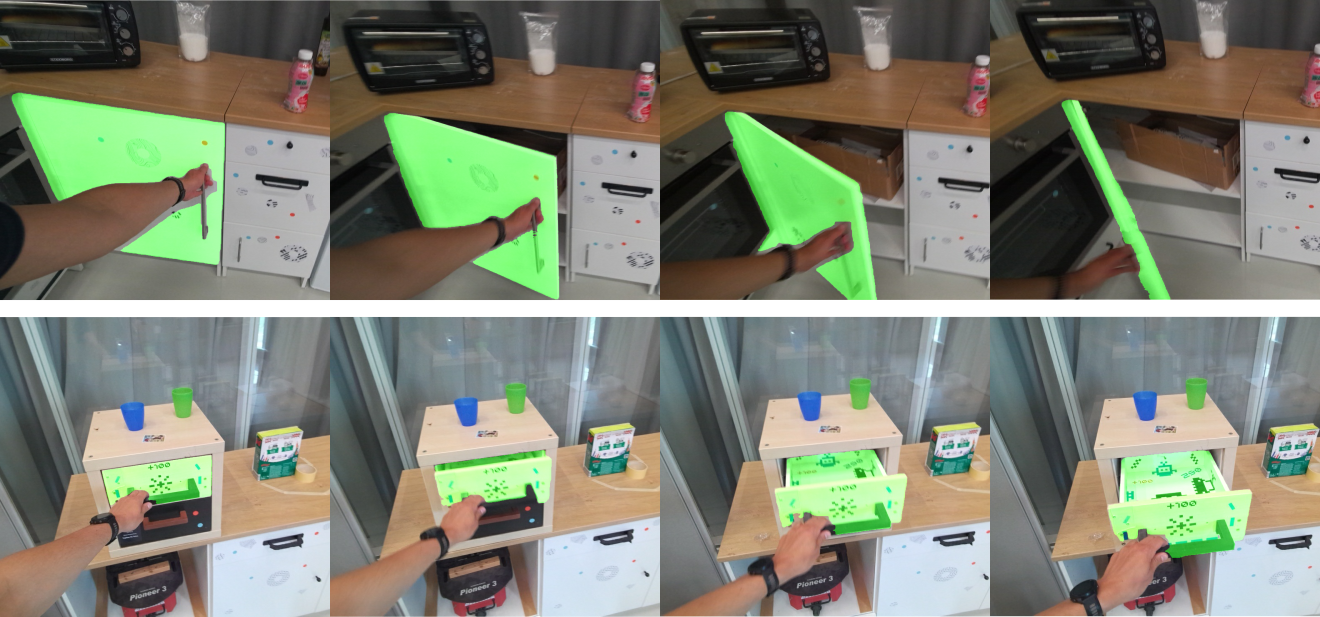}
    \caption{Dense object segmentation for object reconstruction using SAM2 based on repeated TapNext++ initialization along each interaction.}
    \label{fig:sam2_dense_segmentation}
\end{figure}

\noindent{\textbf{Dense Articulated Object Reconstruction:}} If computational means allow, \ours{} includes a dense pixel-wise reconstruction mode yielding frame-wise object masks that enable object reconstruction and provides the means to perform precise collision checking among close-by articulated objects. \ours{} incorporates that by running the TapNext++ repeatedly, \ie, at various points in time (1/4, 1/2, 3/4) throughout a detected interaction. As such, the extracted dynamic point tracks cover the articulated part at each time step. We follow the same articulation estimation pipeline as usual, but evaluate all additional sets of point tracks under the articulation model to filter for meaningful dynamic tracks. Next, at certain keyframes within the interaction, we query Semantic-SAM~\cite{li2023semantic} to obtain part-level masks, which we, in turn, score based on the ratio of dynamic and static point tracks falling into them. Masks exceeding a ratio of 0.5 are fed as mask prompts to SAM2 to obtain a temporal segmentation of the articulated object. We visualize a revolute and a prismatic articulated object and the corresponding pixel-wise segmentations in \cref{fig:sam2_dense_segmentation}.

\subsection{Object-to-Articulation Matching}
\label{sec:supp_obj_articulation_matching}
\begin{figure*}[ht]
\centering
\footnotesize
\caption{Qualitative results of articulation-to-object association scheme introduced in \cref{sec:articulated_scene_graph}. Across all depicted frames, the shaded mask represents the most appropriate identified \textit{free} object when articulated under the estimated articulation model that is projected into the camera frame as a red axis. Solid green image boundaries denote successful pairings, dashed yellow intermediate cases represent either wrong segmentations or misidentified articulation magnitudes under accurate articulation models, and dotted red boundaries denote wrongly associated object masks under permissible estimated axes.} 
\includegraphics[width=\linewidth]{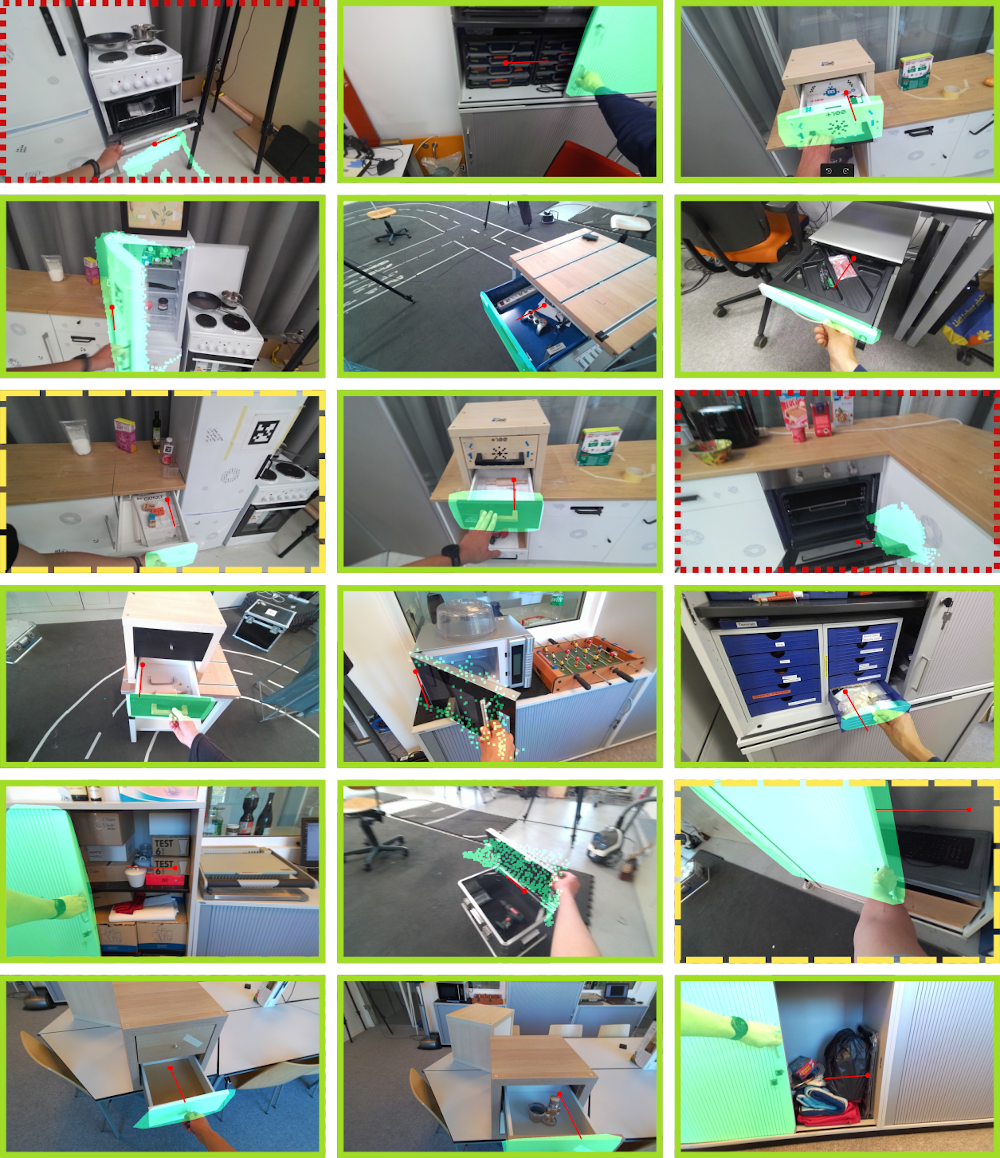}
\label{fig:articulation_object_matching}
\vspace{1em}
\end{figure*}

In the following, we expand on the object-to-articulation matching logic introduced in the main manuscript in \cref{sec:articulated_scene_graph}. We divide the obtained point tracks $\mathcal{X}_n$ into dynamic and static points by evaluating the position variance along each track and filtering out tracks with significant \textit{jumps} due to noisy depth at object boundaries. As outlined in the main manuscript, we measure the ratios of dynamic and static points validating an object $j$. The proposed sum of those ratios yields our object-to-articulation matching score $p_{ij}$. In detail, we measure the ratio of visible dynamic 2D keypoints $x_k$ that fall into the convex set $\Omega(o_j^{2D})$ of the projected 2D object mask of node $j \in \mathcal{O}$ under a certain articulation state $\theta$ and, likewise, penalize the ratio of static keypoints within the interior:

\begin{equation}
\label{eq:pij}
p_{ij} = \sum_{t=t_s}^{t_e} \frac{\sum_{k=1}^{F} \mathbf{1}_{\operatorname{int}(\Omega)}(x_{k,t}^{d})}
     {\sum_{k=1}^{F} \mathbf{1}_{\mathbb{R}^d \setminus \Omega}(x_{k,t}^{{d}})} + \left(1 - 
     \frac{\sum_{k=1}^{F} \mathbf{1}_{\mathbb{R}^d \setminus \Omega}(x_{k,t}^{{s}})}{\sum_{k=1}^{F} \mathbf{1}_{\operatorname{int}(\Omega)}(x_{k,t}^{s})}\right),
\end{equation}

where $x_{k,t}^{d}$ and $x_{k,t}^{s}$ denote dynamic and static points, respectively at time $t$. The above yields a mean association cost $p_{ij}$ per articulation $i$ and object $j$. We visualize the constructed convex sets $\Omega(o_j^{2D})$ in \cref{fig:articulation_object_matching}. Whenever we can not compute the associated convex set we resort to just the 2D-projected points of $j$.

\subsubsection{Contained Object Discovery}
\label{sec:supp_contained_obj_disc}
\begin{figure*}[ht]
\centering
\footnotesize
\setlength{\tabcolsep}{0.01cm}
\begin{tabularx}{\linewidth}{ccc}
\includegraphics[trim=0cm 0cm 0 0cm, clip, width=0.5\linewidth]{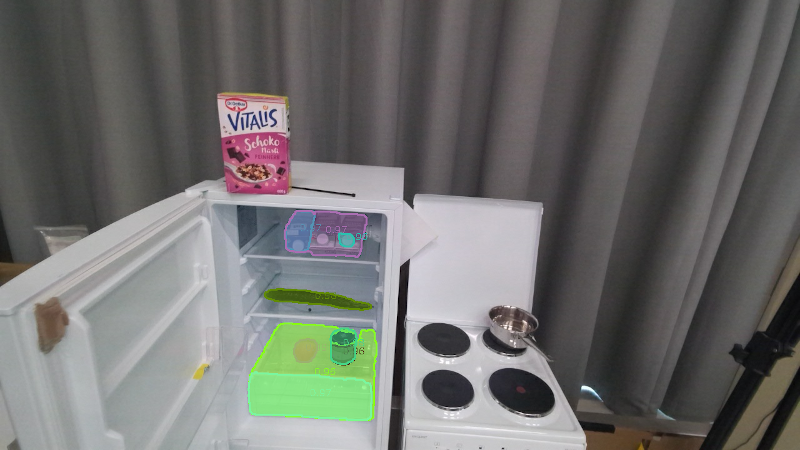} & \includegraphics[trim=0 0 0 0cm, clip, width=0.5\linewidth]{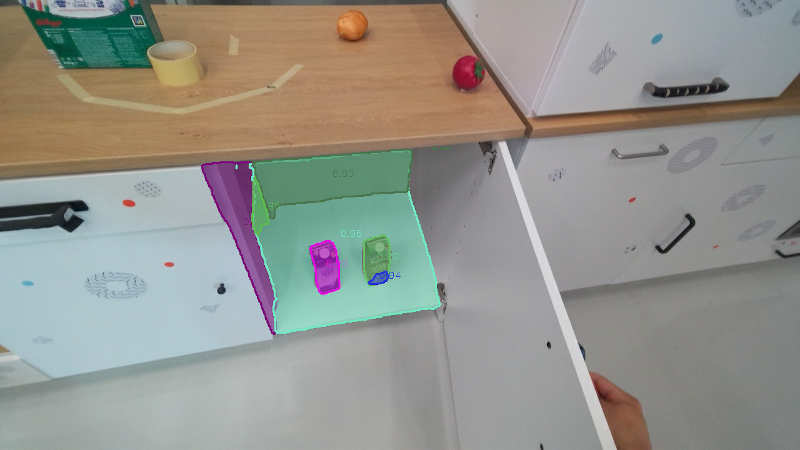} \\[-0.3em]
\includegraphics[trim=0cm 0cm 0 0cm, clip, width=0.5\linewidth]{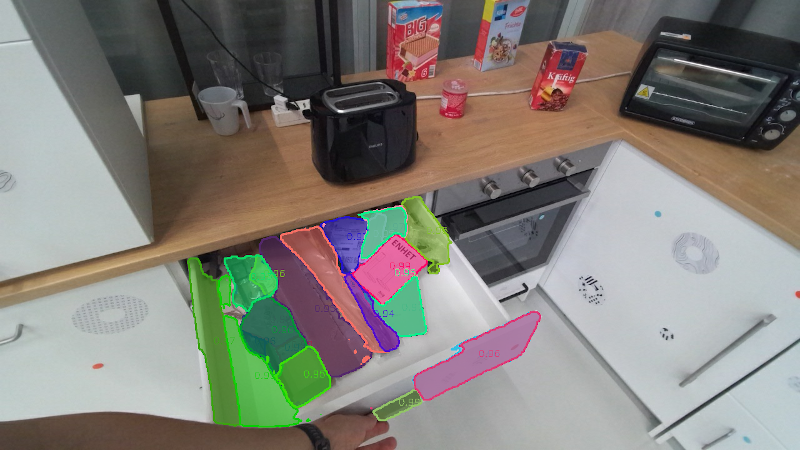} & \includegraphics[trim=0 0 0 0cm, clip, width=0.5\linewidth]{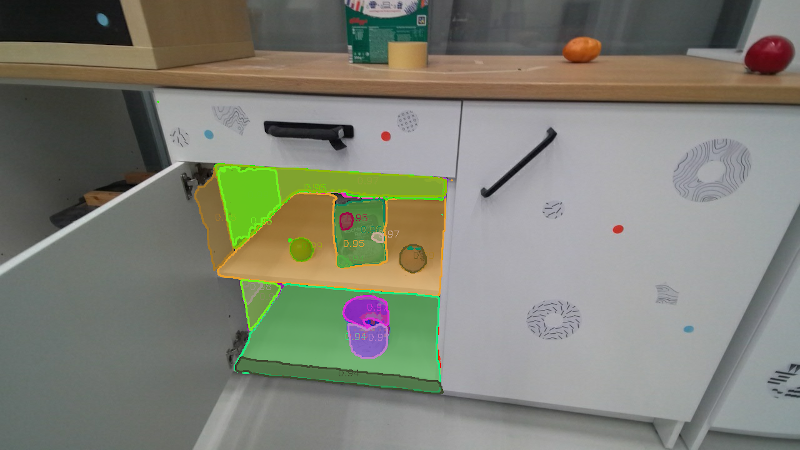} \\[-0.3em]
\includegraphics[trim=0cm 0cm 0 0cm, clip, width=0.5\linewidth]{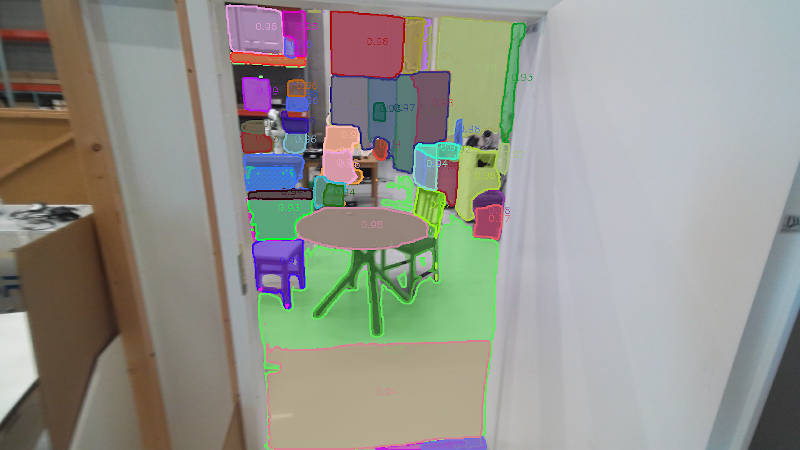} & \includegraphics[trim=0 0 0 0cm, clip, width=0.5\linewidth]{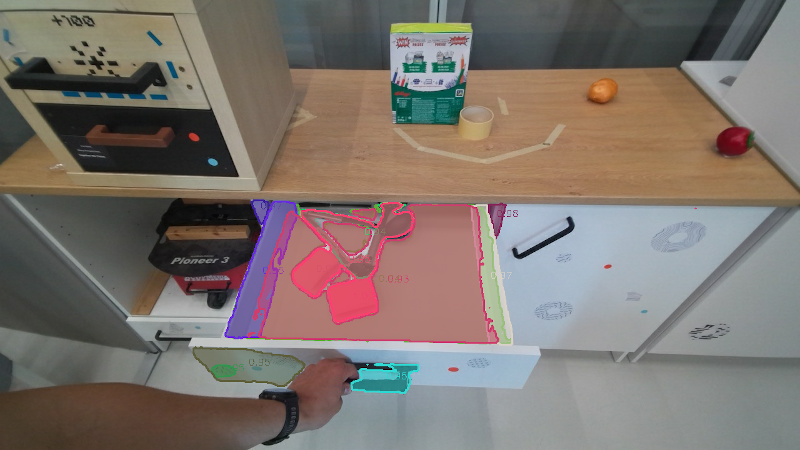} \\[-0.3em]
\includegraphics[trim=0cm 0cm 0 0cm, clip, width=0.5\linewidth]{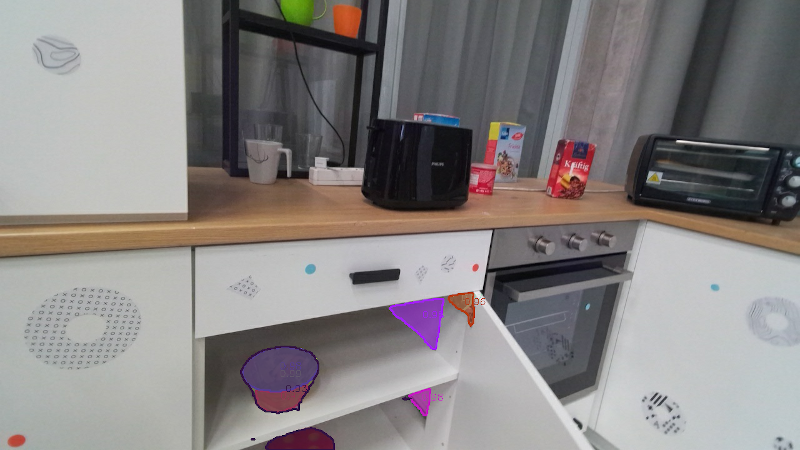} & \includegraphics[trim=0 0 0 0cm, clip, width=0.5\linewidth]{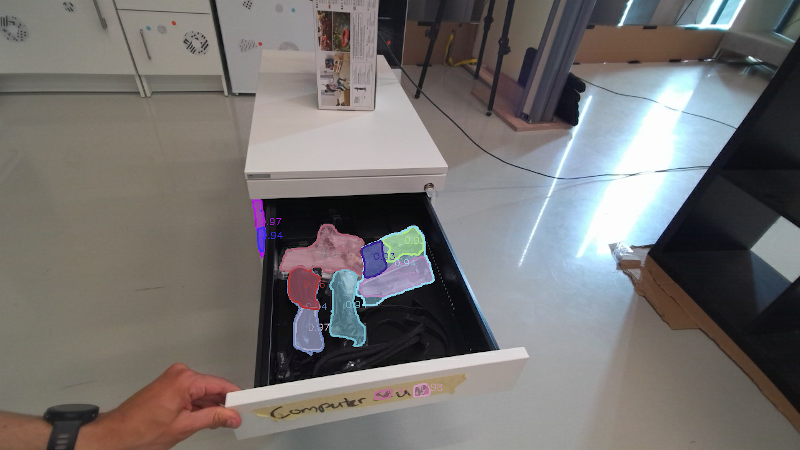} \\[-0.3em]
\includegraphics[trim=0cm 0cm 0 0cm, clip, width=0.5\linewidth]{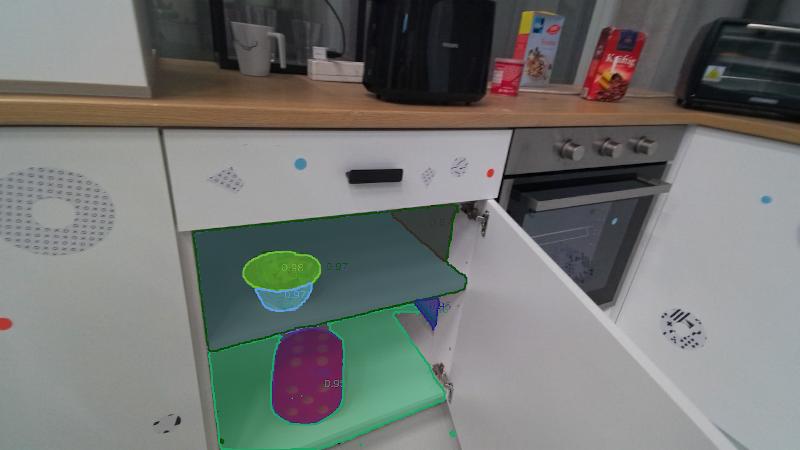} & \includegraphics[trim=0 0 0 0cm, clip, width=0.5\linewidth]{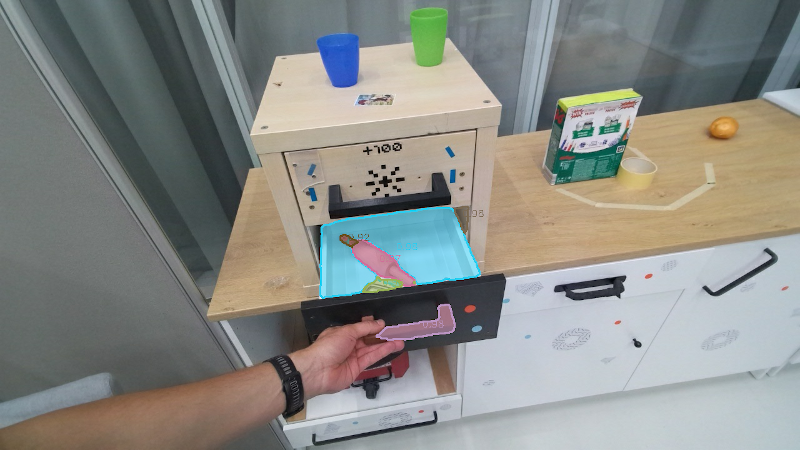} \\[-0.3em]
\end{tabularx}
\caption{We depict the object discovery component of \ours{}. Each of the 10 frames displays identified maximum-opening object states. Instance masks represent identified child objects wrt. the articulated parent.} 
\label{fig:children}
\vspace{-2em}
\end{figure*}

We provide additional qualitative insights on the contained object discovery in \cref{fig:children} and \cref{fig:contained_object_discovery}. 

\clearpage
\subsection{Real-World Mobile Manipulation}
\label{sec:moma_supp}

In this section, we outline the mobile manipulation stack for performing object opening and closing actions in the real world as well as mobile manipulation from natural language inputs.

\subsubsection{Global Localization in Ego-Centric Maps}
\label{sec:egocentric_loc}

In order to localize a robot within, \eg, an ego-centric scene, accurate alignment is required while considering sensor drift throughout map construction. To accomplish this, we first localize the robot at the start of a trial globally. In order to allow for safe object manipulation, we perform an additional registration step before each object manipulation action. In the following, we detail the general approach utilizing 2D LiDAR sensors as this is the most common localization modality in mobile manipulation, as with the Toyota HSR.

\noindent{\textbf{Ground Plane Estimation and Global Alignment:}}
Given the ego-centric 3D map as constructed in \cref{sec:part_mapping_supp}, we first perform robust ground plane estimation to align the recorded point cloud to a single plane normal in order to allow for 2D localization in the bird's-eye-view. We employ random sample consensus (RANSAC) estimation by iterative sampling of three support points of the raw scene point cloud $\mathcal{P}$ in order to construct candidate planes described using a normal $\mathbf{n}_{\mathcal{P}}$. We then score all points $\mathbf{p}_i \in \mathcal{P}$ by evaluating the number of inliers satisfying
\begin{equation}
| \mathbf{n}^\top \mathbf{p}_i + d | < \epsilon,
\end{equation}
where $d$ constitutes a scalar offset and $\epsilon$ is a tuned distance threshold. This yields the predominant $\mathbf{n}$ and a set of inlier points. Given all the inlier points, we employ singular value decomposition (SVD) to refine the RANSAC-estimate $\mathbf{n}$ of the ground plane; thus ensuring an optimal fit to the ground surface. Once the ground plane $(\mathbf{n}, d)$ is estimated, we compute a rigid body transformation ${}^{W}\mathbf{T}_{G} \in SE(3)$ using the Rodrigues' rotation formula to align the ego-centric map such that the ground plane coincides with the $xy$-plane, where $z$ is the upwards-pointing height axis. ${}^{W}\mathbf{T}_{G}$ represents the transformation that allows us to transform objects of the constructed articulated 3D scene graph observed during the mapping stage into world coordinates when at robot deployment time.

\noindent{\textbf{Construction of 2D Occupancy Grid:}}
Given the aligned point cloud expressed in the world frame, we convert its 3D points into a 2D occupancy grid by slicing, projecting, and post-processing free and occupied regions. To do so, we slice the point cloud below \qty{0.05}{m}, which produces floor observations, while points in  $[\qty{0.15}{m}, \qty{0.30}{m}]$ are treated as occupied as they are observed by a 2D LiDAR. All points are then projected to a 2D grid using a world-to-voxel mapping while discarding points outside the grid bounds. Initially, all cells are denoted \texttt{UNEXPLORED}. We add floor points as \texttt{FREE} cells and then overwrite the produced grid with all occupied points, yielding \texttt{OCCUPIED} cells. We apply morphological closing and erosion to remove holes in the observed ground surface. This procedure yields a 2D occupancy grid that represents navigable space as well as object boundaries. At the beginning of each trial, we localize the mobile manipulator using particle filter-based adaptive Monte Carlo localization (AMCL)~\cite{fox1999mcl}.

\subsubsection{Local Online Ego-centric Map Registration}

To maintain spatial consistency between the robot's real-time observations and the precomputed global scene graph, we implement an online registration pipeline. This process aligns a local point cloud, generated from current RGB-D sensor data, with a localized subset of the global 3D map.

\noindent{\textbf{Local Online Submaps:}}
Before each manipulation action, we navigate to the object to be manipulated and acquire a depth image $\mathbf{D}_t$ online and its corresponding RGB frame. Given the current estimated camera pose ${}^{W}\mathbf{T}_{C} \in SE(3)$, we project the depth map to a point cloud $\mathcal{P}_{local}$ and truncate depth values beyond $\delta_{max} = 1.5\text{m}$, thus focusing on the object in question:
\begin{equation}
    {}^{C}\mathcal{P}_{local} = \{ \pi^{-1}(\mathbf{u}, \mathbf{D}_t(\mathbf{u})) \mid \mathbf{D}_t(\mathbf{u}) < \delta_{max} \}
\end{equation}
where $\pi^{-1}$ denotes the inverse projection from image coordinates $\mathbf{u}$ to 3D space. This local cloud is then transformed into the world frame: $^{W}\mathcal{P}_{local} = {}^{W}\mathbf{T}_{C} {}^{C}\mathcal{P}_{local}$.

\noindent{\textbf{Egocentric Reference Submaps:}}
In the next step, we register the live observation against a local crop of our ego-centric map $^{W}\mathcal{P}_{sub} \subseteq {}^{W}\mathcal{P}$ in world coordinates since global fitting is usually prone to local minima. Given the centroid $\mathbf{c}_{local}$ of $^{W}\mathcal{P}_{local}$, we construct a kd-tree over the global 3D map and perform radius search, which yields
\begin{equation}
    ^{W}\mathcal{P}_{sub} = \{ \mathbf{q} \in {}^{W}\mathcal{P} \mid \|\mathbf{q} - \mathbf{c}_{local}\|_2 < r_{local} \}
\end{equation}
where $r_{local} = 0.8\text{m}$ defines the radius of the desired region. Registering the live observation against $^{W}\mathcal{P}_{sub}$ thus allows us to focus the registration process only on the geometry that is relevant to the object to be interacted with.

\noindent{\textbf{Constrained Local Refinement:}}
Finally, we align our submap ${}^{W}\mathcal{P}_{sub}$ against $^{W}\mathcal{P}_{local}$ using point-wise iterative closest point (ICP) registration. This results in a transformation ${}^{\tilde W}\mathbf{T}_{W}$ that allows us to transfer various primitives such as an object pose ${}^{G}\mathbf{T}_{obj} \in SE(3)$ from graph coordinates into a refined world frame $\tilde W$ of the robot:
\begin{equation}
\label{eq:refine}
    {}^{\tilde W}\mathbf{T}_{obj} = {}^{\tilde W} \mathbf{T}_{W} \cdot {}^{W}\mathbf{T}_{G} \cdot {}^{{G}}\mathbf{T}_{obj} 
\end{equation}
If necessary, one can further constrain the registration result ${}^{\tilde W} \mathbf{T}_{W}$ in terms of maximum-allowed rotation and translation.

\subsubsection{Online Grasp Prediction}
\label{sec:supp_grasping}

We sample grasp poses online using the robot's head camera (HSR) / gripper camera (Spot) while facing the object to be manipulated. To do so, we employ SAM3~\cite{carion2025sam3segmentconcepts} with the prompt \textit{``handle``}, which yields handle masks $\{\mathcal{M}_k\}$. For each mask $\mathcal{M}_k$, a partial point cloud is constructed given the corresponding depth image transformed into world coordinates given the above online registration.

After filtering the obtained point cloud, we generate a candidate grasp pose by first computing the handle's oriented bounding box. Let $\mathbf{R} \in \mathbb{R}^{3 \times 3}$ denote its rotation matrix and $\boldsymbol{\ell} \in \mathbb{R}^3$ its half-extents. We identify the predominant axis of the found handle point cloud by identifying the \textit{longest} axis
\begin{equation}
\mathbf{a}_p = \mathbf{R}_{:,\,\arg\max_i \ell_i},
\end{equation}
which corresponds to the pre-dominant principal dimension of the handle. We compute the min and max points along this axis of the bounding box and compute the tool center point as
\begin{equation}
\mathbf{t}_g = \tfrac{1}{2}(\mathbf{p}_{\min} + \mathbf{p}_{\max}),
\end{equation}
representing its midpoint. In order to construct the grasp approach axis, we employ the normal $\mathbf{n}_{\theta}$ of the object to be opened/closed at its respective articulation state $\theta$. Finally, we compute the closing axis given the cross product of the two aforementioned vectors. The resulting grasp pose is then defined as
\begin{equation}
\mathbf{T}_{grasp} = 
\begin{bmatrix}
-\mathbf{n}_{\theta} & -\mathbf{a}_p \times \mathbf{n}_{\theta} & \mathbf{a}_p & \mathbf{t_g}. \\
0 & 0 & 0 & 1
\end{bmatrix}
\end{equation}

Naturally, a single observation of the scene can contain multiple handles. To associate candidate grasps with objects in the scene, we maintain world-frame point clouds and corresponding axis-aligned bounding boxes. For each object--grasp pair, we compute a compatibility score that consists of the IoU between grasp and object, the overlap of the grasp wrt. the object, and the distance of the involved centroids. The resulting score matrix defines a bipartite matching problem, which we solve for using the Hungarian algorithm. The grasp assigned to the target object is selected for manipulation.

\begin{figure}[htbp]
    \centering
    \includegraphics[width=\linewidth]{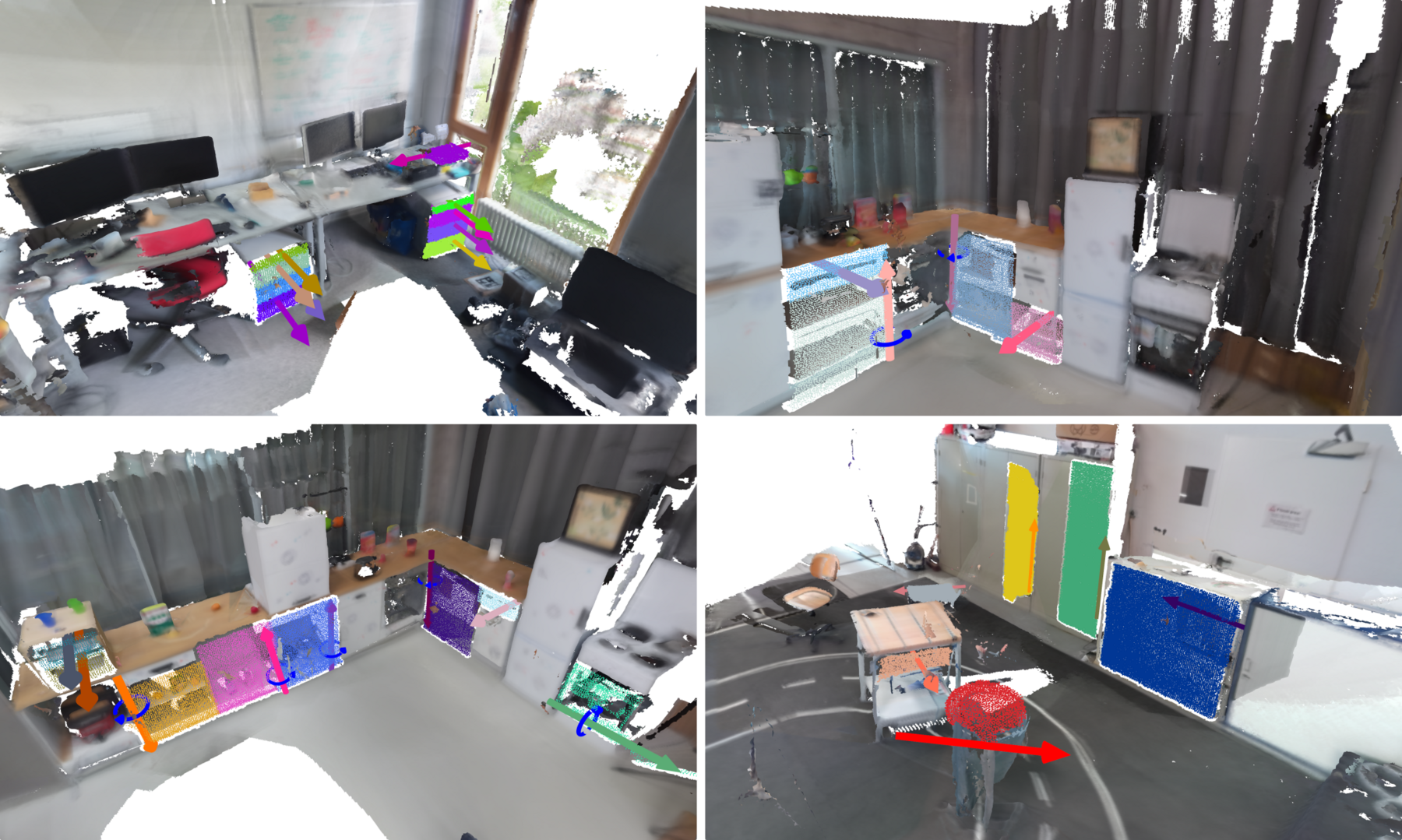}
    \caption{Qualitative full-scene results on articulated object estimation and object association on splits of Arti4D-Semantic.}
    \label{fig:full_scene_plots_arti4d_semantic}
\end{figure}

\begin{figure}[htbp]
    \centering
    \includegraphics[width=1.0\linewidth]{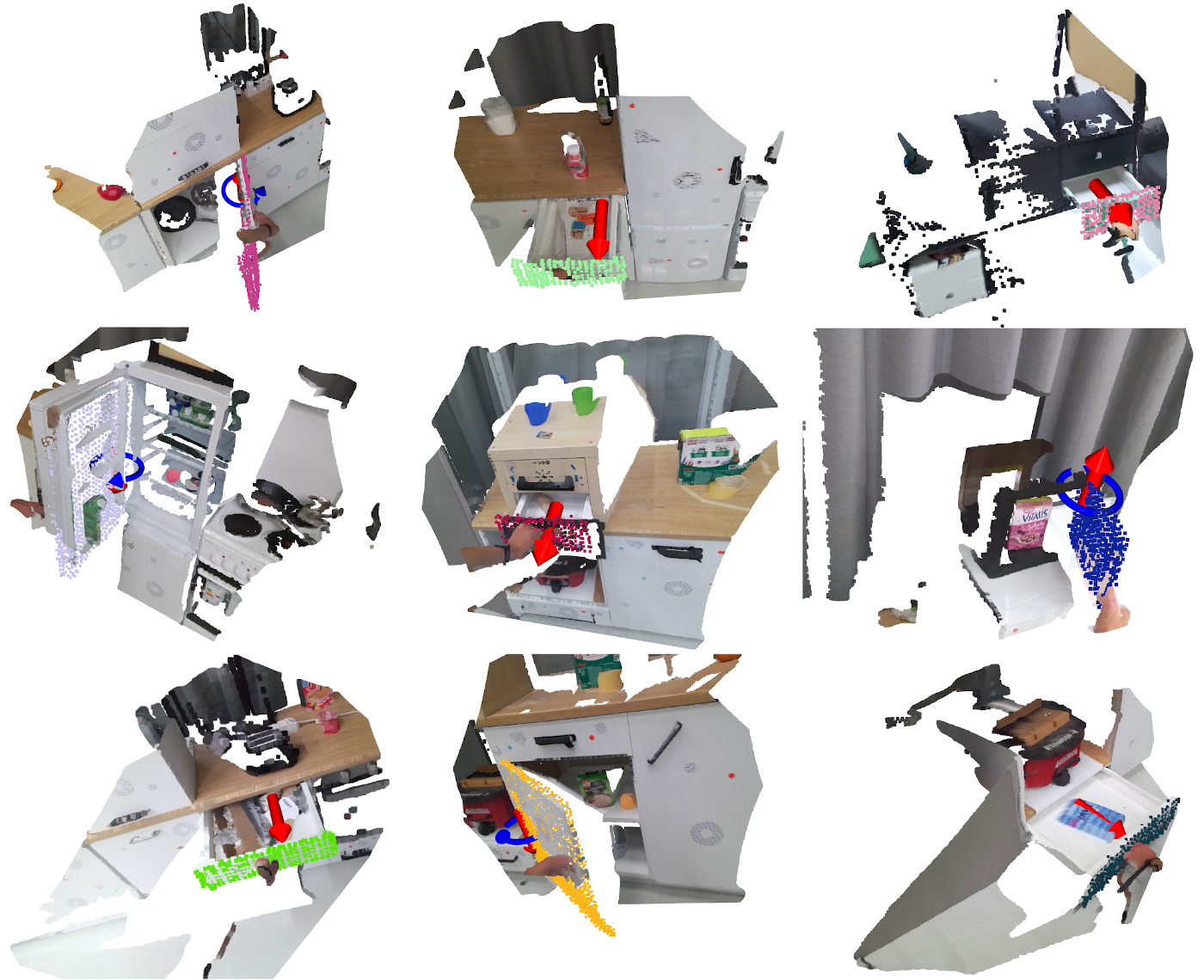}
    \caption{Estimated articulated 3D objects: The colored arrow represents the articulation axis and the distinctly colored point cloud represents the object. A torus around an axis denotes a revolute axis.}
    \label{fig:articulated_3d_objects}
\end{figure}

\subsubsection{End-Effector Trajectory Sampling}
\label{sec:ee_traj_sampling}

To manipulate articulated objects, the robot must generate an end-effector trajectory that is kinematically consistent with the object's articulation model. We synthesize this trajectory by first computing a view-dependent grasp pose and then propagating it under a range of articulation states using the exponential map of the estimated twist $\hat{\xi}$.

Given a grasp pose ${}^{\tilde W}\mathbf{T}_{grasp}$ in refined world coordinates $\tilde W$ and the object's articulation model $\mathcal{A}_{i}$ given as a twist $\hat\xi \in \mathfrak{se}(3)$ in graph coordinates, we sample an end-effector trajectory ${}^{\tilde W}\mathbf{T}_{ee}(\theta)$ under the configuration states $\Theta = \{\theta_0, \theta_1, \dots, \theta_n\}$. For each $\theta \in \Theta$, the global end-effector pose is computed by evaluating the twist under the grasp pose in graph coordinates and transforming back to world coordinates.
\begin{equation}
    {}^{\tilde W}\mathbf{T}_{ee}(\theta) = {}^{\tilde W}\mathbf{T}_{W}  \cdot {}^{W}\mathbf{T}_{G} \cdot \left( \exp(\hat{\xi} \theta) \cdot {}^{G}\mathbf{T}_{grasp} \right)
\end{equation}
where $\exp(\cdot)$ represents the exponential map from $\mathfrak{se}(3)$ to $SE(3)$, $\hat{\xi}$ is the twist governing the articulation and ${}^{G}\mathbf{T}_{grasp}$ is the initial grasp pose transformed into graph coordinates given \cref{eq:refine}. Depending on whether we perform an opening or a closing action, we reverse/negate the range of $\Theta$.

\subsubsection{Language-enabled Mobile Manipulation}
\label{sec:moma_language}
On top of the methodology of MoMa-SG, we demonstrate how its capabilities are integrated into an LLM-based planner, which allows to perform various object manipulations such as opening or closing objects. To do so, we define a discrete set of high-level actions that span observation, navigation, and manipulation skills and specify arguments to be parsed by each of those functions. In addition, we provide the LLM with the current state of all considered articulated objects at execution time: $[(obj_1, M_i), .., (obj_N, M_N)]$, where $M$ constitutes a semantic abstraction mapping from articulation states to semantic states as introduced in \cref{sec:articulated_scene_graph}. Next, the LLM determines the most appropriate high-level action to be executed in an online, receding horizon fashion based on the provided scene context. Given a high-level action call, we propagate this instruction to various low-level controllers responsible for navigation, inspection, and manipulation. We consider the following set of actions:
\begin{itemize}
    \item \texttt{open}$(o)$: Navigate to an articulated object, door, or container $o$ and open it.
    \item \texttt{inspect}$(o)$: Navigate to an articulated object, door, or container $o$, register, and inspect its contents.
    \item \texttt{close}$(o)$: Close an articulated object, door, or container $o$.
    \item \texttt{retrieve}$(o, p)$: Retrieve object $o$ from a parent container $p$.
    \item \texttt{done}$()$: Indicate that the task has been completed.
\end{itemize}
Furthermore, we are also able to run the \texttt{open()}, \texttt{close()}, and \texttt{retrieve()} actions with a boolean retrial flag, which triggers an \texttt{inspect()} call after each performed action in order to check whether the desired state was actually reached. If not, we trigger a retrial. We find that this allows to minimize stochastic gripping failures occurring throughout manipulation. We employ the N2M2 reinforcement learning agent~\cite{honerkamp2023n} as a whole-body controller by adapting it to accept general end-effector $SE(3)$ trajectories, which we sample as outlined in \cref{sec:ee_traj_sampling}. As the language model, we utilize GPT-5. We depict language-based long-horizon object manipulation in \cref{fig:retrieve}

\begin{figure*}[ht]
\centering
\footnotesize
\setlength{\tabcolsep}{0.01cm}
\begin{tabularx}{\linewidth}{ccc}
\includegraphics[trim=10cm 4cm 0 0cm, clip, width=0.5\linewidth]{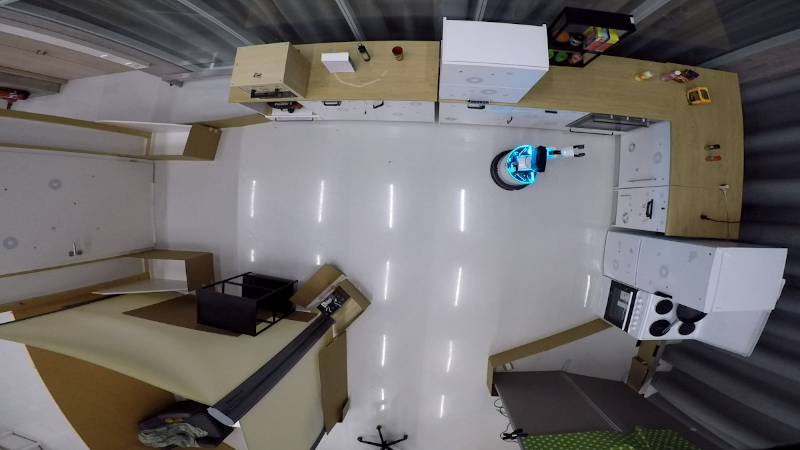} & \includegraphics[trim=10cm 4cm 0 0cm, clip, width=0.5\linewidth]{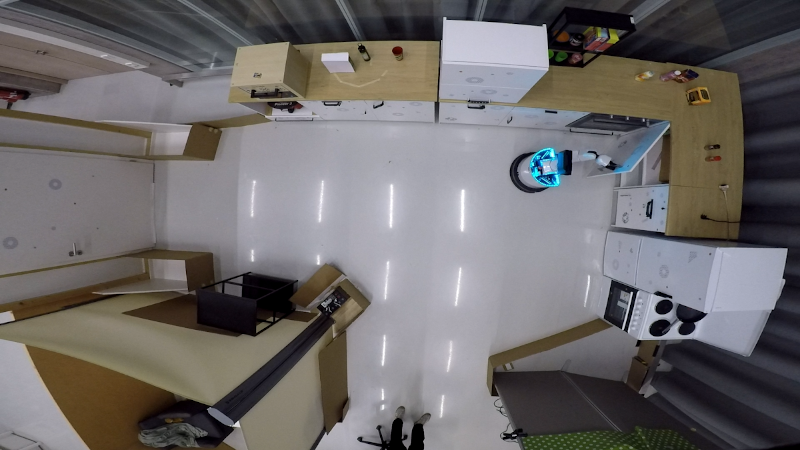} \\[-0.3em]
\includegraphics[trim=10cm 4cm 0 0cm, clip, width=0.5\linewidth]{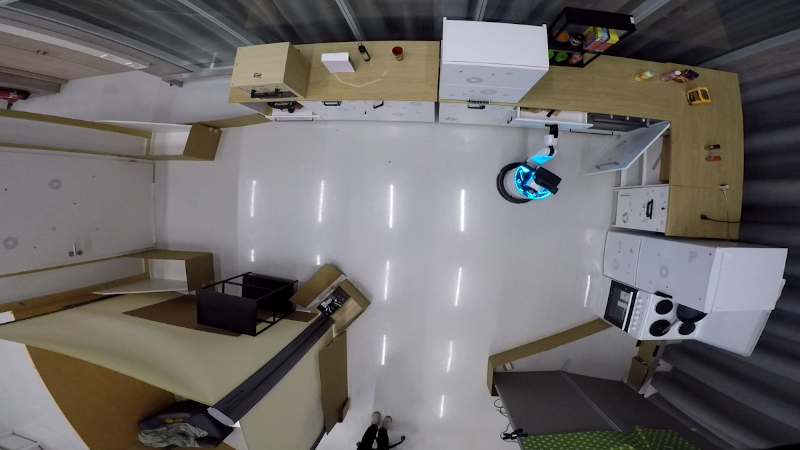} & \includegraphics[trim=10cm 4cm 0 0cm, clip, width=0.5\linewidth]{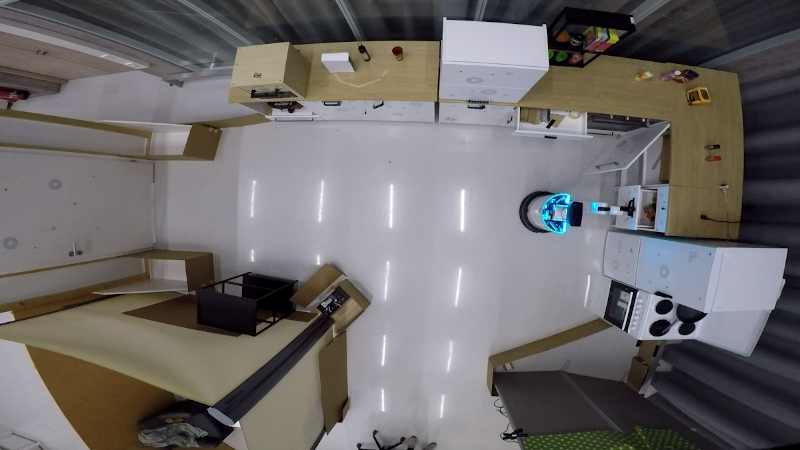} \\[-0.3em]
\includegraphics[trim=10cm 4cm 0 0cm, clip, width=0.5\linewidth]{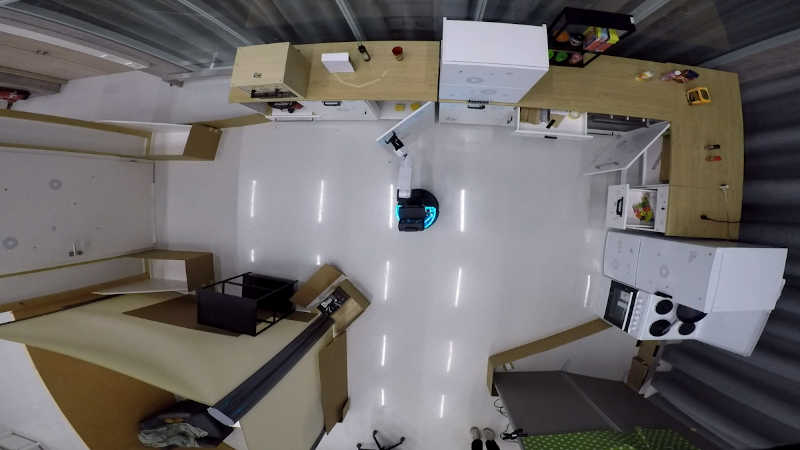} & \includegraphics[trim=10cm 4cm 0 0cm, clip, width=0.5\linewidth]{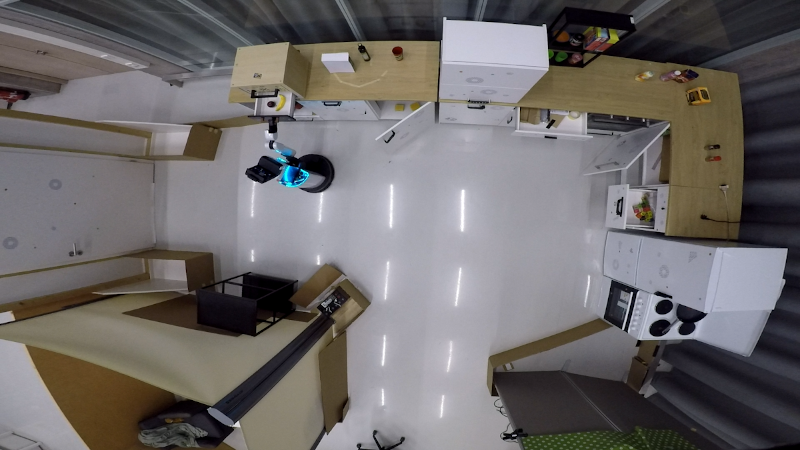} \\[-0.3em]
\includegraphics[trim=10cm 4cm 0 0cm, clip, width=0.5\linewidth]{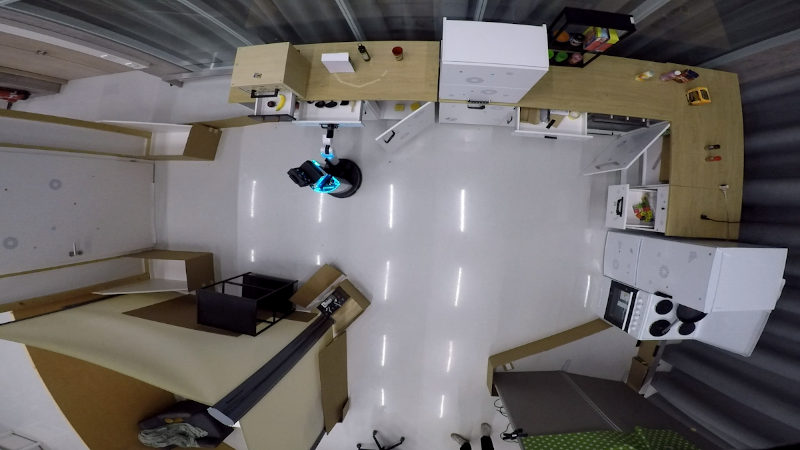} & \includegraphics[trim=10cm 4cm 0 0cm, clip, width=0.5\linewidth]{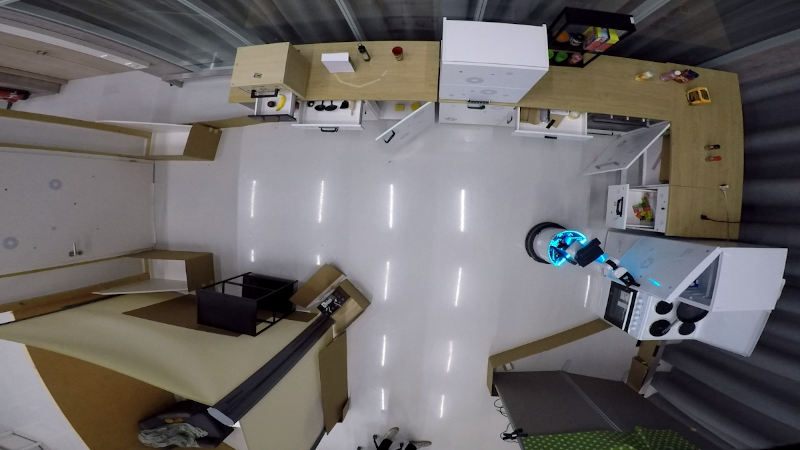} \\
\end{tabularx}
\caption{We depict the ability of \ours{} to perform long-horizon mobile manipulation of articulated scenes using the Toyota HSR. Throughout this episode (left to right, top to bottom), the robot successfully opened 7 different articulated objects, 3 revolute objects, and 4 prismatic objects. Throughout our experiments, we found that grasping failures are much more dominant than errors due to mediocre articulation axes.} 
\label{fig:opening}
\end{figure*}

\begin{figure}[htbp]
\centering
\footnotesize
\includegraphics[width=\linewidth]{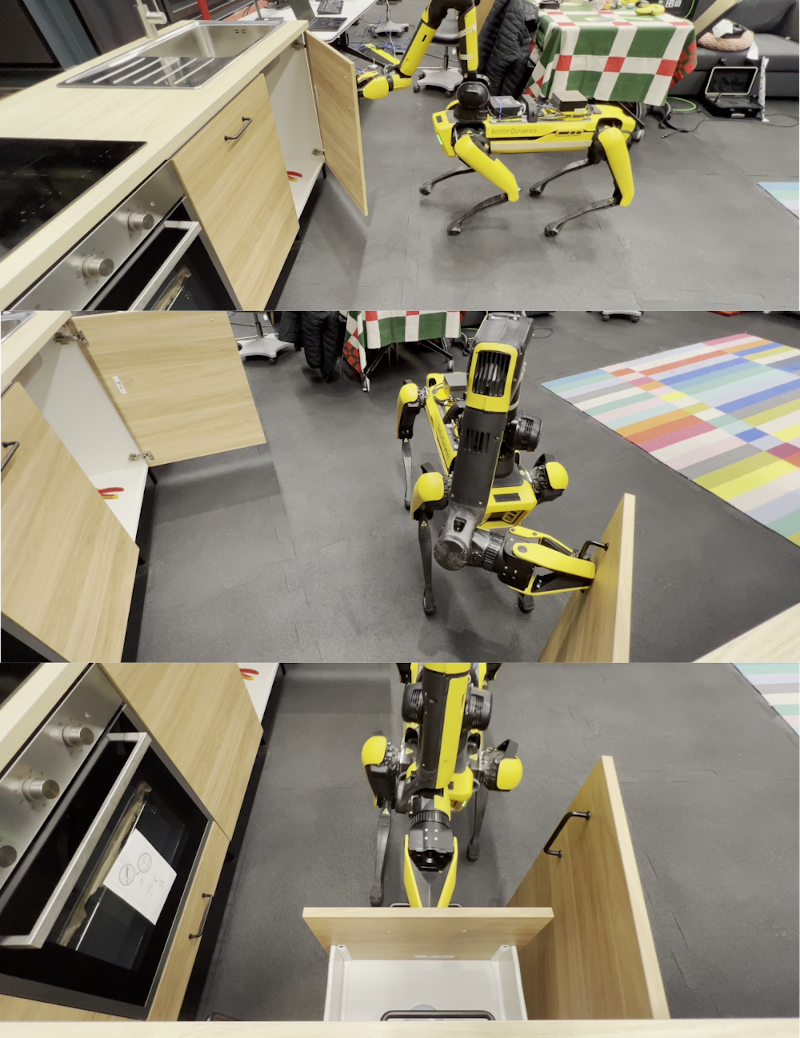}
\caption{We depict three consecutive opening actions conducted using the Boston Dynamics Spot.} 
\label{fig:opening_spot}
\end{figure}

\clearpage
\section{Experimental Evaluation}
\label{sec:supp_experiments}

\subsection{Metrics}
\label{sec:metrics_supp}

\subsubsection{Interaction Segmentation}
We evaluate the intersection segmentation using several metrics being a 1D-IoU over the time series, as well as a segment-based IoU that involves matching predicted segments against ground truth segments. The 1D IoU is computed by evaluating the binary time series of predicted and ground truth timestamps. In the following, we denote the time series as $P$ and $G$ and measure the Jaccard index:
\begin{equation}
\mathrm{IoU}_{\mathrm{1D}} = \begin{cases} 
\frac{|P \cap G|}{|P \cup G|} & \text{if } |P \cup G| > 0 \\ 
0 & \text{if } |P \cup G| = 0 
\end{cases}
\end{equation}
Furthermore, we evaluate temporal segmentation by matching predicted and ground truth interaction segments using bipartite matching. We compute the segment-wise IoU over pairs of predicted $p = (s_p, e_p)$ and a ground-truth interval $g = (s_g, e_g)$, respectively, and compute a segment-IoU:
\begin{equation}
\mathrm{IoU}(p, g) = 
\frac{\max\!\left(0, \min(e_p, e_g) - \max(s_p, s_g)\right)}
{\max(e_p, e_g) - \min(s_p, s_g)} .
\end{equation}
Given all pairs, this produces a cost matrix. We solve for an optimal pairing using the Hungarian algorithm to obtain a one-to-one matching maximizing total overlap. We discard pairs below an IoU of $0.5$. This yields a precision $P$, a recall $R$ as well as the mean segment-based IoU over all obtained pairs.

For matched segment pairs, we additionally report the mean IoU as well as the mean onset and offset errors, where the errors for each pair are defined as $|s_p - s_g|$ and $|e_p - e_g|$, respectively.

\begin{table*}[ht]
\color{black}
\scriptsize
\caption{Split-wise articulated object estimation on Arti4D-Semantic using ground truth interaction segmentation: We report the axis-angle errors $\theta_{err}$, the positional errors $d_{L2}$ of revolute joints, and the accuracy of predicted joint types, including their respective recalls as in \cref{tab:articulation_estimation}.}
\setlength\tabcolsep{3pt}
\begin{threeparttable}
\begin{tabular}{lcc|cccccccc}
      \toprule
          \multirow{2}{*}{Split} & \multirow{2}{*}{Paradigm} & \multirow{2}{*}{GT Poses} & \multicolumn{2}{c}{Prismatic joints} & \multicolumn{2}{c}{Revolute joints} & \multicolumn{3}{c}{Type Quality [\%]} & Mode \\
 & & & $\theta_{err}$[deg] & $d_{L2}$[m] & $\theta_{err}$[deg] & $d_{L2}$[m] & Accuracy & Prismatic Recall & Revolute Recall & Acc.[\%]\\
        \midrule
          \texttt{RH078} & \textit{EGO} & \checkmark & 05.43 & -- & 15.88 & 0.063 & 0.862 & 0.840 & 0.900 & 0.980 \\
          \texttt{RR080} & \textit{EGO} & \checkmark & 19.36 & -- & 14.80 & 0.035 & 0.808 & 0.792 & 1.000 & 0.952 \\
          \texttt{DR080} & \textit{EGO} & \checkmark & 07.99 & -- & 15.67 & 0.094 & 0.940 & 0.922 & 0.970 & 1.000 \\
          \texttt{RH201} & \textit{EGO} & \checkmark & 03.52 & -- & 13.97 & 0.099 & 0.979 & 0.973 & 0.979 & 0.993 \\
          \texttt{MHZH} & \textit{EGO} & \(\times\) & 42.01 & -- & 21.34 & 0.271 & 0.917 & 0.727 & 1.000 & 0.972 \\
          \texttt{MHZH-EXO} & \textit{EXO} & \(\times\) & 30.24 & -- & 36.56 & 0.172 & 0.778 & 0.500 & 0.917 & 1.000 \\
          \texttt{HSR} & \textit{ROBOT} & \(\times\) & 30.99 & -- & 51.05 & 0.367 & 0.605 & 0.684 & 0.526 & 0.852\\
          \midrule
           & Overall & & \textbf{19.94} & -- & \textbf{24.18} & \textbf{0.157} & \textbf{0.841} & \textbf{0.777} & \textbf{0.899} & \textbf{0.964} \\
        \bottomrule
\end{tabular}
\footnotesize
\end{threeparttable}
\label{tab:articulation_scene_results}
\color{black}
\end{table*}

\subsubsection{Articulation Estimation}
\label{sec:articulation_estimation_metrics}
In the main manuscript, we opted for using ground truth interaction segmentation in comparison to \textit{ArtiPoint} in order to ensure comparability among all considered methods. Interactions are labeled based on the first and last timestamps that show each particular articulated object in motion. That obviously means that across non-interaction timestamps, there is still dynamic motion induced by interactors/humans moving throughout the scene or grasping objects. Nonetheless, given ground-truth interaction segments, we hold the means to compare actual prediction-ground truth articulation pairs and follow the evaluation protocol introduced for Arti4D~\cite{arti25werby}:

We compute the positional error $d_i$ between a predicted articulation axis $\hat{\mathbf{a}}$ and the corresponding ground truth axis $\mathbf{a}_{gt}$ while relying on support points $\hat{\mathbf{p}}, \mathbf{p}_{gt}$ along those axes:
\begin{equation}
    d_i = \begin{cases}
        \frac{(\hat{\mathbf{p}}_i - \mathbf{p}_{gt})^\top (\hat{\mathbf{a}}_i \times \mathbf{a}_{gt})}{\norm{\hat{\mathbf{a}}_i \times \mathbf{a}_{gt}}} & \text{if } \norm{\hat{\mathbf{a}}_i \times \mathbf{a}_{gt}} > \epsilon \\
        \norm{(\hat{\mathbf{p}}_i - \mathbf{p}_{gt}) \times \mathbf{a}_{gt}} & \text{else } 
    \end{cases},
\end{equation}
where the first case covers non-parallel axes with $\epsilon = 10^{-4}$. If a method yields a twist, we rely on its own heuristic for reconstructing the associated axis of motion. Furthermore, we follow Arti4D in computing the angular axis errors by computing normalized dot-products:

\begin{equation}
    \Theta_{err,i} = \cos^{-1} \left(\frac{\mathbf{a}_{gt}^\top \hat{\mathbf{a}}_i }{\norm{\mathbf{a}_{gt}}\norm{\hat{\mathbf{a}}_i}}\right).
\end{equation}

Similar to Arti4D, we do only report angular errors for prismatic joints, as the axis location of estimated prismatic objects is not informative.

\subsubsection{Articulated 3D Part Segmentation}
In order to evaluate the articulated parent objects, we consider two evaluations: instance segmentation of articulated parents in their static, non-opened state as well as semantic accuracy. Since the number of observed non-articulated objects may exceed the actual number of articulated objects, we decided to evaluate the IoU and recall at various IoU's when matching all predicted (articulated) objects against the set of ground truth articulated objects that are labeled as part of \ourdataset{}. In general, we opt for evaluating all articulated objects in their closed state. Given its objects' articulation model, we are able to transform its associated mask into the closed state.

\noindent{\textbf{Instance Segmentation:}} 
We define a match between a predicted part $P_i$ and a ground truth part $G_j$ based on their 3D Intersection over Union (IoU) of their axis-aligned bounding boxes. Across all potential pairs $(i,j)$, this constitutes an IoU matrix $\mathbf{A} \in \mathbb{R}^{N \times M}$ over $N$ predictions and $M$ ground truth objects. To find the optimal global assignment, we solve the bipartite matching problem using the Hungarian algorithm. This yields an optimal set of pairs between predicted objects and ground truth objects. First, we compute the mean over all axis-aligned IoUs of the found object pairs, which yields our IoU measure. Second, we compute the recall $\frac{|TP|}{M}$ at the IoU thresholds $\{0.25, 0.50, 0.75\}$ in order to quantify the geometric accuracy. Thus, whenever an object pair exhibits an IoU greater than, \eg, $0.25$, it is counted as a true positive.

\begin{table}
\color{black}
\centering
\scriptsize
\caption{Split-wise articulated 3D part segmentation on Arti4D-Semantic: We report the same set of metrics as in \cref{tab:articulated_parts}. Below, we list the performance using either CoTracker3~\cite{karaev2024cotracker3} or TapNext++\cite{tapnextpp2026}, which is employed in default \ours{}.}
\setlength\tabcolsep{8pt}
\begin{threeparttable}
\begin{tabular}{clcc|ccccccc}
      \toprule
          \multirow{2}{*}{Method} & \multirow{2}{*}{Split} & \multirow{2}{*}{Paradigm} & \multirow{2}{*}{GT Poses} & \multirow{2}{*}{IoU} & \multicolumn{3}{c}{Recall} \\
         & & & & & \textit{@0.25} & \textit{@0.50} & \textit{@0.75} \\
        \midrule
          \multirow{9}{*}{\rotatebox{90}{w/ CoTracker3}} & \texttt{RH078} & \textit{EGO} & \checkmark & 0.184 & 0.327 & 0.181 & 0.000 \\
          & \texttt{RR080} & \textit{EGO} & \checkmark & 0.196 & 0.347 & 0.154 & 0.035 \\
          & \texttt{DR080} & \textit{EGO} & \checkmark & 0.243 & 0.387 & 0.241 & 0.019 \\
          & \texttt{RH201} & \textit{EGO} & \checkmark & 0.262 & 0.416 & 0.287 & 0.049 \\
          \cmidrule{2-8}
          & \texttt{MHZH} & \textit{EGO} & \(\times\) & 0.208 & 0.317 & 0.108 & 0.025 \\ 
          & \texttt{MHZH-EXO} & \textit{EXO} & \(\times\) & 0.238 & 0.396 & 0.229 & 0.000 \\
          & \texttt{HSR} & \textit{ROBOT} & \(\times\) & 0.300 & 0.537 & 0.237 & 0.025 \\
          \cmidrule{2-8} 
          & & Overall & & 0.233 & 0.391 & 0.212 & 0.025 \\
          \midrule           
          \midrule
          \multirow{9}{*}{\rotatebox{90}{w/ TapNext++}} & \texttt{RH078} & \textit{EGO} & \checkmark & 0.436 & 0.740 & 0.402 & 0.144 \\
          & \texttt{RR080} & \textit{EGO} & \checkmark & 0.459 & 0.744 & 0.552 & 0.131 \\
          & \texttt{DR080} & \textit{EGO} & \checkmark & 0.404 & 0.667 & 0.398 & 0.091 \\
          & \texttt{RH201} & \textit{EGO} & \checkmark & 0.434 & 0.717 & 0.505 & 0.087 \\
          \cmidrule{2-8}
          & \texttt{MHZH} & \textit{EGO} & \(\times\) & 0.393 & 0.565 & 0.269 & 0.090 \\ 
          & \texttt{MHZH-EXO} & \textit{EXO} & \(\times\) & 0.387 & 0.683 & 0.289 & 0.074 \\
          & \texttt{HSR} & \textit{ROBOT} & \(\times\) & 0.403 & 0.541 & 0.305 & 0.077 \\
          \cmidrule{2-8}
          & & Overall & & 0.417 & 0.665 & 0.389 & 0.099 \\
        \bottomrule
\end{tabular}
\footnotesize
\end{threeparttable}
\label{tab:scene_results_part_seg}
\color{black}
\end{table}

\noindent{\textbf{Semantic Accuracy:}} To evaluate the quality of the articulated parts' CLIP features, we resort to a retrieval-based evaluation. We evaluate whether the actual ground truth category is among the $k$ highest-similarity responses when scoring the CLIP feature of the object against all considered GT objects. For this evaluation, we follow the optimal assignment obtained above. For each matched pair $(P_i, G_j)$, we extract the predicted CLIP feature vector $\mathbf{f}_i$. We then compute the cosine similarity between $\mathbf{f}_i$ and the set of all potential category embeddings $\mathcal{C}_{sem}$.
\begin{equation}
    S(i, c) = \frac{\mathbf{f}_i \cdot \mathbf{w}_c}{\|\mathbf{f}_i\| \|\mathbf{w}_c\|}
\end{equation}
where $\mathbf{w}_c$ is the text embedding for category $c \in \mathcal{C}_{sem}$. Thus, we report the Top-$k$ accuracy (for $k \in \{1, 2, 5, 10\}$), measuring to which degree the correct category $c_{gt}$ is ranked among the top-$k$ highest similarity scores.

\begin{table}
\color{black}
\centering
\scriptsize
\caption{Articulated 3D part semantics evaluation on Arti4D-Semantic: We report open-vocabulary top-k retrieval scores as outlined in \cref{sec:metrics_supp}.}
\setlength\tabcolsep{10pt}
\begin{threeparttable}
\begin{tabular}{cl|cccc}
      \toprule
          Object Set & Split & $top_1$ & $top_2$ & $top_5$ & $top_{10}$ \\
          \midrule
          \multirow{8}{*}{$\mathcal{O}$} & \texttt{RH078} & 0.387 & 0.506 & 0.660 & 0.830 \\
          & \texttt{RR080} & 0.069 & 0.174 & 0.452 & 0.532 \\
          & \texttt{DR080} & 0.489 & 0.663 & 0.774 & 0.888 \\
          & \texttt{RH201} & 0.410 & 0.605 & 0.781 & 0.915 \\
          & \texttt{MHZH} & 0.357 & 0.523 & 0.725 & 0.980 \\ 
          & \texttt{MHZH-EXO} & 0.333 & 0.562 & 0.896 & 1.000 \\
          & \texttt{HSR} & 0.444 & 0.669 & 0.850 & 0.975 \\
          \cmidrule{2-6}
          &  Overall & 0.355 & 0.529 & 0.734 & 0.875 \\
          \midrule
          \multirow{8}{*}{$\mathcal{O}^{\mathcal{A}}$} & \texttt{RH078}    & 0.212     & 0.356 & 0.641 & 0.781 \\ 
          & \texttt{RR080}    & 0.026     & 0.123 & 0.403 & 0.549 \\ 
          & \texttt{DR080}    & 0.305     & 0.424 & 0.609 & 0.778 \\ 
          & \texttt{RH201}    & 0.374     & 0.480 & 0.758 & 0.935 \\ 
          & \texttt{MHZH}     & 0.287     & 0.543 & 0.667 & 0.782 \\ 
          & \texttt{MHZH-EXO} & 0.271     & 0.396 & 0.583 & 0.812 \\ 
          & \texttt{HSR}      & 0.531   & 0.662 & 0.769 & 0.875 \\ 
          \cmidrule{2-6}
          &  Overall & 0.287 & 0.426 & 0.633 & 0.787 \\
        \bottomrule
\end{tabular}
\footnotesize
\end{threeparttable}
\label{tab:scene_results_part_semantics}
\color{black}
\end{table}

\subsubsection{Contained Objects}
Next, we provide additional details on the evaluation of retrieved child objects contained within articulated objects. This includes two types of objects: \texttt{ARTICULATED} child objects exhibit the same motion pattern as their articulated parents (\eg, milk carton in fridge door), while the pose of \texttt{STATIC} child objects is independent of their articulated parent (\eg, jar behind cabinet door). We reflect this discrimination when evaluating contained objects in the following and follow the above evaluation routine by computing the IoU as well as the recall. The logic behind this is that many potential object parts observed could constitute a child object, which is task dependent. A simple example of this are multiple shelves contained in the fridge, which are natural child objects wrt. the fridge. Nonetheless, for our work, we aim to discover distinct contained objects such as the milk carton, the marmalade jar, or the butter, for instance.

First of all, we assign detected child objects $c$ to ground truth interaction segments by assessing whether the timestamp of observation lies within the ground truth interaction segment of the articulated parent. Once assigned, each child $c$ is matched to the nearest ground truth child $g \in \mathcal{C}_{parent}$ by maximizing the voxelized point cloud-based interaction over union:
\begin{equation}
    \text{IoU}_{max}(c) = \max_{g \in \mathcal{C}_{parent}} \left( \frac{P_c \cap P_g}{P_c \cup P_g} \right)
\end{equation}
at a resolution of $1.5\text{cm}$, which yields the IoU as reported in \cref{tab:articulated_parts}. Given this assignment, we also report the recall of found child objects above an IoU threshold of $0.25$. Finally, we measure whether the containment type (\texttt{STATIC} vs. \texttt{ARTICULATED}) for each associated child is predicted correctly in terms of accuracy. Results are aggregated scene-wise and averaged across the entire dataset.

\clearpage
\subsection{Additional Experiments}

\subsubsection{Interaction Segmentation}
\label{sec:interaction_segmentation_eval_supp}
We provide interaction segmentation results in \cref{tab:interaction_segmentation}. The obtained results indicate that \ours{} greatly outperforms both ArtiPoint and Pandora in terms of 1D-IoU and segment-based IoU. Furthermore, we find that a simple Hidden Markov Model (HMM) achieves high precision but suffers from low recalls. Ultimately, this signifies that the fusion of two complementary signals, depth disparity and interaction prior, using the proposed conditional probability model in \cref{sec:interaction_disc_supp} provides greater robustness to noise. The obtained interaction segments are used as input to the articulation estimation and articulated part segmentation evaluations in \cref{tab:articulation_pred_seg} and \cref{tab:scene_results_part_seg_pred_seg}, respectively.

\begin{table}
\color{black}
\scriptsize
\caption{Temporal interaction segmentation evaluation on Arti4D-Semantic: The 1D-IoU metric denotes the intersection over union of the dense binary timeseries signals, P and R, as well as the Segment-IoU denote the precision and recall, as well as the segment-wise IoU, following a linear sum assignment between predicted and ground truth segments under an IoU $>$ 0.5. $I_{\text{on}}$ and $I_{\text{off}}$ measure the absolute start and end errors of matched segments.}
\centering
\setlength\tabcolsep{12pt}
\begin{threeparttable}
\begin{tabular}{l|cccccc}
      \toprule
        Method & 1D-IoU & P & R & Segment-IoU & $I_{\text{on}}$ [\si{\second}] & $I_{\text{off}}$ [\si{\second}] \\
        \midrule
         GPT-5-mini & 0.55 & 0.862 & 0.6111 & 0.753 & 8.179 & 10.12 \\
         \greyrule
         Pandora~\cite{yu25pandora} & 0.359 & 0.690 & 0.400 & 0.689 & \textbf{0.7442} & \textbf{0.725} \\
         HMM & 0.471 & \textbf{0.871} & 0.532 & 0.683 & \underline{0.770} & 0.762 \\  
         ArtiPoint~\cite{arti25werby} & \underline{0.575} & 0.678 & \underline{0.714} & \underline{0.698} & 1.05 & 0.558 \\ 
         MoMa-SG & \textbf{0.649} & \underline{0.786} & \textbf{0.800} & \textbf{0.718} & 0.816 & \underline{0.742} \\
      \bottomrule
\end{tabular}
\footnotesize
\end{threeparttable}
\label{tab:interaction_segmentation}
\color{black}
\end{table}

\begin{table}
\centering
\scriptsize
\caption{Real-world articulation state estimation: Experiments are conducted across 54 various object articulation states.}
\setlength\tabcolsep{13.6pt}
\begin{threeparttable}
\begin{tabular}{l|cccc}
      \toprule
        Object & \multicolumn{2}{c}{Trans. Error} & \multicolumn{2}{c}{Rot. Error}  \\
        Type & $\Delta d$ [\unit{m}] & $\sigma$ [\unit{m}] & $\Delta \theta$ [\unit{\degree}] & $\sigma$ [\unit{\degree}] \\
        \midrule
        Prismatic & 0.0174 & 0.006 & - & - \\
        Revolute & - & - & 6.477 & 3.604 \\
      \bottomrule
\end{tabular}
\footnotesize
\end{threeparttable}
\label{tab:real_world_articulation_state_estimation}
\color{black}
\end{table}

\subsubsection{Articulation Estimation}
\label{sec:articulation_eval_supp}

\begin{table*}
\color{black}
\centering
\scriptsize
\caption{Articulated object estimation on Arti4D across methods under predicted interaction segments: Our evaluation matched predicted articulations against ground truth articulations if a 1D-IoU of at least 0.25 is exceeded. We report the axis-angle errors $\theta_{err}$, the positional errors $d_{L2}$ of revolute joints, and the accuracy of predicted joint types, including their respective recalls as in \cref{tab:articulation_estimation}.}
\setlength\tabcolsep{6.5pt}
\begin{threeparttable}
\begin{tabular}{l|cccccccc}
      \toprule
          \multirow{2}{*}{Method} & \multicolumn{2}{c}{Prismatic joints} & \multicolumn{2}{c}{Revolute joints} & Failure & \multicolumn{2}{c}{Joint Type [\%]} & Mode \\
 & $\theta_{err}$[deg] & $d_{L2}$[m] & $\theta_{err}$[deg] & $d_{L2}$[m] & Rate [\%] & Accuracy & Pris./Rev. Recall & Acc.[\%]\\
        \midrule
           Pandora~\cite{yu25pandora} & {46.31} & -- & {50.23} & \underline{0.182} & \textbf{00.0} & {55.8} & {45.0}\,/\,{75.9} & - \\
           ArtiPoint~\cite{arti25werby} & \underline{24.88} & -- & \underline{30.24} & {0.293} & {23.2} & \underline{77.4} & {88.2}\,/\,{72.2} & - \\
            MoMa-SG (ours) & \textbf{20.66} & -- & \textbf{17.31} & \textbf{0.071} & \underline{10.5} & \textbf{82.0} & {73.8}\,/\,{98.4} & \textbf{80.7} \\
        \bottomrule
\end{tabular}
\footnotesize
\end{threeparttable}
\label{tab:articulation_pred_seg}
\color{black}
\end{table*}

\begin{figure}[htbp]
    \centering
    \includegraphics[width=1.0\linewidth]{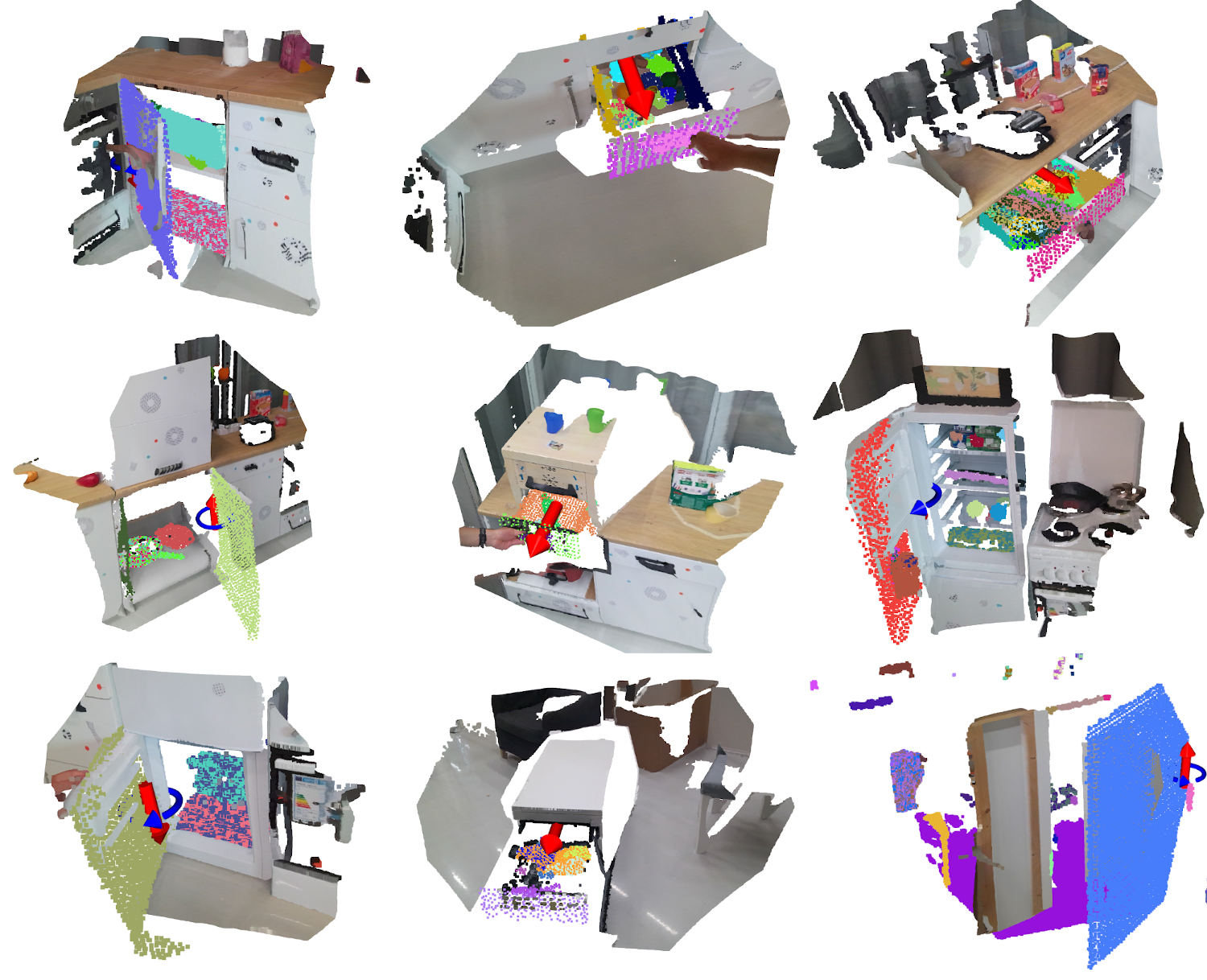}
    \caption{Contained object discovery given the articulated 3D parent objects.}
    \label{fig:contained_object_discovery}
\end{figure}

\begin{table*}
\color{black}
\centering
\scriptsize
\caption{Articulated object estimation on Arti4D across dataset splits: We report the axis-angle errors $\theta_{err}$, the positional errors $d_{L2}$ of revolute joints, and the accuracy of predicted joint types, including their respective recalls as in \cref{tab:articulation_estimation}.}
\setlength\tabcolsep{3pt}
\begin{threeparttable}
\begin{tabular}{lcc|cccccccc}
      \toprule
          \multirow{2}{*}{Split} & \multirow{2}{*}{Paradigm} & \multirow{2}{*}{GT Poses} & \multicolumn{2}{c}{Prismatic joints} & \multicolumn{2}{c}{Revolute joints} & \multicolumn{3}{c}{Type Quality [\%]} & Mode \\
 & & & $\theta_{err}$[deg] & $d_{L2}$[m] & $\theta_{err}$[deg] & $d_{L2}$[m] & Accuracy & Prismatic Recall & Revolute Recall & Acc.[\%]\\
        \midrule
          \texttt{RH078} & \textit{EGO} & \checkmark & 17.58 & -- & 06.42 & 0.035 & 0.794 & 0.667 & 1.000 & 0.825 \\
          \texttt{RR080} & \textit{EGO} & \checkmark & 29.84 & -- & 32.76 & 0.118 & 0.657 & 0.626 & 1.000 & 0.879 \\
          \texttt{DR080} & \textit{EGO} & \checkmark & 16.32 & -- & 21.57 & 0.062 & 0.889 & 1.000 & 0.816 & 0.815 \\
          \texttt{RH201} & \textit{EGO} & \checkmark & 12.94 & -- & 17.34 & 0.085 & 0.919 & 0.877 & 0.966 & 0.734 \\
          \midrule 
           & Overall & & \textbf{20.66} & -- & \textbf{17.31} & \textbf{0.071} & \textbf{0.820} & \textbf{0.738} & \textbf{0.984} & \textbf{0.807} \\
        \bottomrule
\end{tabular}
\footnotesize
\end{threeparttable}
\label{tab:articulation_pred_seg_splits}
\color{black}
\end{table*}

In the following, we expand on the results provided in the main manuscript and provide articulation estimation results using predicted interaction segments in \cref{tab:articulation_pred_seg}. We find that the obtained results are generally in line with the results obtained using GT interaction segments, albeit with a slight decrease in performance across both ArtiPoint and \ours{}, while Pandora is comparably stable. This is expected given the interaction segmentation performance as outlined in \cref{tab:interaction_segmentation}. Furthermore, \ours{} outperforms both ArtiPoint and Pandora in terms of axis prediction results by a considerable margin. Since we rely on predicted interactions for \cref{tab:articulation_pred_seg}, we also list a failure rate, which indicates the sum of missed object interactions under a 1D-IoU of 0.25 and non-successful articulation estimates. 

\begin{figure}[htbp]
\centering
\footnotesize
\includegraphics[trim=0cm 0cm 0cm 0cm, clip, width=\linewidth]{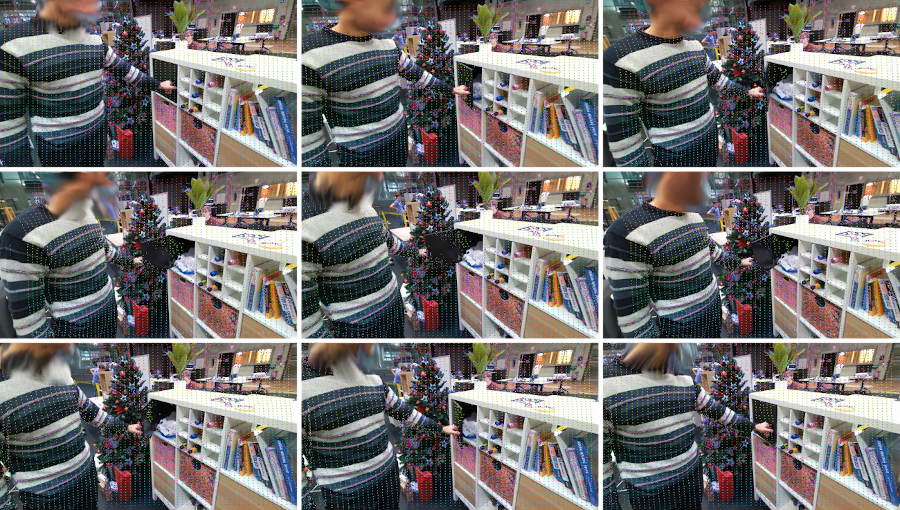} 
\caption{Tracked keypoints of TapNext++~\cite{tapnextpp2026} on the \texttt{MHZH-EXO} split (exo-centric) of \ourdataset{} along an interaction (left to right, top to bottom).} 
\label{fig:cotracker_mhzh_exo}
\end{figure}

In addition, we provide split-wise articulated object estimation results in \cref{tab:articulation_scene_results}. While we observe roughly similar errors on the original splits of Arti4D, we also note considerably higher errors on the newly introduced splits of \ourdataset{}, which we interpret in the following. In general, both \texttt{MHZH} and \texttt{MHZH-EXO} do not provide ground truth camera poses. As a consequence, performing depth lifting will lead to considerable errors. Furthermore, we generally expect the exo-centric splits to be more difficult as the observation paradigm induces more dynamic motion due to humans close-by, which we visualize in \cref{fig:cotracker_mhzh_exo}. This is further exacerbated by the number of objects that are quite texture-less as depicted in \cref{fig:cotracker_mhzh}. As a result, we do not obtain a significant number of keypoint trajectories required for accurate articulation estimation. Regarding the \textit{HSR} split we note significantly higher depth noise, which similarly increases articulation estimation errors.

Similar to \cref{tab:articulation_scene_results}, we list split-wise results under predicted interaction segments in \cref{tab:articulation_pred_seg_splits}.

Finally, we display error uncertainties in Tab.~\ref{tab:errors_articulation}, indicating that \ours{} yields smaller mean errors and standard deviations compared to ArtiPoint \& Pandora. However, the error distribution is highly skewed, i.e., not normally distributed, due to the underlying type understanding problem. Thus, we also display \textit{\{50,75,90\}\%} percentile error scores in Tab.~\ref{tab:errors_articulation}, demonstrating improved robustness. 

A Wilcoxon signed-rank significance test between MoMa-SG and ArtiPoint/Pandora confirms that predictions are significant in favor of MoMa-SG at $\alpha = 0.05$ with $p<0.0001$ in 5 of 6 tests and $p = 0.02$ for revolute position errors vs.\, ArtiPoint. 

\begin{table}[h]
\centering
\scriptsize
\caption{Error variance and percentiles of articulated axes on Arti4D}
\setlength\tabcolsep{7pt}
\begin{threeparttable}
\begin{tabular}{l|lll}
      \toprule
          \multirow{2}{*}{Method} & \multicolumn{1}{l}{Prismatic joints} & \multicolumn{2}{c}{Revolute joints}  \\
  & \multicolumn{1}{l}{$\theta_{err}$[deg]$\pm \sigma$} & \multicolumn{1}{l}{$\theta_{err}$[deg]$\pm \sigma$ }  & \multicolumn{1}{l}{$d_{L2}$[m]$\pm \sigma$} \\
        \midrule
           \multirow{1}{*}{Pandora} & 46.81$\pm$37.49  & {50.62}$\pm$28.58 & \underline{0.20}$\pm$0.18 \\
            & \textit{\{47.31, 85.58, 88.43\}} & \textit{\{51.72, 79.69, 86.34\}} & {\textit{\{0.11, 0.31, 0.42\}}} \\[0.2em]
           \multirow{1}{*}{ArtiPoint}  & \underline{23.27}$\pm$31.67 & \underline{26.36}$\pm$25.53  & {0.25}$\pm$0.26 \\
            & \textit{\{04.85, 45.19, 81.80\}} & \textit{\{12.86, 38.38, 62.22\}} & \textit{\{0.05, 0.10, 0.31\}} \\[0.2em]
             \multirow{1}{*}{MoMa-SG} & \textbf{10.37}$\pm$\textbf{24.10} & \textbf{14.80}$\pm$\textbf{22.99}  & \textbf{0.09}$\pm$\textbf{0.15} \\
              &  \textit{\{02.56, 03.90, 14.93\}} & \textit{\{04.30, 11.21, 55.24\}} & \textit{\{0.03, 0.07, 0.27\}} \\
        \bottomrule
        
\end{tabular}
The values below the errors denote 50,75,90\% percentile error scores, respectively.
\end{threeparttable}
\label{tab:errors_articulation}
\end{table}

\begin{figure*}
\centering
\footnotesize
\setlength{\tabcolsep}{0.01cm}
\includegraphics[width=\linewidth]{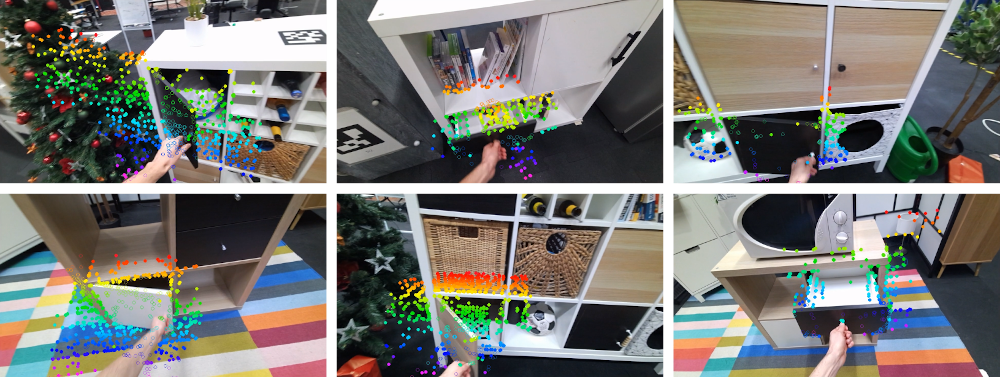}
\caption{Tracked keypoints of CoTracker3~\cite{karaev2024cotracker3} throughout interactions of texture-less objects on the \texttt{MHZH} split (ego-centric) of \ourdataset{}.} 
\label{fig:cotracker_mhzh}
\end{figure*}

\clearpage
\subsection{3D Part Mapping}
\label{sec:part_mapping_eval_supp}
In addition to the articulation estimation results, we provide results of \ours{} on 3D part segmentation under predicted interaction segments in \cref{tab:scene_results_part_seg_pred_seg}. As depicted, we do not find a performance decrease in terms of articulated 3D part segmentation, which signifies that the 3D part mapping stage is robust to errors in the interaction segmentation stage. 

Furthermore, we list split-wise results in \cref{tab:scene_results_part_seg}. Overall, we find that our method performs similarly across all considered splits. Furthermore, we have evaluated the quality of the CLIP features associated with each object in \cref{tab:scene_results_part_semantics}. We follow the evaluation routine outlined in \cref{sec:metrics_supp} and generally find a slight decrease in performance when pairing \textit{free} objects with articulations to obtain $\mathcal{O}^{\mathcal{A}}$. Overall, we observe that the correct label is the top response in 28.7\% of the cases, while it is among the top-5 in 63.3\% of cases.

\begin{table}
\color{black}
\centering
\scriptsize
\caption{Articulated 3D part segmentation performance of \ours{} on Arti4D-Semantic using predicted interaction segments. \ours{} employs TapNext++ point tracks. We report the same set of metrics as in \cref{tab:articulated_parts}.}
\setlength\tabcolsep{8pt}
\begin{threeparttable}
\begin{tabular}{lcc|ccccccc}
      \toprule
          \multirow{2}{*}{Split} & \multirow{2}{*}{Paradigm} & \multirow{2}{*}{GT Poses} & \multirow{2}{*}{IoU} & \multicolumn{3}{c}{Recall} \\
         & & & & \textit{@0.25} & \textit{@0.50} & \textit{@0.75} \\   
          \midrule
          \texttt{RH078} & \textit{EGO} & \checkmark & 0.508 & 0.692 & 0.415 & 0.134 \\
          \texttt{RR080} & \textit{EGO} & \checkmark & 0.411 & 0.605 & 0.429 & 0.146 \\
          \texttt{DR080} & \textit{EGO} & \checkmark & 0.548 & 0.889 & 0.556 & 0.111 \\
          \texttt{RH201} & \textit{EGO} & \checkmark & 0.408 & 0.674 & 0.501 & 0.124 \\
          \midrule
           & Overall & & \textbf{0.448} & \textbf{0.675} & \textbf{0.460} & \textbf{0.132} \\
        \bottomrule
\end{tabular}
\footnotesize
\end{threeparttable}
\label{tab:scene_results_part_seg_pred_seg}
\color{black}
\end{table}

We provide qualitative insights into scene-level articulated object predictions, object-level predictions, and contained object discovery in \cref{fig:full_scene_plots_arti4d_semantic}, \cref{fig:articulated_3d_objects}, and \cref{fig:contained_object_discovery}, respectively. We find that the obtained articulated 3D part segmentations and axis predictions are highly accurate given the level of occlusions and tracking drift. In most cases, the articulation state is accurately estimated. However, under strong occlusions of the articulated part, we notice errors in angle estimation for revolutes.

Regarding the contained object discovery, we find that \ours{} is able to discover child objects and correctly predict the relation to their parent object (\texttt{STATIC}/\texttt{ARTICULATED}) in most cases, \eg, contained objects in drawers in non-closed articulation states moving along with their parent or the segmented milk carton moving along with the fridge door. We also find that some child objects are not discovered, which we either attribute to faulty articulated object estimation or imprecise granularity in object segmentation.

We further obtained robustness estimates and note the following articulated part IoU standard deviations: MoMa-SG ($0.434 \pm 0.261$), MoMa-SG w/ Mobile-SAM ($0.417 \pm 0.270$), MoMa-SG w/ greedy matching ($0.423 \pm 0.265$), and Pandora ($0.065 \pm 0.061$).

\subsection{Real-World Long-Horizon Mobile Manipulation}
\label{sec:real_world_long_horizon_moma}

As discussed in the manuscript, we investigate to which degree \ours{} allows to perform state estimation of various articulated objects. Overall, we find that estimating the configuration state of prismatics works well while we observe slightly higher state estimation errors on revolute objects, as displayed in \cref{tab:real_world_articulation_state_estimation}.

We provide additional qualitative insights into the performance of \ours{} in long-horizon mobile manipulation. First, we demonstrate that \ours{} is capable to perform 7 consecutive object manipulations in a row as visualized in \cref{fig:opening}. We find that the predominant error source when performing long-horizon mobile manipulation are failed grasps, \ie, loss of grip while manipulating. Occasionally, we find that handles are wrongly associated with close-by objects when the object to be manipulated is already opened. In addition, we depict consecutive long-horizon opening trials using the Boston Dynamics Spot quadruped in \cref{fig:opening_spot}.

We obtain similar long-horizon results on the task of object closing, which is depicted in the supplementary video. We consider the task of closing to be more complicated, as one needs to first infer the precise articulation state of the object to be manipulated, while the robot needs to be positioned well to accomplish this. Furthermore, detecting handles is more challenging for the closing vs. the opening task. 

Finally, we also show how \ours{} is tightly integrated with natural language prompts as depicted in \cref{fig:retrieve}, where the task is to "get milk from the fridge". 
The executed high-level actions (see \cref{sec:moma_language}) for this trial were: \texttt{open(fridge)}, which involves navigating to an inspection position based on the object normal of the object to be manipulated, registration to perform fine-grained localization (see \cref{sec:moma_supp}), and the sole opening action from sampled end-effector trajectories (see \cref{sec:ee_traj_sampling}); \texttt{inspect(fridge)} from the previous inspection position to look inside of the fridge; \texttt{retrieve(milk, fridge)} to estimate a grasp for the prompted category "milk" and to execute an end-effector trajectory that takes the milk carton from the fridge along the approach axis of the grasp. In general, the retrieval trajectory is interchangeable with another more acceptable motion.

\subsubsection{Robustness Study}
\label{sec:eval_robustness_study_supp}
\begin{figure}[htbp]
\vspace{-0.3cm}
\centering
\footnotesize
\includegraphics[trim=0 0cm 0 0cm, clip, width=\linewidth]{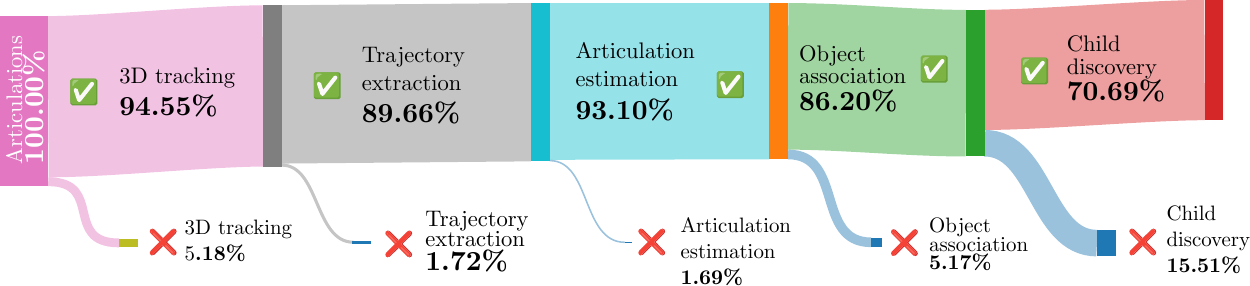}
\caption{Robustness study across the stages of \ours{} on the RH201 split of the Arti4D-Semantic dataset.} 
\label{fig:robustness_study_sankey}
\vspace{-0.3cm}
\end{figure}

We conducted a systematic analysis of the robustness of different components of \ours{} on the RH201 split of \ourdataset{}, which is displayed in \cref{fig:robustness_study_sankey}. The dominant error sources are specular objects causing failed 3D tracking, faulty trajectory filtering, object-articulation mismatches, and erroneous articulation magnitudes, in turn complicating child discovery. We attribute the high failure rate in child discovery to imperfect articulation estimation and articulation-object association. Independent of this, we hold matchable parts in 94.8\% and predict true articulation modes in 99.3\% of cases.

Furthermore, we investigate how sensitive our estimator is to variations in $\eta$ in \cref{fig:twist_eta_study}. $\eta$ is the main hyperparameter that dictates type discrimination as outlined in \cref{sec:articulation_estimation}.

\begin{figure}[h]

\resizebox{\linewidth}{!}{
\begin{tikzpicture}
 
\begin{axis}[
    myplotstyle,
    at={(0,0)},
    ylabel={Type Acc [\%]},
    xlabel={$\eta$},
    ylabel style={at={(axis description cs:-0.2,0.5)}, anchor=center},
    ymin=40, ymax=100,
    xmin=0.985, xmax=1.0,
    legend pos=north west,
    scaled x ticks=false,
    xticklabel style={/pgf/number format/fixed, /pgf/number format/precision=3},
    xtick={0.985,0.987,0.989,0.991,0.993,0.995,0.997,0.999},
    ytick={20,40,60,80,100},
    title={Type prediction accuracy},
]
\addplot[orange!90!black, mark=*, mark size=1pt] coordinates {
(0.986388,66.8966) (0.992189,81.3793) (0.993833,86.8966) (0.994871,92.4138) (0.995424,94.4828)
(0.995940,95.8621) (0.996233,96.5517) (0.996466,96.5517) (0.996649,96.5517) (0.996870,96.5517)
(0.997233,95.1724) (0.997591,96.5517) (0.997981,93.7931) (0.998374,92.4138) (0.998653,90.3448)
(0.998885,87.5862) (0.999136,85.5172) (0.999284,84.1379) (0.999368,81.3793) (0.999660,60.6897)
(0.999668,60.6897) (0.999670,60.6897) (0.999940,51.0345)
};
\addplot[black, dashed, forget plot] coordinates {(0.9969, 0) (0.9969, 100)};
\addlegendentry{Type Acc}
\end{axis}
 
\begin{axis}[
    myplotstyle,
    at={(5.2cm,0)},
    ylabel={$\theta_{\mathrm{pris}}$ [$^{\circ}$]},
    xlabel={$\eta$},
    ylabel style={at={(axis description cs:-0.2,0.5)}, anchor=center},
    legend pos=north west,
    ymin=0, ymax=100,
    xmin=0.985, xmax=1.0,
    scaled x ticks=false,
    xticklabel style={/pgf/number format/fixed, /pgf/number format/precision=3},
    xtick={0.985,0.987,0.989,0.991,0.993,0.995,0.997,0.999},
    ytick={20,40,60,80,100},
    title={Rotational error (prismatic)},
]
\addplot[red!40, draw=none, forget plot, name path=tp_upper] coordinates {
(0.986388,6.9940) (0.992189,7.1223) (0.993833,14.2505) (0.994871,14.2556) (0.995424,14.2556)
(0.995940,14.2545) (0.996233,14.2556) (0.996466,14.2470) (0.996649,14.2969) (0.996870,14.7856)
(0.997233,26.8801) (0.997591,26.7334) (0.997981,38.9975) (0.998374,46.5081) (0.998653,53.2473)
(0.998885,59.9719) (0.999136,67.5079) (0.999284,71.5653) (0.999368,77.3493) (0.999660,100.0000)
(0.999668,100.0000) (0.999670,100.0000) (0.999940,95.5877)
};
\addplot[red!40, draw=none, forget plot, name path=tp_lower] coordinates {
(0.986388,0.0000) (0.992189,0.0000) (0.993833,0.0000) (0.994871,0.0000) (0.995424,0.0000)
(0.995940,0.0000) (0.996233,0.0000) (0.996466,0.0000) (0.996649,0.0000) (0.996870,0.0000)
(0.997233,0.0000) (0.997591,0.0000) (0.997981,0.0000) (0.998374,0.0000) (0.998653,0.0000)
(0.998885,0.0000) (0.999136,0.0000) (0.999284,1.4887) (0.999368,5.6178) (0.999660,44.7467)
(0.999668,44.5379) (0.999670,44.5287) (0.999940,77.1293)
};
\addplot[red!25, fill opacity=0.5, forget plot]
    fill between[of=tp_upper and tp_lower];
\addplot[red!80!black, mark=*, mark size=1pt] coordinates {
(0.986388,3.4096) (0.992189,3.4267) (0.993833,4.2406) (0.994871,4.2412) (0.995424,4.2412)
(0.995940,4.2408) (0.996233,4.2417) (0.996466,4.2428) (0.996649,4.2965) (0.996870,4.6077)
(0.997233,7.5571) (0.997591,7.9157) (0.997981,12.8874) (0.998374,16.3477) (0.998653,19.9722)
(0.998885,24.6044) (0.999136,31.7075) (0.999284,36.5270) (0.999368,41.4836) (0.999660,72.4403)
(0.999668,72.3074) (0.999670,72.3677) (0.999940,86.3585)
};
\addplot[black, dashed, forget plot] coordinates {(0.9969, 0) (0.9969, 120)};
\addlegendentry{$\theta_{\mathrm{pris}}$ $\pm\,\sigma$}
\end{axis}
 
\begin{axis}[
    myplotstyle,
    at={(10.4cm,0)},
    ylabel={$\theta_{\mathrm{rev}}$ [$^{\circ}$]},
    ylabel style={at={(axis description cs:-0.2,0.5)}, anchor=center},
    xlabel={$\eta$},
    legend pos=north west,
    ymin=0, ymax=110,
    xmin=0.985, xmax=1.0,
    scaled x ticks=false,
    xticklabel style={/pgf/number format/fixed, /pgf/number format/precision=3},
    xtick={0.985,0.987,0.989,0.991,0.993,0.995,0.997,0.999},
    ytick={20,40,60,80,100},
    title={Rotational error (revolute)},
]
\addplot[teal!40, draw=none, forget plot, name path=tr_upper] coordinates {
(0.986388,98.6364) (0.992189,79.9449) (0.993833,69.5302) (0.994871,61.9969) (0.995424,53.6139)
(0.995940,48.9713) (0.996233,46.9639) (0.996466,47.0595) (0.996649,48.9424) (0.996870,43.9472)
(0.997233,42.9044) (0.997591,40.6902) (0.997981,39.4833) (0.998374,37.0060) (0.998653,36.4830)
(0.998885,36.6613) (0.999136,36.0838) (0.999284,36.0816) (0.999368,36.0489) (0.999660,36.0984)
(0.999668,36.0823) (0.999670,36.0820) (0.999940,36.0811)
};
\addplot[teal!40, draw=none, forget plot, name path=tr_lower] coordinates {
(0.986388,33.7132) (0.992189,9.2343) (0.993833,0.2549) (0.994871,0.0000) (0.995424,0.0000)
(0.995940,0.0000) (0.996233,0.0000) (0.996466,0.0000) (0.996649,0.0000) (0.996870,0.0000)
(0.997233,0.0000) (0.997591,0.0000) (0.997981,0.0000) (0.998374,0.0000) (0.998653,0.0000)
(0.998885,0.0000) (0.999136,0.0000) (0.999284,0.0000) (0.999368,0.0000) (0.999660,0.0000)
(0.999668,0.0000) (0.999670,0.0000) (0.999940,0.0000)
};
\addplot[teal!30, fill opacity=0.5, forget plot]
    fill between[of=tr_upper and tr_lower];
\addplot[teal, mark=*, mark size=1pt] coordinates {
(0.986388,66.1748) (0.992189,44.5896) (0.993833,34.8926) (0.994871,30.5524) (0.995424,25.4798)
(0.995940,22.5765) (0.996233,21.8136) (0.996466,21.7822) (0.996649,22.5175) (0.996870,20.3651)
(0.997233,20.0271) (0.997591,19.1865) (0.997981,18.5649) (0.998374,17.3795) (0.998653,17.0708)
(0.998885,17.1724) (0.999136,16.6716) (0.999284,16.6588) (0.999368,16.6334) (0.999660,16.6722)
(0.999668,16.6602) (0.999670,16.6600) (0.999940,16.6589)
};
\addplot[black, dashed, forget plot] coordinates {(0.9969, 0) (0.9969, 120)};
\addlegendentry{$\theta_{\mathrm{rev}}$ $\pm\,\sigma$}
\end{axis}
 
\begin{axis}[
    myplotstyle,
    at={(15.6cm,0)},
    ylabel={$d$ [\unit{\meter}]},
    xlabel={$\eta$},
    ylabel style={at={(axis description cs:-0.2,0.5)}, anchor=center},
    legend pos=north west,
    ymin=0, ymax=0.55,
    xmin=0.985, xmax=1.0,
    scaled x ticks=false,
    xticklabel style={/pgf/number format/fixed, /pgf/number format/precision=3},
    xtick={0.985,0.987,0.989,0.991,0.993,0.995,0.997,0.999},
    ytick={0.1,0.2,0.3,0.4,0.5},
    title={Translational error (revolute)},
]
\addplot[blue!30, draw=none, forget plot, name path=d_upper] coordinates {
(0.986388,0.4204) (0.992189,0.3446) (0.993833,0.3038) (0.994871,0.3053) (0.995424,0.1482)
(0.995940,0.1900) (0.996233,0.1303) (0.996466,0.1231) (0.996649,0.1137) (0.996870,0.1144)
(0.997233,0.1219) (0.997591,0.4585) (0.997981,0.3697) (0.998374,0.4198) (0.998653,0.1133)
(0.998885,0.0954) (0.999136,0.0955) (0.999284,0.0956) (0.999368,0.0971) (0.999660,0.3104)
(0.999668,0.2897) (0.999670,0.2895) (0.999940,0.2885)
};
\addplot[blue!30, draw=none, forget plot, name path=d_lower] coordinates {
(0.986388,0.0000) (0.992189,0.0000) (0.993833,0.0000) (0.994871,0.0000) (0.995424,0.0000)
(0.995940,0.0000) (0.996233,0.0000) (0.996466,0.0000) (0.996649,0.0000) (0.996870,0.0000)
(0.997233,0.0000) (0.997591,0.0000) (0.997981,0.0000) (0.998374,0.0000) (0.998653,0.0000)
(0.998885,0.0000) (0.999136,0.0000) (0.999284,0.0000) (0.999368,0.0000) (0.999660,0.0000)
(0.999668,0.0000) (0.999670,0.0000) (0.999940,0.0000)
};
\addplot[blue!25, fill opacity=0.5, forget plot]
    fill between[of=d_upper and d_lower];
\addplot[blue, mark=*, mark size=1pt] coordinates {
(0.986388,0.1657) (0.992189,0.1147) (0.993833,0.0888) (0.994871,0.0829) (0.995424,0.0450)
(0.995940,0.0482) (0.996233,0.0381) (0.996466,0.0358) (0.996649,0.0329) (0.996870,0.0330)
(0.997233,0.0342) (0.997591,0.0628) (0.997981,0.0558) (0.998374,0.0614) (0.998653,0.0326)
(0.998885,0.0276) (0.999136,0.0277) (0.999284,0.0278) (0.999368,0.0286) (0.999660,0.0488)
(0.999668,0.0471) (0.999670,0.0471) (0.999940,0.0471)
};
\addplot[black, dashed, forget plot] coordinates {(0.9969, 0) (0.9969, 0.55)};
\addlegendentry{$d \pm\,\sigma$}
\end{axis}
 
\end{tikzpicture}
}
\caption{Sensitivity of the proposed regularized twist estimation to various $\eta$ on the RH201 split of \ourdataset{}.}
\label{fig:twist_eta_study}
\end{figure}

\subsubsection{Runtime}
\label{sec:runtime_supp}
\ours{} constructs 3D scene graphs w/ 10 objects in 21.89~min (avg), i.e., 2~min 11s per object. This time is allocated to the following stages of the approach: data buffering (6\%), interaction segmentation (9.8\%), mapping (42.1\%), point tracking (10.1\%), trajectory filtering (12.9\%), articulation and mode estimation (9.9\%), and axis-object association (8.7\%). We conducted our experiments using an RTX A6000, 128GB RAM, and an AMD Threadripper PRO 3975WX. We recommend a minimum of 64 GB of RAM and 24GB of VRAM to run \ours{}.

\clearpage
\section{Arti4D-Semantic Dataset \& Benchmark}
\label{sec:supp_arti4d_semantic}
We present the \textit{\ourdataset{}} dataset, which, to the best of our knowledge, constitutes the first hierarchical interactive 3D scene graph dataset tailored towards scene-level articulated object understanding. Our goal is for \textit{\ourdataset{}} to connect the research areas of articulated object estimation and semantic 3D mapping, which have been studied mostly in isolation to date. \textit{\ourdataset{}} extends \textit{Arti4D}~\cite{arti25werby} by providing a rich label set of interaction segments, labeled articulated 3D parts, including semantic categories, as well as labels of objects contained within articulated objects. While ego-centric sequences represented the only observation paradigm in \textit{Arti4D}, the novel \textit{\ourdataset{}} dataset provides two additional ones: exo-centric demonstrations and robot-centric interaction observations using a teleoperated mobile manipulator. We believe the dataset will be invaluable to study the problem of constructing semantic-kinematic representations of 3D scenes under minimal assumptions about camera positions and demonstration protocol.  

\begin{figure}[htbp]
\centering
\footnotesize
\setlength{\tabcolsep}{0.01cm}
\begin{tabularx}{\linewidth}{ll}
\includegraphics[trim=0 0cm 0 0cm, clip, width=0.49\linewidth]{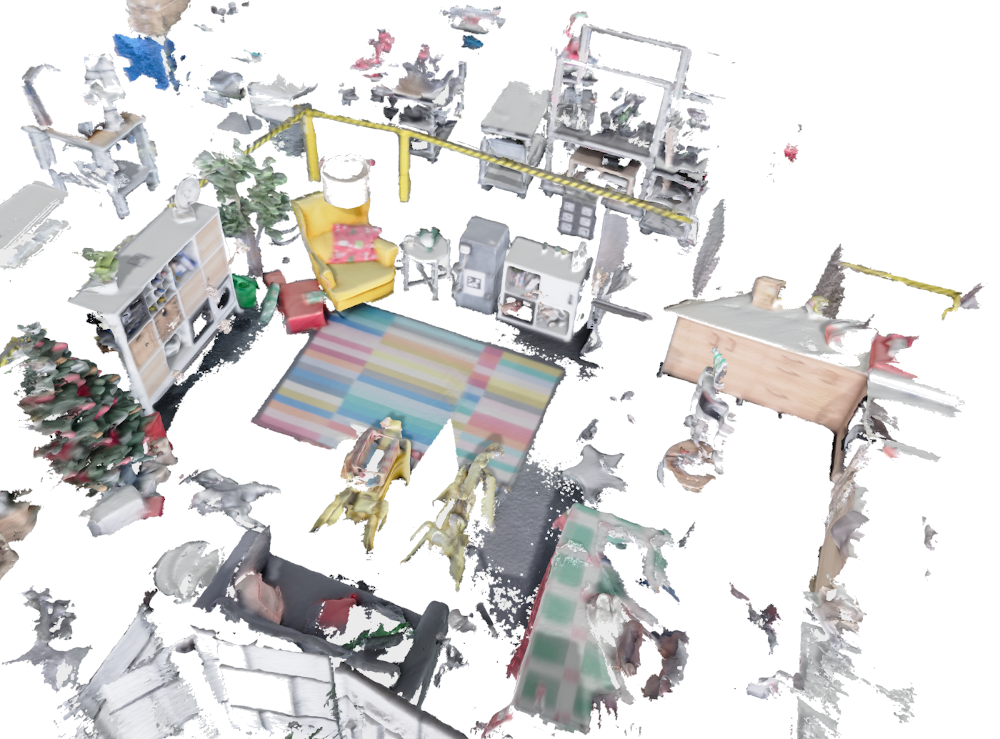} &
\includegraphics[width=0.49\linewidth]{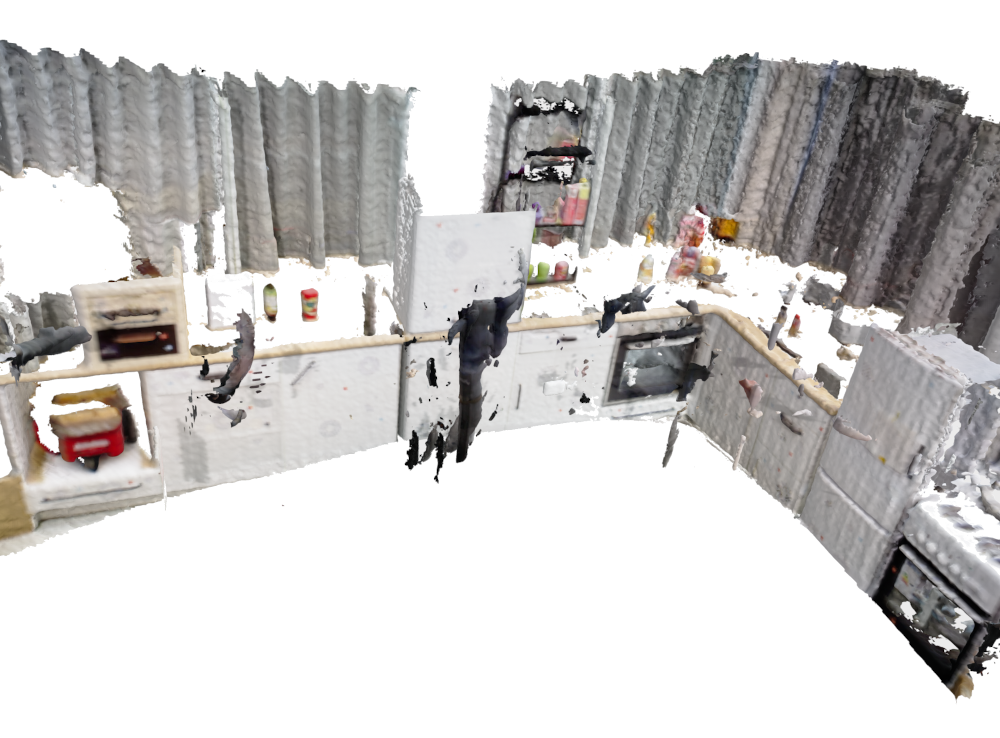} \\
\end{tabularx}
\caption{We visualize the three new dataset splits of Arti4D-Semantic: The ego-centric \texttt{MHZH} / exo-centric \texttt{MHZH-EXO} splits (left) and the robot-centric \texttt{HSR} split (right).} 
\label{fig:qualitative_arti4d_semantic}
\end{figure}

Across all observation paradigms, \textit{\ourdataset{}} contains 62 RGB-D sequences across 5 distinct scenes depicting humans demonstrating interactions with various articulated objects such as drawers, cabinets, doors, or sliding cabinets. For each interacted object, we thus provide instance and semantic information coupled with its axis of motion.
While the labeling of articulated 3D parts is conducted on static scene reconstructions, the labeling of contained objects is more involved. 
Since contained objects are either static in world coordinates or follow the motion pattern of their articulated parent, our dataset provides contained object labels that are themselves a function of the articulation state. Therefore, misidentifications of maximum-opening states are resolvable at the time of evaluation. We provide additional details in the supplementary material.

As part of our work, we do not only provide three additional dataset splits that contribute robot-centric (\texttt{HSR}) as well as exo-centric (\texttt{MHZH}/\texttt{MHZH-EXO}) demonstrations, but also provide labels for all articulated objects as well as the objects contained in them. The labels include instance segmentation of all objects and also semantic categories. We pair these new labels with the articulation axes of the interacted objects and their associated ground truth interaction segments. We visualize the three new splits of \ourdataset{} in \cref{fig:qualitative_arti4d_semantic}. We construct pseudo ground truth camera trajectories using DROID-SLAM~\cite{teed2021droid}.

\begin{figure}[htbp]
\centering
\footnotesize
\setlength{\tabcolsep}{0.01cm}
\begin{tabularx}{\linewidth}{c}
\includegraphics[trim=0 5cm 0 6cm, clip, width=0.995\linewidth]{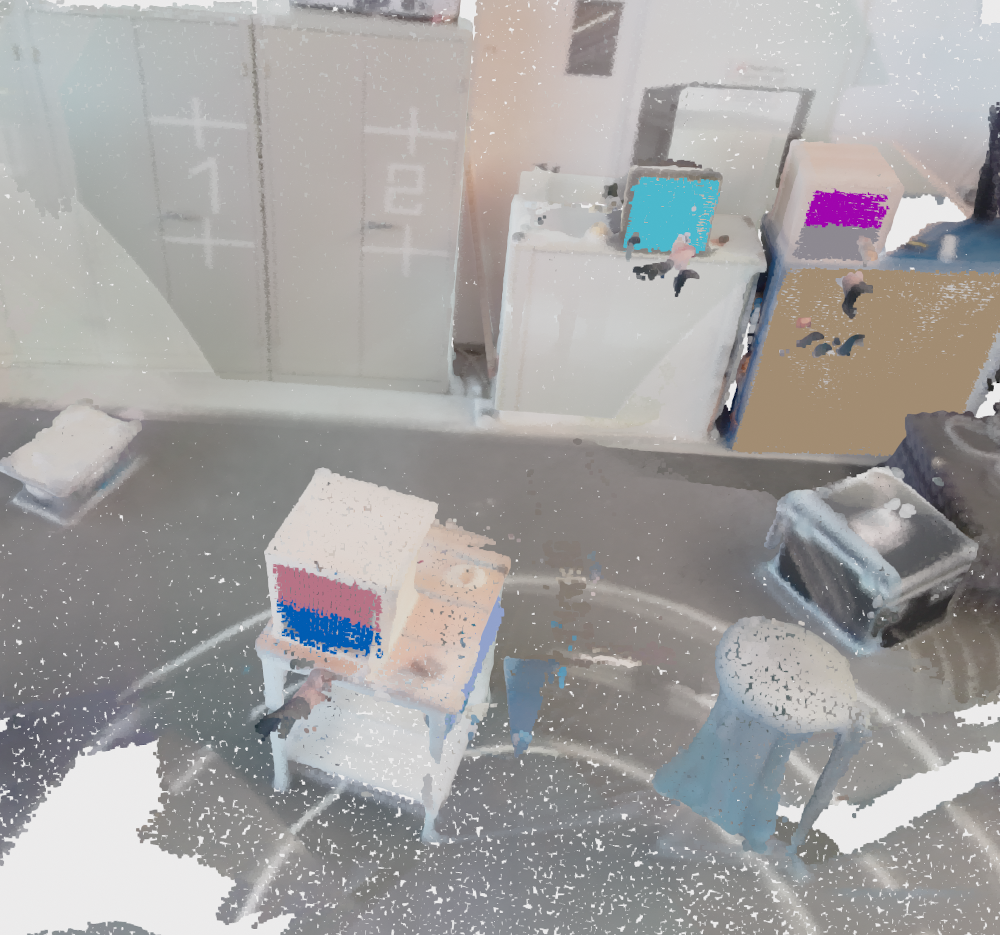} \\
\includegraphics[trim=0 10cm 0 10cm, clip, width=0.995\linewidth]{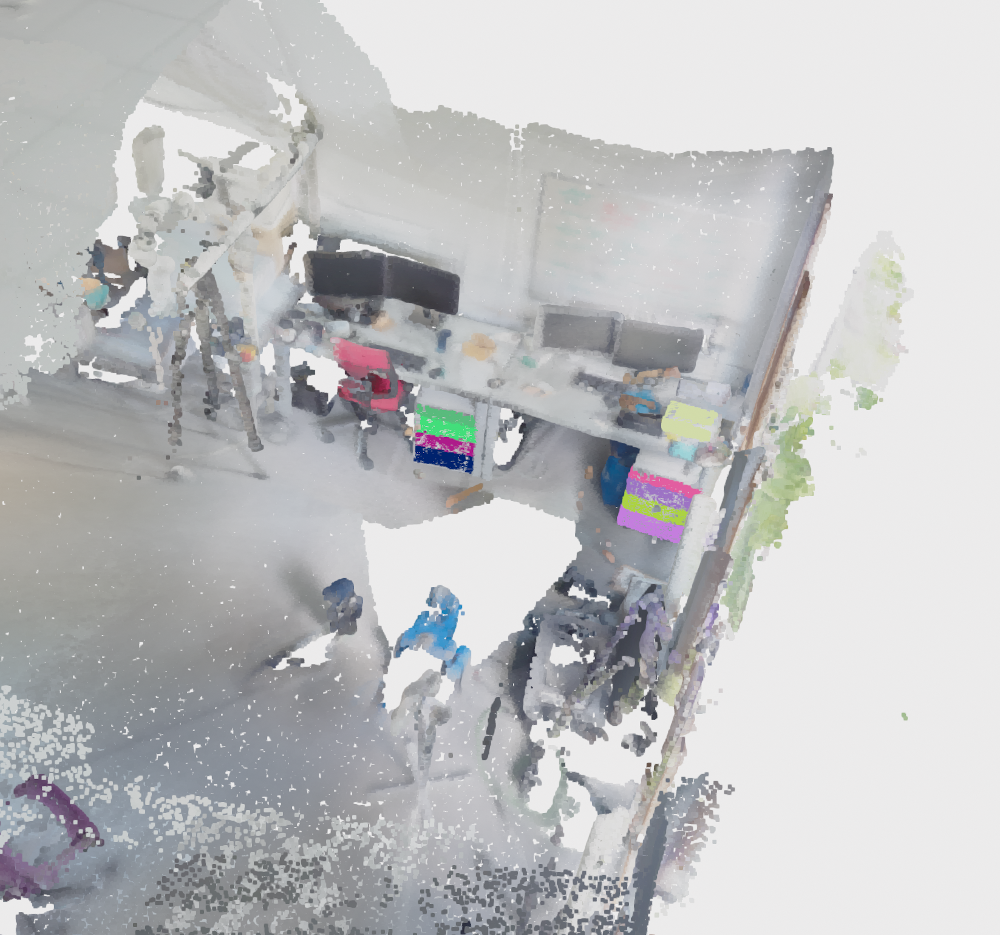} \\
\end{tabularx}
\caption{We visualize articulated 3D parent instance labels for two scenes contained in \texttt{RH078} (top) and \texttt{DIN080} (bottom), respectively. Each distinctly colored mask represents a closed articulated object.} 
\label{fig:arti4d_labels}
\end{figure}

In the following, we provide details on the involved labeling procedure. Across all articulated objects, we manually select a sufficient number of keyframes from which the articulated object is visible in its closed state. We label each articulated object given those keyframes and later fuse those labels by aggregating them in 3D world coordinates as depicted in \cref{fig:arti4d_labels}. Furthermore, we provide a semantic category to each parent object that is independently verified by another reviewer. Regarding contained objects, we select frames of maximum opening for each articulated parent. Given those frames, we label each distinct, contained object and provide it with a semantic category. In addition to that, we also label the visible parts of the articulated parent object by differentiating between areas seen in its closed state (front face) and areas seen only during articulation, which provides us with regions-of-interest that let us infer whether a child object is \texttt{STATIC} or \texttt{ARTICULATED}. Thus, whenever a child label mask overlaps with the articulated part mask during interaction, we consider it \texttt{ARTICULATED}, otherwise as \texttt{STATIC}. Overall, our dataset features 29 articulated parent categories and 54 child object categories. 

We provide an overview of all scenes contained in \ourdataset{} in \cref{tab:sequence_overview}.

\begin{table*}[ht]
    \scriptsize
    \setlength{\tabcolsep}{2pt}
    \centering
    \caption{Overview of all demonstration sequences of \ourdataset{}. The \texttt{RH078}, \texttt{RR080}, \texttt{DR080}, \texttt{RH201}, and \texttt{MHZH} splits are ego-centric. The \texttt{MHZH-EXO} split is exo-centric and the \texttt{HSR} split is robot-centric. We list the distribution of articulation types (\texttt{PRISM} / \texttt{REV}), difficulties, and the number of distinct interactions vs. the number of articulated objects.}
    \begin{threeparttable}
        \begin{tabularx}{0.99\linewidth}{cccclcccccc}
          \toprule
            \multirow{2}{*}{\textit{Split}} & \textit{GT } & \multirow{2}{*}{\textit{Paradigm}} & \textit{Sequence} & \multirow{2}{*}{\textit{Recording}} & \textit{\#} & \textit{\#} & \#  & \# & Parent & Child \\
                          & Poses & & \textit{ID}  &  & \textit{Objects} & \textit{Interactions} & \texttt{PRISM} / \texttt{REV} &  \texttt{EASY} / \texttt{HARD} & Labels & Labels \\
             \midrule
                \multirow{11}{*}{\rotatebox{90}{\texttt{RH078}}} & \checkmark & EGO & RH078-00 & scene\_2025-04-04-19-14-38 &  7  &  7  &  3 / 4  &  7 / 0  & \checkmark & \checkmark \\
                 & \checkmark & EGO & RH078-01 & scene\_2025-04-04-19-18-54 &  6  &  7  &  2 / 4  &  4 / 2  & \checkmark & \checkmark \\
                 & \checkmark & EGO & RH078-02 & scene\_2025-04-07-11-39-17 &  7  &  7  &  3 / 4  &  7 / 0  & \checkmark & \checkmark \\
                 & \checkmark & EGO & RH078-03 & scene\_2025-04-07-11-41-52 &  8  &  9  &  3 / 5  &  6 / 2  & \checkmark & \checkmark \\
                 & \checkmark & EGO & RH078-04 & scene\_2025-04-07-11-48-40 &  7  &  7  &  3 / 4  &  7 / 0  & \checkmark & \checkmark \\
                 & \checkmark & EGO & RH078-05 & scene\_2025-04-09-10-30-11 &  8  &  8  &  5 / 3  &  3 / 5  & \checkmark & \checkmark \\
                 & \checkmark & EGO & RH078-06 & scene\_2025-04-09-10-32-52 &  6  &  6  &  5 / 1  &  3 / 3  & \checkmark & \checkmark \\
                 & \checkmark & EGO & RH078-07 & scene\_2025-04-09-10-35-47 &  7  &  7  &  6 / 1  &  2 / 5  & \checkmark & \checkmark \\
                 & \checkmark & EGO & RH078-08 & scene\_2025-04-09-10-38-38 &  7  &  7  &  5 / 2  &  3 / 4  & \checkmark & \checkmark \\
                 & \checkmark & EGO & RH078-09 & scene\_2025-04-09-10-46-48 &  7  &  8  &  7 / 0  &  4 / 3  & \checkmark & \checkmark \\
                 & \checkmark & EGO & RH078-10 & scene\_2025-04-09-10-49-20 &  7  &  7  &  6 / 1  &  2 / 5  & \checkmark & \checkmark \\
        
                \greyrule
                \multirow{10}{*}{\rotatebox{90}{\texttt{RR080}}} & \checkmark & EGO & RR080-00 & scene\_2025-04-10-13-11-16 &  17  &  17  &  14 / 3  &  9 / 8  & \checkmark & \checkmark \\
                 & \checkmark & EGO & RR080-01 & scene\_2025-04-10-16-05-09 &  14  &  14  &  13 / 1  &  8 / 6  & \checkmark & \checkmark \\
                 & \checkmark & EGO & RR080-02 & scene\_2025-04-17-15-25-14 &  11  &  12  &  11 / 0  &  7 / 4  & \checkmark & \checkmark \\
                 & \checkmark & EGO & RR080-03 & scene\_2025-04-17-15-33-44 &  9  &  9  &  9 / 0  &  7 / 2  & \checkmark & \checkmark \\
                 & \checkmark & EGO & RR080-04 & scene\_2025-04-22-09-53-49 &  8  &  8  &  8 / 0  &  5 / 3  & \checkmark & \checkmark \\
                 & \checkmark & EGO & RR080-05 & scene\_2025-04-22-09-56-24 &  10  &  10  &  9 / 1  &  9 / 1  & \checkmark & \checkmark \\
                 & \checkmark & EGO & RR080-06 & scene\_2025-04-22-09-58-49 &  7  &  8  &  7 / 0  &  4 / 3  & \checkmark & \checkmark \\
                 & \checkmark & EGO & RR080-07 & scene\_2025-04-22-11-45-15 &  9  &  9  &  8 / 1  &  7 / 2  & \checkmark & \checkmark \\
                 & \checkmark & EGO & RR080-08 & scene\_2025-04-22-11-48-01 &  9  &  9  &  8 / 1  &  7 / 2  & \checkmark & \checkmark \\
                 & \checkmark & EGO & RR080-09 & scene\_2025-04-22-11-50-40 &  8  &  8  &  7 / 1  &  6 / 2  & \checkmark & \checkmark \\

                 \greyrule
                \multirow{8}{*}{\rotatebox{90}{\texttt{DR080}}} & \checkmark & EGO & DR080-00 & scene\_2025-04-11-11-44-32 &  11  &  11  &  7 / 4  &  5 / 6  & \checkmark & \checkmark \\
                 & \checkmark & EGO & DR080-01 & scene\_2025-04-11-12-58-58 &  10  &  10  &  6 / 4  &  5 / 5  & \checkmark & \checkmark \\
                 & \checkmark & EGO & DR080-02 & scene\_2025-04-11-13-01-59 &  9  &  9  &  5 / 4  &  4 / 5  & \checkmark & \checkmark \\
                 & \checkmark & EGO & DR080-03 & scene\_2025-04-11-13-18-00 &  9  &  10  &  4 / 5  &  3 / 6  & \checkmark & \checkmark \\
                 & \checkmark & EGO & DR080-04 & scene\_2025-04-11-13-43-03 &  11  &  11  &  7 / 4  &  5 / 6  & \checkmark & \checkmark \\
                 & \checkmark & EGO & DR080-05 & scene\_2025-04-11-14-01-06 &  11  &  11  &  7 / 4  &  5 / 6  & \checkmark & \checkmark \\
                 & \checkmark & EGO & DR080-06 & scene\_2025-04-11-15-43-24 &  11  &  12  &  7 / 4  &  5 / 6  & \checkmark & \checkmark \\
                 & \checkmark & EGO & DR080-07 & scene\_2025-04-11-15-46-48 &  11  &  11  &  6 / 5  &  4 / 7  & \checkmark & \checkmark \\

                \greyrule
                \multirow{16}{*}{\rotatebox{90}{\texttt{RH201}}} & \checkmark & EGO & RH201-00 & scene\_2025-04-24-17-52-21 &  11  &  11  &  5 / 6  &  6 / 5  & \checkmark & \checkmark \\
                 & \checkmark & EGO & RH201-01 & scene\_2025-04-24-17-54-13 &  9  &  9  &  5 / 4  &  5 / 4  & \checkmark & \checkmark \\
                 & \checkmark & EGO & RH201-02 & scene\_2025-04-24-19-18-42 &  11  &  11  &  5 / 6  &  7 / 4  & \checkmark & \checkmark \\
                 & \checkmark & EGO & RH201-03 & scene\_2025-04-24-19-21-50 &  8  &  8  &  2 / 6  &  5 / 3  & \checkmark & \checkmark \\
                 & \checkmark & EGO & RH201-04 & scene\_2025-04-24-19-24-09 &  8  &  8  &  5 / 3  &  5 / 3  & \checkmark & \checkmark \\
                 & \checkmark & EGO & RH201-05 & scene\_2025-04-25-10-36-37 &  9  &  9  &  5 / 4  &  7 / 2  & \checkmark & \checkmark \\
                 & \checkmark & EGO & RH201-06 & scene\_2025-04-25-10-53-40 &  8  &  8  &  4 / 4  &  3 / 5  & \checkmark & \checkmark \\
                 & \checkmark & EGO & RH201-07 & scene\_2025-04-25-10-56-33 &  8  &  8  &  3 / 5  &  3 / 5  & \checkmark & \checkmark \\
                 & \checkmark & EGO & RH201-08 & scene\_2025-04-25-11-11-47 &  15  &  16  &  6 / 9  &  6 / 9  & \checkmark & \checkmark \\
                 & \checkmark & EGO & RH201-09 & scene\_2025-04-25-11-15-47 &  7  &  7  &  5 / 2  &  4 / 3  & \checkmark & \checkmark \\
                 & \checkmark & EGO & RH201-10 & scene\_2025-04-25-14-58-42 &  9  &  9  &  6 / 3  &  4 / 5  & \checkmark & \checkmark \\
                 & \checkmark & EGO & RH201-11 & scene\_2025-04-25-15-02-14 &  7  &  7  &  3 / 4  &  4 / 3  & \checkmark & \checkmark \\
                 & \checkmark & EGO & RH201-12 & scene\_2025-04-25-15-04-48 &  7  &  7  &  4 / 3  &  3 / 4  & \checkmark & \checkmark \\
                 & \checkmark & EGO & RH201-13 & scene\_2025-04-25-15-16-29 &  9  &  9  &  5 / 4  &  3 / 6  & \checkmark & \checkmark \\
                 & \checkmark & EGO & RH201-14 & scene\_2025-04-25-15-19-22 &  10  &  10  &  5 / 5  &  5 / 5  & \checkmark & \checkmark \\
                 & \checkmark & EGO & RH201-15 & scene\_2025-04-25-15-22-54 &  8  &  8  &  4 / 4  &  3 / 5  & \checkmark & \checkmark \\   

                \greyrule
                \multirow{5}{*}{\rotatebox{90}{\texttt{MHZH}}} & $\times$ & EGO & MHZH-00 & scene\_20251205\_171539 &  6  &  6  &  2 / 4  &  - / -   & \checkmark & - \\
                 & $\times$ & EGO & MHZH-01 & scene\_20251206\_152615 &  11  &  11  &  2 / 9  &   - / -  & \checkmark & - \\
                 & $\times$ & EGO & MHZH-02 & scene\_20251206\_153729 &  8  &  8  &  4 / 4 &   - / -  & \checkmark & - \\
                 & $\times$ & EGO & MHZH-03 & scene\_20251206\_154127 &  4  &  4  &  2 / 2  &  - / -   & \checkmark & - \\
                 & $\times$ & EGO & MHZH-04 & scene\_20251206\_154926 &  8  &  8  & 1 / 7  &   - / -  & \checkmark & - \\

                \greyrule
                \multirow{4}{*}{\rotatebox{90}{\shortstack{\texttt{MHZH} \\ \texttt{EXO}}}} & $\times$ & EXO & MHZH-EXO-00 & scene\_20251210\_123247 &  4 & 4   &  0 / 4  & - / -   & \checkmark & - \\
                 & $\times$ & EXO & MHZH-EXO-01 & scene\_20251210\_123710 & 4  & 4 &  1 / 3 & - / - & \checkmark & - \\
                 & $\times$ & EXO & MHZH-EXO-02 & scene\_20251210\_123920-0 & 6 & 6 & 3 / 3  & - / - & \checkmark & - \\
                 & $\times$ & EXO & MHZH-EXO-03 & scene\_20251210\_123920-1 & 6 & 6 & 2 / 4 & - / - & \checkmark & - \\

                \greyrule
                \multirow{8}{*}{\rotatebox{90}{\texttt{HSR}}} & $\times$ & ROBOT & HSR-00 & scene\_2026-01-05-15-44-54 &  5  &  5  &  2 / 3  &  - / -  & \checkmark & - \\
                 & $\times$ & ROBOT & HSR-01 & scene\_2026-01-05-15-49-46 & 5  & 5  & 2 / 3   &  - / -  & \checkmark & - \\
                 & $\times$ & ROBOT & HSR-02 & scene\_2026-01-05-15-53-13 & 5  &  5  &  2 / 3  &  - / -  & \checkmark & - \\
                 & $\times$ & ROBOT & HSR-03 & scene\_2026-01-05-16-00-17 &  5 &  5  &  3 / 2 &  - / -  & \checkmark & - \\
                 & $\times$ & ROBOT & HSR-04 & scene\_2026-01-05-16-02-59 &  5  &  5  &  3 / 2  &  - / -  & \checkmark & - \\
                 & $\times$ & ROBOT & HSR-05 & scene\_2026-01-05-16-05-49 &  5  &  5  &  1 / 3  &  - / -  & \checkmark & - \\
                 & $\times$ & ROBOT & HSR-06 & scene\_2026-01-05-16-09-34 &  4  &  4  &  2 / 2  &  - / -  & \checkmark & - \\
                 & $\times$ & ROBOT & HSR-07 & scene\_2026-01-05-16-13-36 &  5  &  5  &  4 / 1  &  - / -  & \checkmark & - \\
                
            \bottomrule
        \end{tabularx}
    \end{threeparttable}
    \label{tab:sequence_overview}
\end{table*}

\clearpage

\end{document}